%% file: article.tex
\documentclass[default,iicol]{sn-jnl}

\jyear{2021}%

\theoremstyle{thmstyleone}%
\newtheorem{theorem}{Theorem}
\usepackage{booktabs} 
\usepackage{amsfonts,amssymb}
\usepackage{subfigure}
\usepackage{color,xcolor}
\usepackage{times}
\usepackage{epsfig}
\usepackage{amsmath}
\usepackage{textcomp}
\usepackage{gensymb}
\usepackage{wasysym}
\usepackage{multirow}
\usepackage{graphicx} 
\usepackage{colortbl}
\usepackage{algorithm}
\usepackage{algorithmicx}

\usepackage{mathrsfs}
\usepackage{url}
\usepackage{lettrine}

\input{math_commands.tex}

\usepackage{amsfonts}
\usepackage{colortbl}

\def\eg{\emph{e.g.}}
\def\ie{\emph{i.e.}}

\def\wrt{\emph{w.r.t.}}

\def\cf{\emph{c.f.}}

\newcommand{\revise}[1]{\textcolor{black}{#1}}

\newcommand{\revision}[1]{\textcolor{black}{#1}}

\theoremstyle{thmstyletwo}%

\theoremstyle{thmstylethree}%

\begin{document}

\title[Article Title]{Towards Defending Multiple $\ell_p$-norm Bounded Adversarial Perturbations via Gated Batch Normalization}


\author[1]{\fnm{Aishan} \sur{Liu}}\email{liuaishan@buaa.edu.cn}

\author[1]{\fnm{Shiyu} \sur{Tang}}\email{sytang@buaa.edu.cn}

\author[2]{\fnm{Xinyun} \sur{Chen}}\email{xinyunchen@google.com}

\author[1]{\fnm{Lei} \sur{Huang}}\email{huangleiai@buaa.edu.cn}

\author[1]{\fnm{Haotong} \sur{Qin}}\email{qinhaotong@buaa.edu.cn}

\author*[1]{\fnm{Xianglong} \sur{Liu}}\email{xlliu@buaa.edu.cn}

\author[3,4]{\fnm{Dacheng} \sur{Tao}}\email{taocheng.tao@gmail.com}

\affil[1]{\orgname{Beihang University}, \orgaddress{\state{Beijing}, \country{China}}}

\affil[2]{\orgname{Google Brain},\orgaddress{\state{ Mountain View}, \country{USA}}}

\affil[3]{\orgname{JD Explore Academy},\orgaddress{\state{ Beijing}, \country{China}}}

\affil[4]{\orgname{The University of Sydney},\orgaddress{\state{ Sydney}, \country{Australia}}}


\abstract{There has been extensive evidence demonstrating that deep neural networks are vulnerable to adversarial examples, which motivates the development of defenses against adversarial attacks. Existing adversarial defenses typically improve model robustness against individual specific perturbation types (\eg, $\ell_{\infty}$-norm bounded adversarial examples). However, adversaries are likely to generate multiple types of perturbations in practice (\eg, $\ell_1$, $\ell_2$, and $\ell_{\infty}$ perturbations). Some recent methods improve model robustness against adversarial attacks in multiple $\ell_p$ balls, but their performance against each perturbation type is still far from satisfactory. In this paper, we observe that different $\ell_p$ bounded adversarial perturbations induce different statistical properties that can be separated and characterized by the statistics of Batch Normalization (BN). We thus propose Gated Batch Normalization (GBN) to adversarially train a perturbation-invariant predictor for defending multiple $\ell_p$ bounded adversarial perturbations. GBN consists of a multi-branch BN layer and a gated sub-network. Each BN branch in GBN is in charge of one perturbation type to ensure that the normalized output is aligned towards learning perturbation-invariant representation. Meanwhile, the gated sub-network is designed to separate inputs added with different perturbation types. We perform an extensive evaluation of our approach on commonly-used dataset including MNIST, CIFAR-10, and Tiny-ImageNet, and demonstrate that GBN outperforms previous defense proposals against multiple perturbation types (\ie, $\ell_1$, $\ell_2$, and $\ell_{\infty}$ perturbations) by large margins.}




\maketitle

\input{1_introduction}
\input{2_preliminaries}
\input{3_method}
\input{4_experiments}

\input{5_discussion}
\input{6_conclusion}

\section{Acknowledge}
This work was supported by National Natural Science Foundation of China (62206009, 62022009, and 61872021), and the National Key Research and Development Plan of China (2020AAA0103502).
\bibliographystyle{unsrt}
\bibliography{0-main}


\end{document}


\title[Article Title]{Supplementary Material: Towards Defending Multiple $\ell_p$-norm Bounded Adversarial Perturbations via Gated Batch Normalization}


\author[1]{\fnm{Aishan} \sur{Liu}}\email{liuaishan@buaa.edu.cn}

\author[1]{\fnm{Shiyu} \sur{Tang}}\email{sytang@buaa.edu.cn}

\author[2]{\fnm{Xinyun} \sur{Chen}}\email{xinyunchen@google.com}

\author[1]{\fnm{Lei} \sur{Huang}}\email{huangleiai@buaa.edu.cn}

\author[1]{\fnm{Haotong} \sur{Qin}}\email{qinhaotong@buaa.edu.cn}

\par\author*[1]{\fnm{Xianglong} \sur{Liu}}\email{xlliu@buaa.edu.cn}

\author[3,4]{\fnm{Dacheng} \sur{Tao}}\email{taocheng.tao@gmail.com}

\affil[1]{\orgname{Beihang University}, \orgaddress{\state{Beijing}, \country{China}}}

\affil[2]{\orgname{Google Brain},\orgaddress{\state{ Mountain View}, \country{USA}}}

\affil[3]{\orgname{JD Explore Academy},\orgaddress{\state{ Beijing}, \country{China}}}

\affil[4]{\orgname{The University of Sydney},\orgaddress{\state{ Sydney}, \country{Australia}}}




\maketitle

\section{GBN Implemenation Details}
\label{sec:gbn_implementation}

\subsection{Network architecture}


To train models containing GBN, we add the GBN block into all layers within a model. Specifically, we set the $\rvg$ in the first GBN as the \emph{Conv gate} and use the \emph{FC gates} for other GBN blocks to further capture domain-specific information and improve model robustness. We set $\xi$ as 0.00001 and $\alpha$ as 0.1.

To empirically prove the effectiveness of the above strategy, we conduct additional experiments using the \emph{Conv gate} and \emph{FC gate} for all GBN blocks. In other words, we train a VGG-16 model with all GBN blocks using the \emph{Conv gate} denoted ``Conv$_{all}$'', and train another model with all GBN blocks using the \emph{FC gate} denoted ``FC$_{all}$''. As shown in Table~\ref{tab:conv_fc_test}, our strategy (denoted ``Conv+FC'') achieves the greatest robustness. Thus, we use \emph{Conv gate} for all the GBN blocks in the single layer study, and use \emph{Conv gate} for the first GBN and \emph{FC gates} for the other GBN blocks in the layer group study (Section 4.3 in the main body).

We conjecture that there are two reasons for this: (1) the running statistics between different domains in the first layer are almost indistinguishable. Thus, solely using the FC layer (\emph{FC gate}) fails to extract sufficient features from the first layer to perform correct classification; (2) using conv layers for all GBN blocks may suffer from over-fitting problem, since only adding GBN into the first layer achieves considerable robustness (as shown in Section 4.4 in the main body). We will further address it in the future studies.

\begin{table}[!t]
\caption{Model robustness of VGG-16 on CIFAR-10 (the higher the better).}
\label{tab:conv_fc_test}
\begin{center}
\begin{tabular}{cccc}
\toprule
& Conv$_{all}$ & FC$_{all}$ & Conv+FC \\
\hline \\
PGD-$\ell_1$&  30.4\% & 35.1\% & 59.6\%\\
PGD-$\ell_2$&  29.5\% & 34.2\% & 69.9\%\\
PGD-$\ell_{\infty}$ &  20.2\% & 32.1\% & 58.1\%\\

Clean accuracy &  41.3\% & 39.3\% & 83.6\% \\
\bottomrule
\end{tabular}
\end{center}
\end{table}

\subsection{Training details} 

During GBN training, we generate corresponding PGD-$\ell_1$, PGD-$\ell_2$, and PGD-$\ell_{\infty}$ adversarial mini-batches based on the clean mini-batch data. Here we tried two different strategies for adversarial example generation. For the first strategy, we set the BN as $eval()$ mode and let the adversarial gradient flow randomly from different BN branches; For the second strategy, we also set the BN as $eval()$ mode, while we force the adversarial gradient for each adversarial example flow through four BN branches. We found the second strategy turn out to be slightly more effective due to the higher diversity of adversarial training data.

\section{Attacks/Defenses Setups}

\subsection{Adversarial attacks}
\textbf{PGD-$\ell_1$}. We set the iteration number $k$=50. On MNIST, we set the step size $\alpha$=$\epsilon$/10; on CIFAR-10 and Tiny-ImageNet, we set the step size $\alpha$=0.05.

\textbf{PGD-$\ell_2$}. On MNIST, we set the iteration number $k$=100, and step size $\alpha$=0.1. On CIFAR-10 and Tiny-ImageNet, we set the iteration number $k$=50 and step size $\alpha$=$\epsilon$/10.

\textbf{PGD-$\ell_{\infty}$}. On MNIST, we set the iteration number $k$=50 and step size $\alpha$=0.01. On CIFAR-10 and Tiny-ImageNet, we set the iteration number $k$=40, and step size $\alpha$=$\epsilon$/10.

\textbf{PGD-1000-$\ell_{\infty}$}. On CIFAR-10, we set the perturbation magnitude $\epsilon$=0.03, iteration number $k$=1000, and step size $\alpha$=$\epsilon$/10.

\textbf{BBA}. For all datasets, we set the number of optimization steps as 1000, learning rate as 0.001, momentum as 0.8, and the binary search steps as 10.

\textbf{C\&W-$\ell_2$}. For all datasets, we set the number of optimization steps as 10000, each step size as 0.01, and the confidence required for an example to be marked as adversarial as 0.

\textbf{C\&W-$\ell_{\infty}$}. We set the number of optimization steps as 10000, each step size as 0.01, and the confidence required for an example to be marked as adversarial as 0.

\textbf{BA}. For all datasets, we set the maximum number of steps as 25000, initial step size for the orthogonal step as 0.01, and initial step size for the step towards the target as 0.01.

\textbf{MI-FGSM}. For all datasets, we set the decay factor $\mu$=1 in terms of $\ell_{\infty}$ norm, the step number $k$=10, and the step size $\alpha$=$\epsilon$/$k$.

\textbf{SPSA}. We set the maximum iteration as 100, the batch size as 8192, and the learning rate as 0.01.

\textbf{NATTACK}. We set $T$=600 as the maximum number of optimization iterations, $b$=300 for the sample size, variance of the isotropic Gaussian $\sigma^2$=0.01, and
learning rate as 0.008. 

\textbf{AutoAttack}. AutoAttack selects the following variants of adversarial attacks: APGD$_{CE}$ without random restarts, APGD$_{DLR}$, the targeted version of FAB as FAB$^T$, and Square Attack with one run of 5000 queries. We use 100 iterations for each run of the white-box attacks. For
APGD, we set the
momentum coefficient $\alpha$=0.75, $\rho$=0.75,
initial step size $\eta^{(0)}$=2$\epsilon$, where $\epsilon$ is the perturbation magnitude in terms of the $\ell_{\infty}$ norm. For FAB, we keep the standard hyper-parameters based on AdverTorch. For Square Attack, we set the initial value for the size of the squares p = 0.8.

\subsection{Adversarial defenses}

\textbf{ABS.} ABS uses multiple variational autoencoders to construct a complex generative architecture to defend against adversarial examples in the MNIST dataset. ABS uses an $\ell_0$ perturbation model of a higher radius and evaluated against $\ell_0$ attacks. So the reported number is a near estimate of the $\ell_1$ adversarial accuracy.

\textbf{AVG.} For each batch of clean data (size=64), we generate corresponding adversarial examples ($\ell_1$, $\ell_2$, and $\ell_{\infty}$) using PGD attack. We train the model using a combination of clean, $\ell_1$, $\ell_2$, and $\ell_{\infty}$ adversarial examples simultaneously. The hyper-parameters of PGD adversaries can be found in the previous sub-section.

\textbf{MAX.} For each batch of clean data (size=64), we generate the strongest adversarial examples (one of the $\ell_1$, $\ell_2$, and $\ell_{\infty}$ attack) using PGD attack. We train the model using a combination of clean examples and the strongest attack. The hyper-parameters of PGD adversaries are the same to AVG.

\textbf{MSD.} MSD creates a single adversarial perturbation by simultaneously maximizing the worst-case loss over all perturbation models at each projected steepest descent step. For MNIST, we set the iteration number $k$=100; for CIFAR-10 and Tiny-ImageNet, we set the iteration number $k$=50.

\textbf{TRADES.} TRADES is an adversarial defense method trading adversarial robustness off against accuracy which won 1st place in the NeurIPS 2018 Adversarial Vision Challenge. We set $1/\lambda$=3.0 and set other hyper-parameters as the default values following the original paper. 

\textbf{PAT.} PAT is an adversarial training method that can generalize to unforeseen perturbation types without training on them. It generates adversarial examples with a bounded neural perceptual distance to natural images and then uses the generated images to train models. Specifically, we use the proposed Fast Lagrange Perceptual Attack to conduct adversarial training, with setting attack iter=10 and perturbation bound=0.5 for MNIST and CIFAR-10 and 0.25 for Tiny-ImageNet.

We also adversarially train 3 models using PGD-$\ell_1$, PGD-$\ell_2$, and PGD-$\ell_{\infty}$ attacks, respectively. The settings are drawn from the commonly-used benchmark \cite{croce2020robustbench,tang2021robustart}. They are denoted $P_1$, $P_2$, and $P_{\infty}$.

\subsection{Normalization techniques}

\textbf{MN}. We set the number of modes in MN to 2, which achieves the best performance according to the original paper~\citep{deecke2018mode}. During training, we feed the model with a mixture of clean, $\ell_1$, $\ell_2$, and $\ell_{\infty}$ adversarial examples using the same setting as AVG.

\textbf{MBN}. The original MBN~\citep{xie2020adversarial,Xie2020intriguing} manually selects the BN branches for clean and adversarial examples during inference, which is infeasible in adversarial defense setting (the model is unaware of the type of inputs). Thus, we add the 2-way gated sub-network in MBN to predict the input domain label; we then keep the following 2 BN branches the same. During training, we compel the clean examples to go through the first BN branch and the adversarial examples (\ie, $\ell_1$, $\ell_2$, and $\ell_{\infty}$) to the second BN branch. The adversarial examples are generated via PGD using the same setting as AVG.


\section{Proof of Theorem 1}
\label{ProofofTheorem1}

\begin{theorem}
\revise{For a specific batch normalization layer, the corresponding input feature set of the dataset with $N$ adversarial perturbation types can be expressed as $\mathbb F=\left(\mathbb F_1, ..., \mathbb F_N\right)$.
When we assume that the feature $\mathbb F_k$ (generated $k$-th type data in feature) inputted to the batch normalization layer follows a Gaussian distribution $\mathcal{N}\left(\mu_k, \sigma_k\right)$ and the sampling probability of $\mathbb F$ is $\mathbf{w};\|\mathbf{w}\|_{\ell1}=1$ and $\mathbf{w}'=\mathbf{w}+\mathbf{e}_w;\|\mathbf{w'}\|_{\ell1}=1$ in two different sets $\mathbb{S}_1$ and $\mathbb{S}_2$, respectively, the difference of the mixture distribution statistics between  $\mathbb{S}_1$ and $\mathbb{S}_2$ can be expressed as $\Delta_\mu=\mathbf{e}_w\bm{\mu}^T$ and $\Delta_\sigma=\mathbf{e}_w\mathbf{t}^T-2\left(\mathbf{w}\bm{\mu}^T\right)\left(\mathbf{e}_w\bm{\mu}^T\right)-\left(\mathbf{e}_w\bm{\mu}^T\right)^2$, where $\bm{\mu}=\left(\mu_1, ..., \mu_k\right)$ and $\mathbf{t}=\left(\mu_1^2+\sigma_1^2, ..., \mu_k^2+\sigma_k^2\right)$, $\Delta_\mu=0 \ and \ \Delta_\sigma=0$ $\iff$ $\mathbf{e}_w=\mathbf{0}$.}
\end{theorem}

\begin{proof}
{Given the corresponding input feature set of the dataset with $N$ adversarial perturbation types can be expressed as $\mathbb F=[\mathbb F^1, ..., \mathbb F^N]$, where the feature $\mathbb F^k$ (generated $k$-th type data) following the Gaussian distribution $\mathcal{N}(\mu_k, \sigma_k)$.} When the sampling probability of $\mathbb F$ is $\mathbf{w};\|\mathbf{w}\|_{\ell1}=1$ in training data $\mathbf{F}_\text{train}$, the probability density function of its distribution can be expressed as

\begin{equation}
p(x)=\sum_{k=1}^{n} w_{k} p_{k}(x),
\end{equation}

where $p_{k}(x)$ is the probability density function of $\mathbb F^k$ and is expressed as

\begin{equation}
p_k(x)
=
\frac{1}{\sigma_k\sqrt{2\pi}} \, \exp \left( -\frac{(x- \mu_k)^2}{2\sigma_k^2} \right).
\end{equation}

Let $X_1, X_2, \cdots, X_N$ denote random variables with $N$ component distributions of training data sampled from $\mathbb{F}$, and $X$ denote random variables with mixture distributions. Therefore, for any function $H(\cdot)$, if $\mathbb E [H(X_k)]$ exists, and assuming that the component distribution $p_k(x)$ exists, we have

\begin{align}
\mathbb{E}[H(X)] &=\int_{-\infty}^{\infty} H(x) \sum_{k=1}^{N} w_{k} p_{k}(x) d x \\
&\revise{=\sum_{k=1}^{N} w_{k} \int_{-\infty}^{\infty} p_{k}(x) H(x) d x }\\
&=\sum_{k=1}^{N} w_{k} \mathbb{E}\left[H\left(X_{k}\right)\right]
\end{align}

and when $H(x)=(x-\mu)^i$, where $\mu$ is the mean of $X$, we have

\begin{align}
\mathbb{E}\left[(X&-\mu)^{j}\right] \revise{=\sum_{k=1}^{N} w_{k} \mathbb{E}\left[\left(X_{k}-\mu_{k}+\mu_{k}-\mu\right)^{j}\right]} \\
&=\revise{\sum_{k=1}^{N} w_{k} \sum_{i=0}^{j}\left(\begin{array}{l}
i \\
j
\end{array}\right)\left(\mu_{k}-\mu\right)^{i-j} \mathbb{E}\left[\left(X_{k}-\mu_{k}\right)^{j}\right].}
\end{align}

Therefore, the mean and variance of $X$ can be expressed as

\begin{align}
\mathbb{E}[X]=&\revise{\mu=\sum_{k=1}^{N} w_{k} \mu_{k},}\\
E\left[(X-\mu)^{2}\right]=&\revise{\sigma^{2}=\sum_{k=1}^{N} w_{k}\left[\left(\mu_{k}-\mu\right)^{2}+\sigma_{k}^{2}\right]}\\
=&\revise{\sum_{k=1}^{N} w_{k}\left(\mu_{k}^{2}+\sigma_{k}^{2}\right)-\mu^{2}}\\
=&\revise{\sum_{k=1}^{N} w_{k}\left(\mu_{k}^{2}+\sigma_{k}^{2}\right)-\left(\sum_{k=1}^{N} w_{k} \mu_{k}\right)^{2}}.
\end{align}

when we let $\mathbf{w}=[w_1, ..., w_N]$, $\bm{\mu}=[\mu_1, ..., \mu_N]$, $\bm{\sigma}=[\sigma_1, ..., \sigma_N]$, $\mathbf{t}=[t_1, ..., t_N]$, where $t_k=\mu_k^2+\sigma_k^2$, the the mean and variance of $X$ can be expressed as

\begin{align}
\mathbb{E}[X]=&\mathbf{w} \bm{\mu}^{\top},\\
E\left[(X-\mu)^{2}\right]=&\mathbf{w} \bm{t}^{\top}-(\mathbf{w} \bm{\mu}^{\top})^2.
\end{align}

And for the test data ${X}'$, when the sampling probability is expressed as $\mathbf{w}'=\mathbf{w}+\mathbf{e}_w;\|\mathbf{w'}\|_{\ell1}=1$, we have 

\begin{align}
\mathbb{E}[X']=&\mathbf{w'} \bm{\mu}^{\top}=\mathbf{w}\bm{\mu}^{\top}+\mathbf{e}_w\bm{\mu}^{\top},\\
E\left[(X'-\mu')^{2}\right]=&\mathbf{w'} \bm{t}^{\top}-(\mathbf{w'} \bm{\mu}^{\top})^2\\
=&\mathbf{w}\bm{t}^{\top}+\mathbf{e}_w\bm{t}^{\top}-\left(\mathbf{w}\bm{\mu}^{\top}+\mathbf{e}_w\bm{\mu}^{\top}\right)^2\\
=&\mathbf{w}\bm{t}^{\top}+\mathbf{e}_w\bm{t}^{\top}-\left(\mathbf{w}\bm{\mu}^{\top}\right)^2\\
&-2\left(\mathbf{w}\bm{\mu}^{\top}\right)\left(\mathbf{e}_w\bm{\mu}^{\top}\right)-\left(\mathbf{e}_w\bm{\mu}^{\top}\right)^2.
\end{align}

The we get

\begin{align}
\Delta_\mu=&\mathbb{E}[X']-\mathbb{E}[X]=\mathbf{e}_w\bm{\mu}^{\top},\\
\Delta_\sigma=&E\left[(X'-\mu')^{2}\right]-E\left[(X-\mu)^{2}\right]\\
=&\mathbf{e}_w\bm{t}^{\top}-2\left(\mathbf{w}\bm{\mu}^{\top}\right)\left(\mathbf{e}_w\bm{\mu}^{\top}\right)-\left(\mathbf{e}_w\bm{\mu}^{\top}\right)^2.
\end{align}

Since $\mathbf{t}>0$, we can trivially get that $\Delta_\mu=0 \ and \ \Delta_\sigma=0$ \iff $\mathbf{e}_w=\mathbf{0}$.

\end{proof}

\section{More Experimental Results}

\subsection{Alternative prediction approach of the gated sub-network}
\label{sec:hard_label}
In the main body of our paper, we calculate the normalized output using the gated sub-network in a soft-gated way (denoted ``soft'') based on Eqn. (3). In this section, we also try taking the top-1 prediction of $\rvg$ (the hard label) to normalize the output as an alternative approach (denoted ``hard''). As shown in Table~\ref{tab:soft_hard}, our soft-label version achieves slightly
better results than the hard-label one.

\begin{table}[!thb]
\caption{Model robustness of ResNet-20 on CIFAR-10 using different prediction approaches of the gated sub-network (the higher the better). The hyper-parameters for these attacks are the same for the main experiment.}
\label{tab:soft_hard}
\begin{center}
\begin{tabular}{cccc}
\toprule
& vanilla & hard & soft \\
\hline \\
PGD-$\ell_1$&  0.1\% & 57.8\% & 58.1\%\\
PGD-$\ell_2$&  0.0\% & 68.4\% & 68.9\%\\
PGD-$\ell_{\infty}$ &  0.0\% & 57.2\% & 58.0\%\\

Clean accuracy &  89.4\% & 79.8\% & 80.2\%\\
\bottomrule
\end{tabular}
\end{center}
\end{table}

\subsection{Adversarial Robustness against multiple perturbations}
\label{sec:adv_eval}

In this part, we provide the breakdown for each individual attack on MNIST and Tiny-ImageNet in \revise{Table} ~\ref{tab:mnist-breakdown} and ~\ref{tab:tiny-breakdown}. Further, we provide the breadkdown for each individual attack on CIFAR-10 using VGG-16, WideResNet-28-10, and ResNet-20 in Table~\ref{tab:cifar10-breakdown-vgg16}, and Table~\ref{tab:cifar10-breakdown-wrn}, and Table~\ref{tab:cifar10-res20-break}.

According to the results, our GBN outperforms other methods for almost all attacks by large margins. However, it is reasonable to notice that our GBN shows slightly weaker or comparable performance on some individual attacks compared to defenses trained for the specific perturbation types. For example, $P_1$ outperforms GBN for PGD-$\ell_1$ on CIFAR-10 and TRADES shows better performance for some $\ell_{\infty}$ attacks on MNIST.



In summary, our proposed GBN trains robust models in terms of multiple perturbation types (\ie, $\ell_1$, $\ell_2$, $\ell_{\infty}$) and outperforms other methods by large margins.

\begin{figure}[]
\vspace{-0.1in}
\centering

\subfigure[\revise{$layer1.1.bn1$}]{
\includegraphics[width=0.46\linewidth]{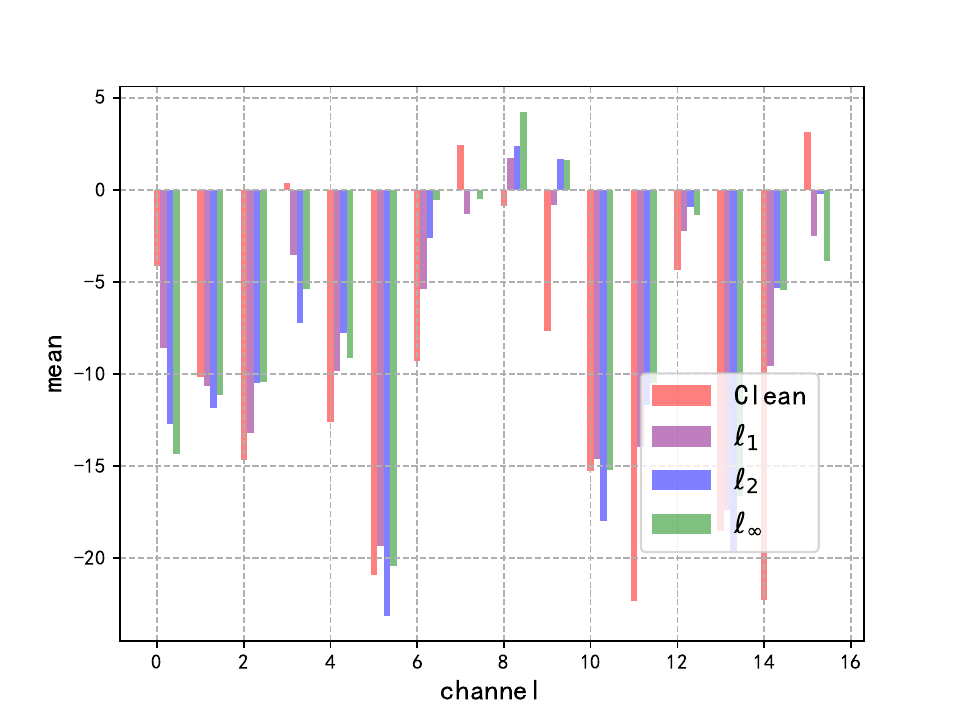}
}
\subfigure[\revise{$layer1.1.bn1$}]{
\includegraphics[width=0.46\linewidth]{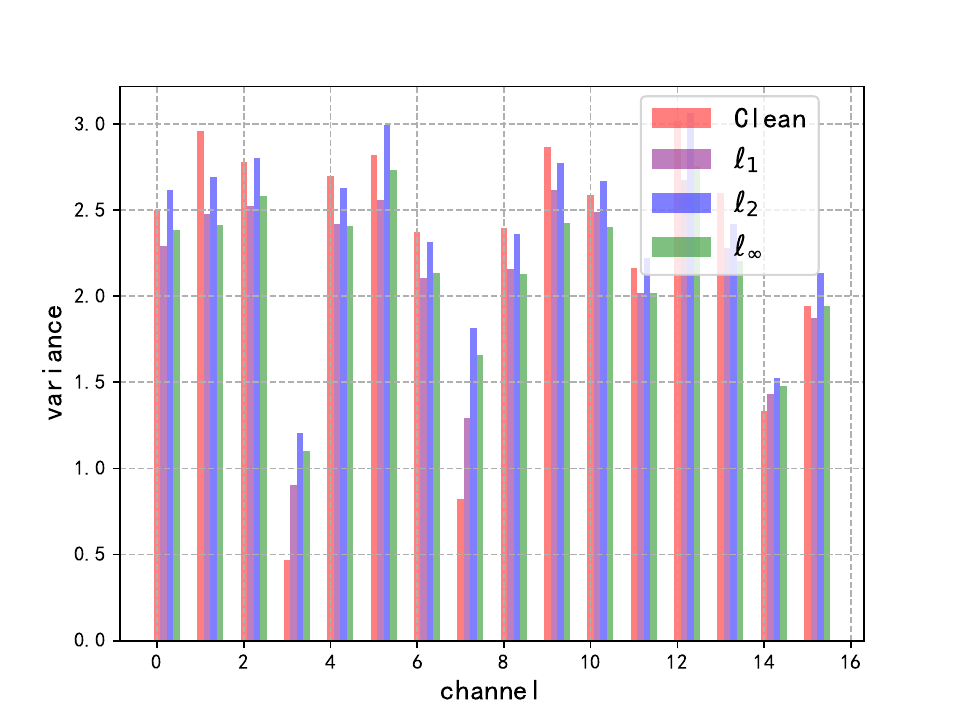}
}

\subfigure[\revise{$layer2.1.bn1$}]{
\includegraphics[width=0.46\linewidth]{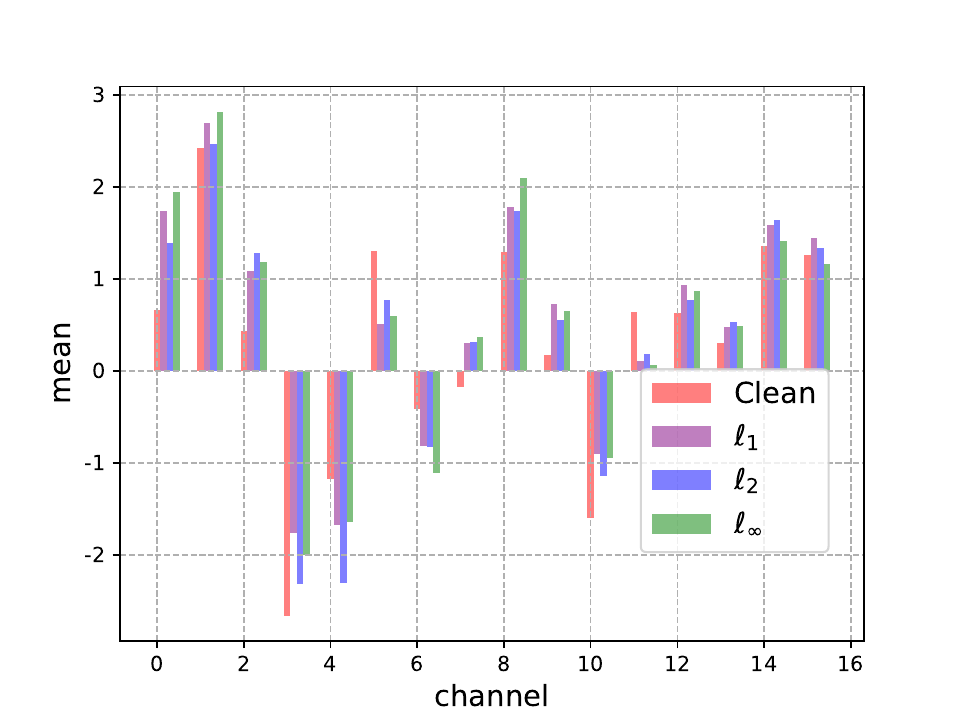}
}
\subfigure[\revise{$layer2.1.bn1$}]{
\includegraphics[width=0.46\linewidth]{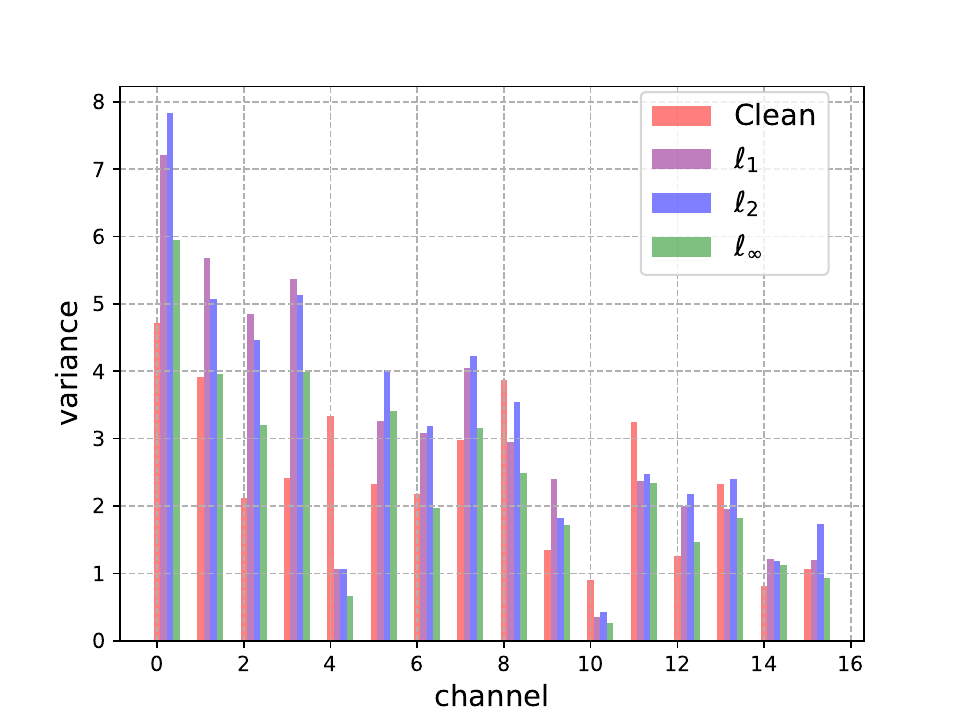}
}

\subfigure[\revise{$layer3.0.bn1$}]{
\includegraphics[width=0.46\linewidth]{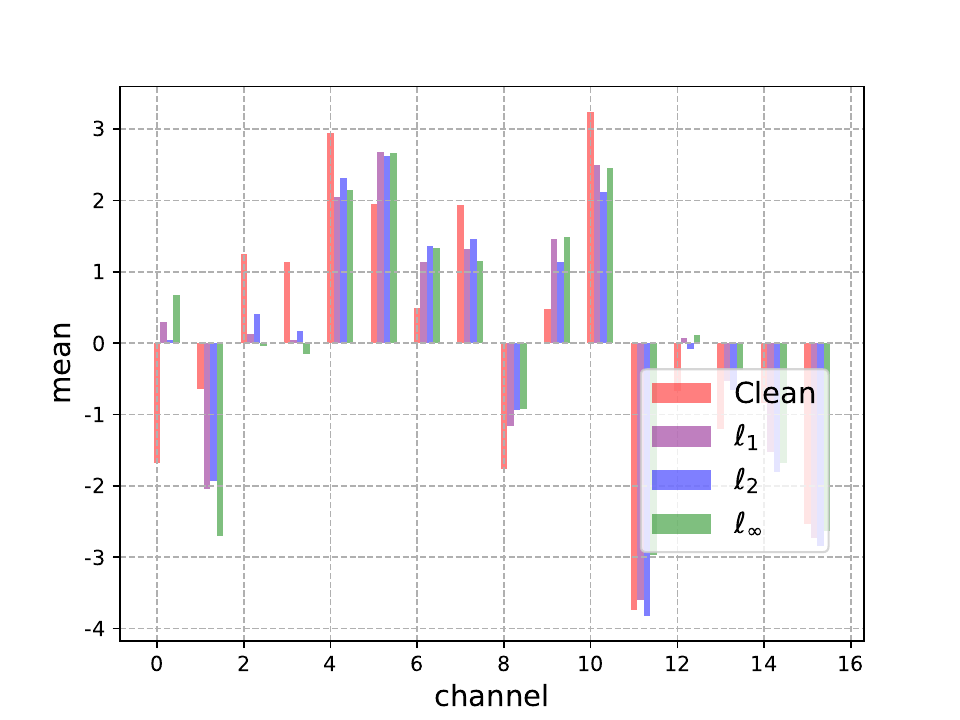}
}
\subfigure[\revise{$layer3.0.bn1$}]{
\includegraphics[width=0.46\linewidth]{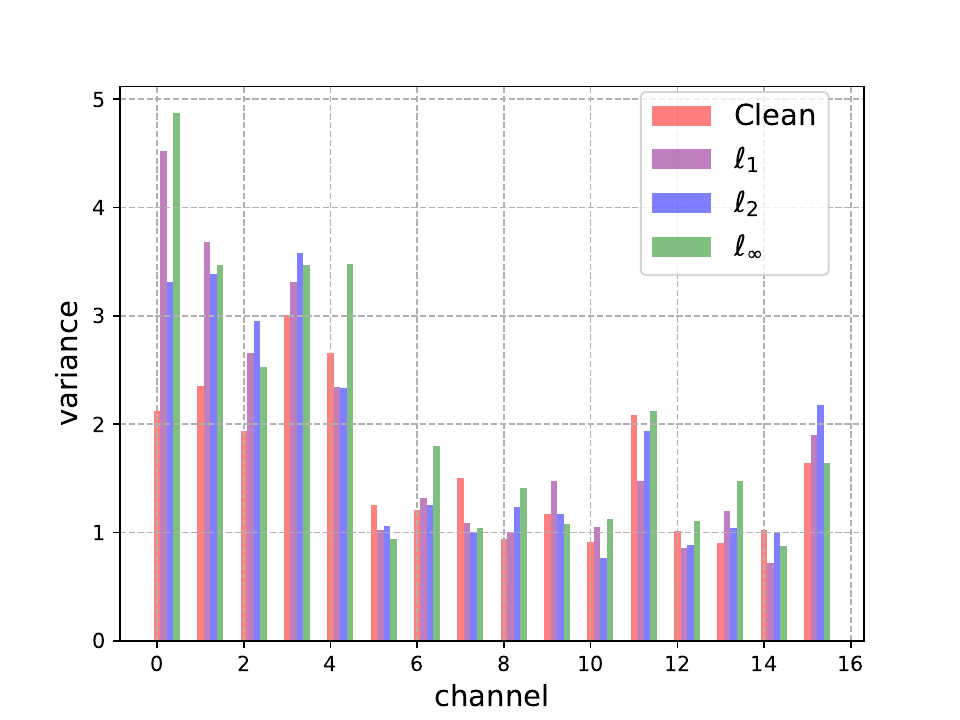}
}

\subfigure[\revise{$layer3.2.bn1$}]{
\includegraphics[width=0.46\linewidth]{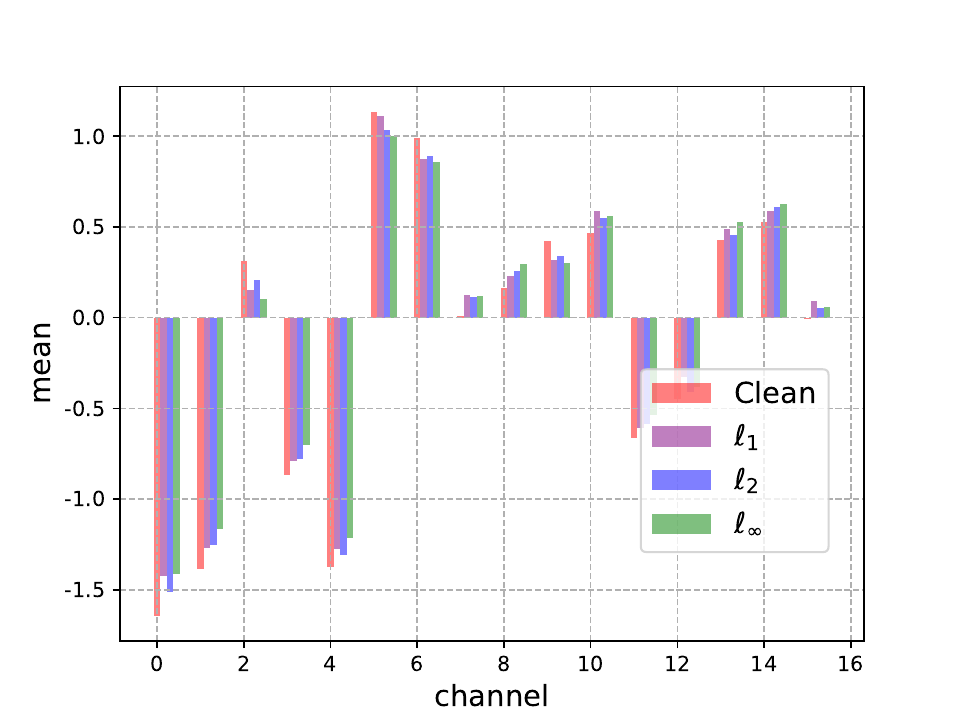}
}
\subfigure[\revise{$layer3.2.bn1$}]{
\includegraphics[width=0.46\linewidth]{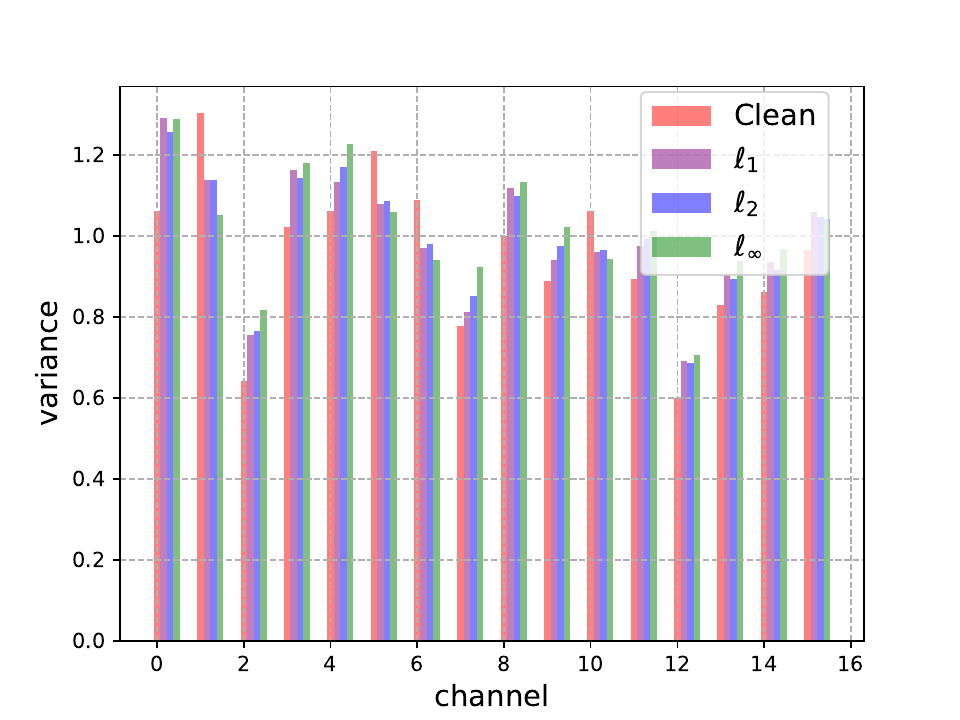}
}
\caption{\revise{Running statistics (running mean and variance) of each BN in the multiple BN branches at different layers on ResNet-20 trained on CIFAR-10.}}
\label{fig:running_res-supp}
\end{figure}

\begin{figure}[]
\vspace{-0.1in}
\centering

\subfigure[\revise{$block1.layer1.bn2$}]{
\includegraphics[width=0.46\linewidth]{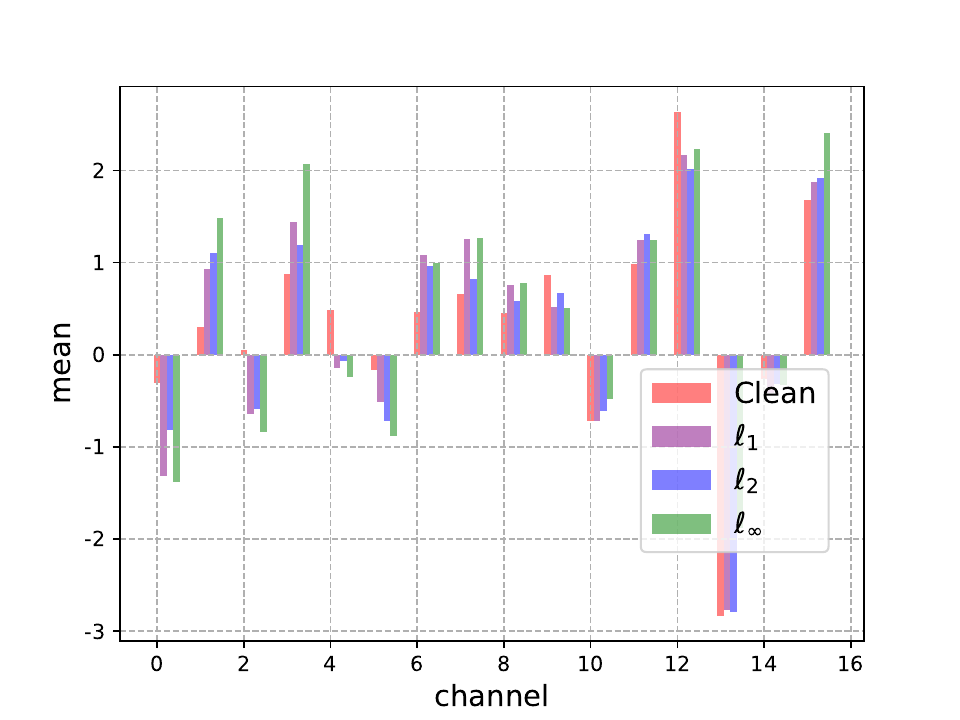}
}
\subfigure[\revise{$block1.layer1.bn2$}]{
\includegraphics[width=0.46\linewidth]{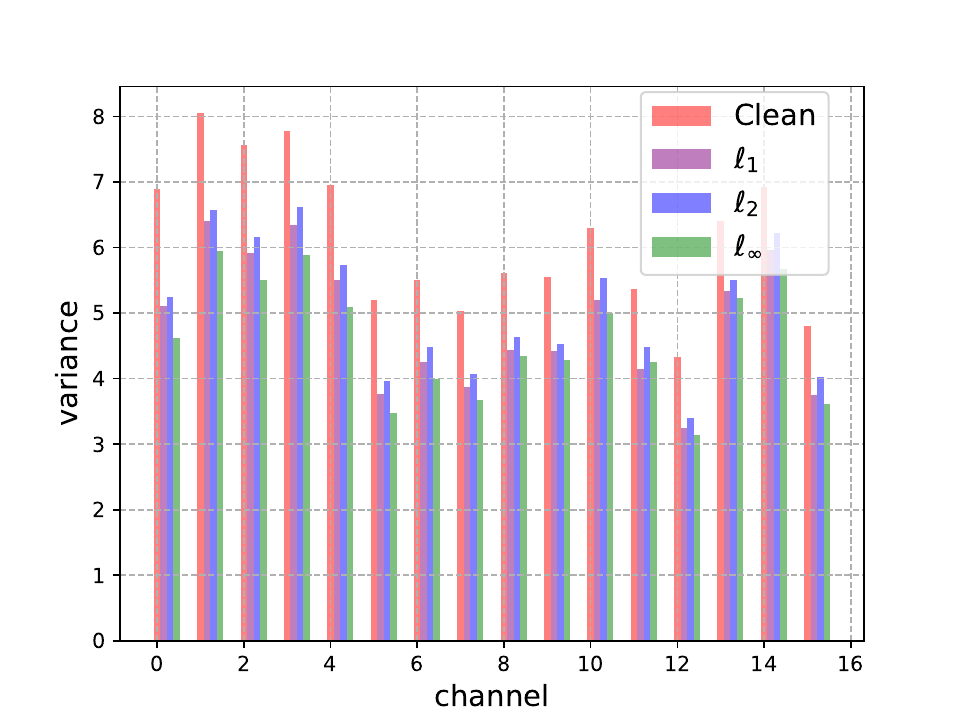}
}

\subfigure[\revise{$block2.layer1.bn1$}]{
\includegraphics[width=0.46\linewidth]{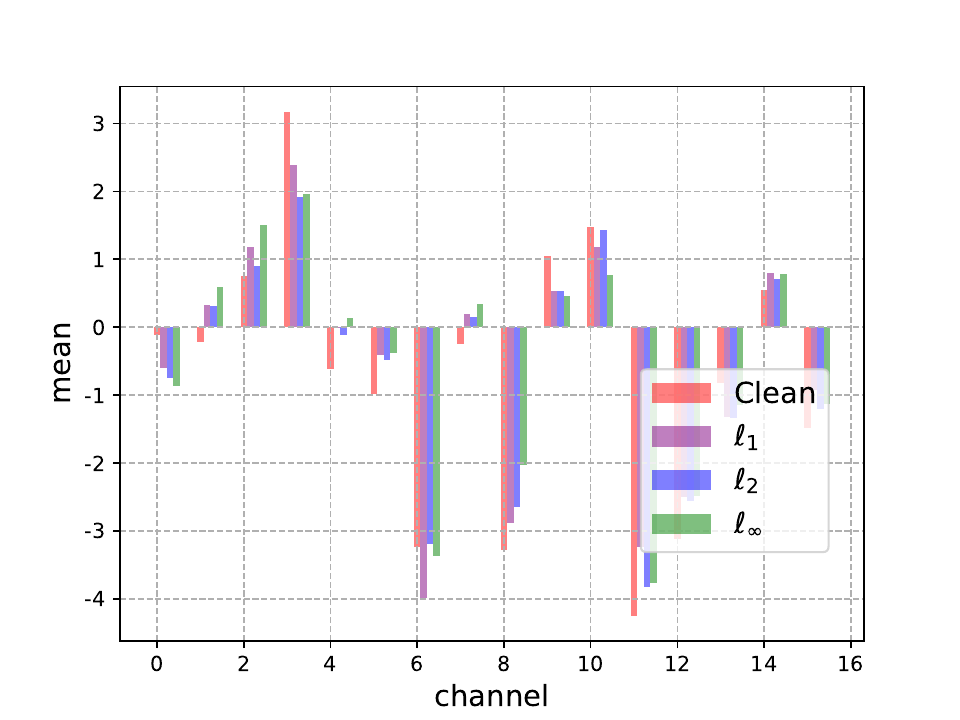}
}
\subfigure[\revise{$block2.layer1.bn1$}]{
\includegraphics[width=0.46\linewidth]{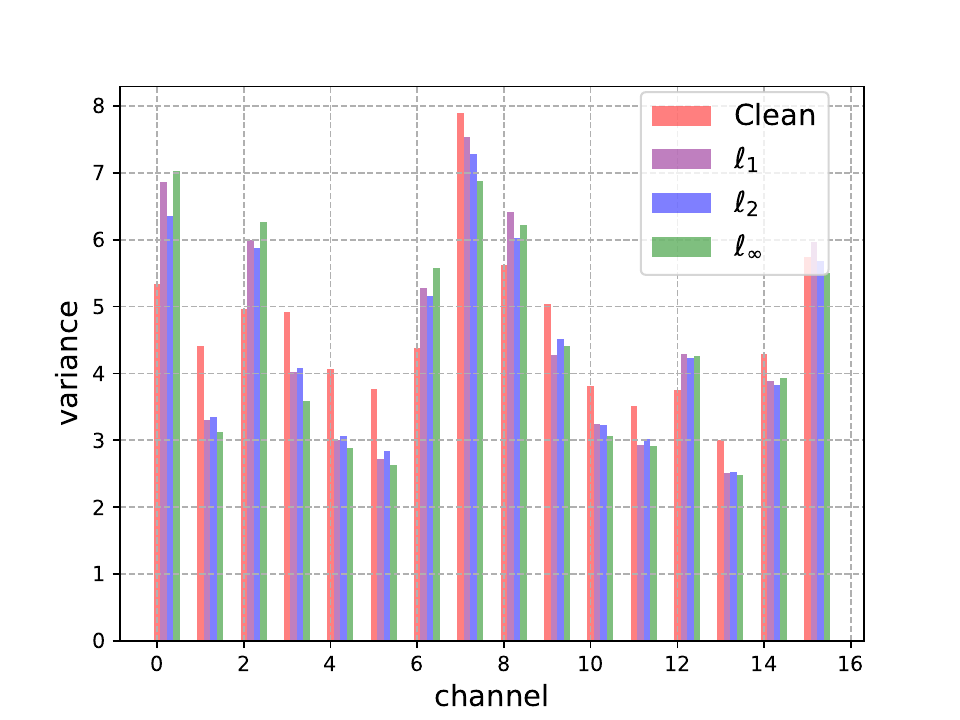}
}

\subfigure[\revise{$block2.layer2.bn2$}]{
\includegraphics[width=0.46\linewidth]{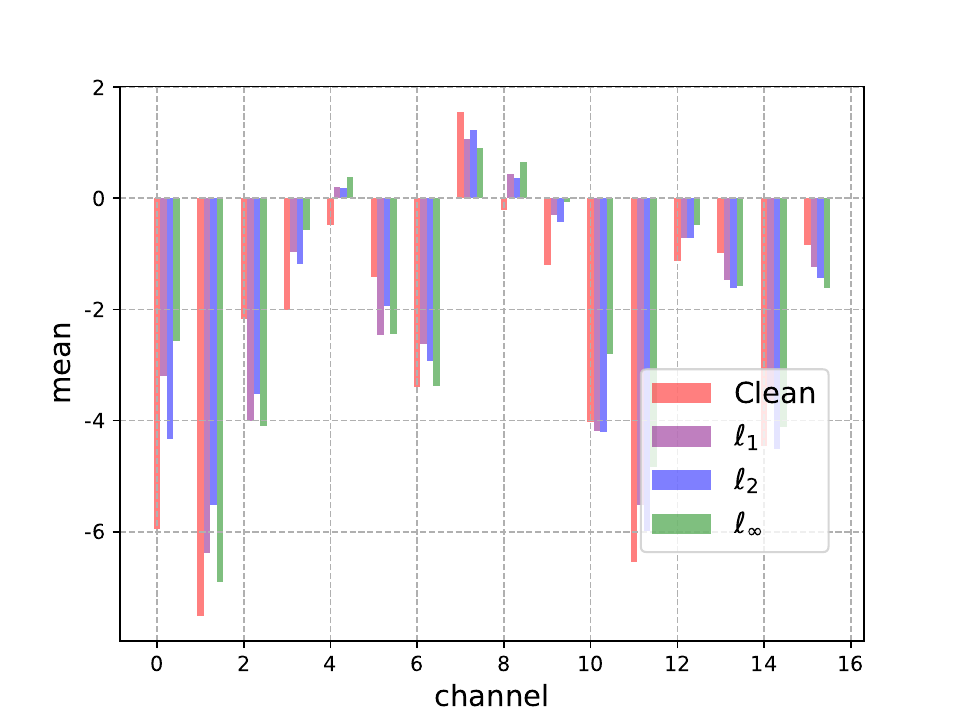}
}
\subfigure[\revise{$block2.layer2.bn2$}]{
\includegraphics[width=0.46\linewidth]{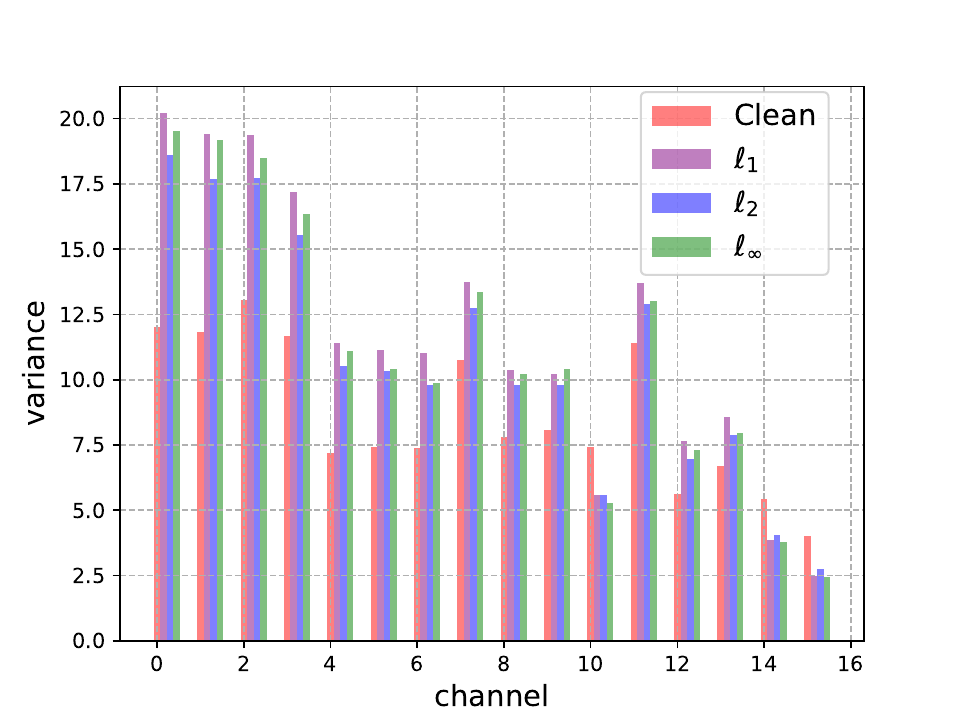}
}

\subfigure[\revise{$block3.layer3.bn1$}]{
\includegraphics[width=0.46\linewidth]{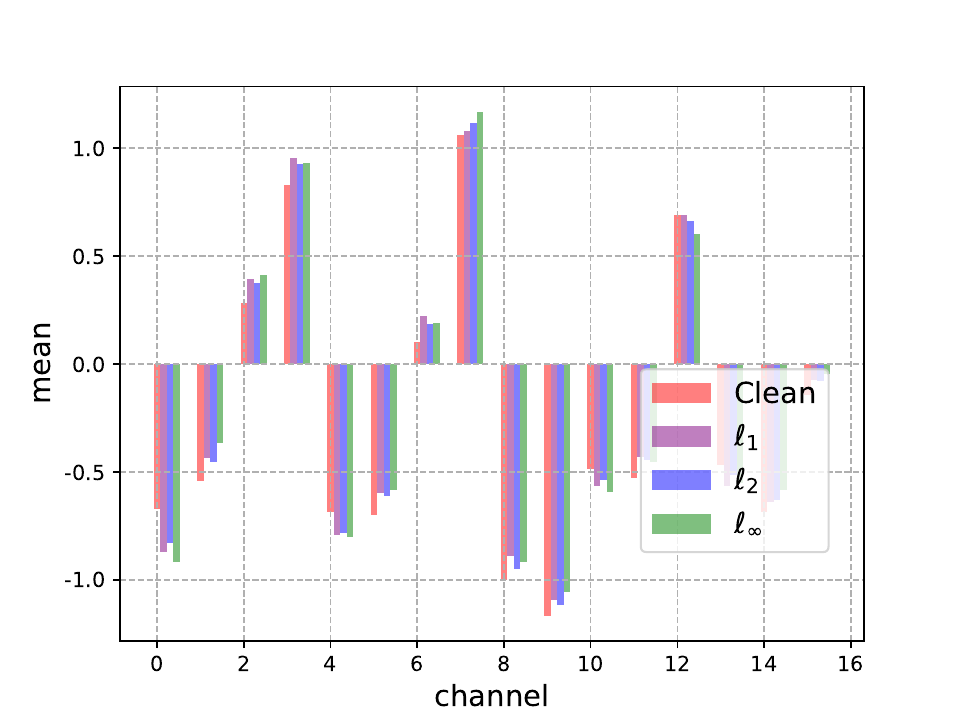}
}
\subfigure[\revise{$block3.layer3.bn1$}]{
\includegraphics[width=0.46\linewidth]{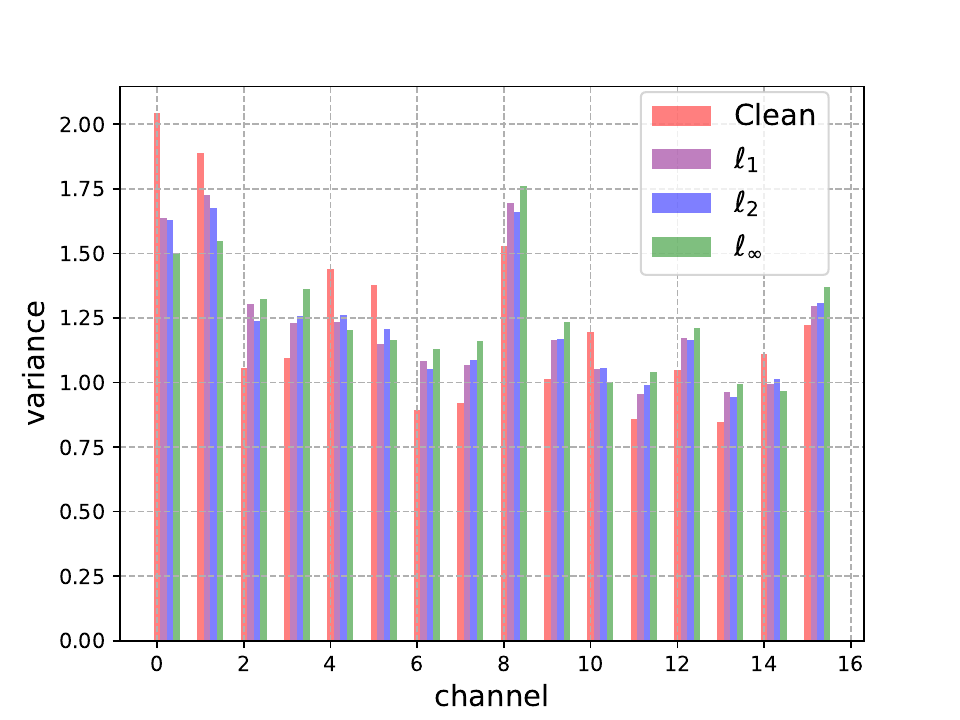}
}
\caption{\revise{Running statistics (running mean and variance) of each BN in the multiple BN branches at different layers on WideResNet-34-10 trained on CIFAR-10.}}
\label{fig:running_wide-supp}
\end{figure}

\subsection{\revise{More visualization results}}

\revise{In this part, we provide more visualization results of the running statistics of multiple BN branches on different models as shown in Figure \ref{fig:running_res-supp} and \ref{fig:running_wide-supp}.}

\begin{table*}[]
\caption{Model robustness of LeNet on MNIST over each individual attack (the higher the better).}
\label{tab:mnist-breakdown}

\begin{center}
\footnotesize
\setlength{\tabcolsep}{1.2mm}
\begin{tabular}{clc|ccccccccc|cc|c}
\toprule
& & Vanilla & PAT & TRADES & $P_1$ & $P_2$ & $P_{\infty}$ &AVG & MAX & ABS & MSD& MN & MBN & \textbf{GBN}\\
\hline \\
\multirow{2}*{$\ell_1$ attacks}& PGD-$\ell_1$ & 24.4\%& 69.2\%& 80.1\%& 58.2\%& 	77.1\%& 	52.1\%& 	77.2\%& 	76.7\% & /& 76.5\%& 79.7\%& 79.2\%&\textbf{86.1\%}\\
& BBA & 6.1\%&53.2\%& 9.6\%& 33.1\%& 	68.7\%& 	8.1\%& 	30.6\%& 	27.7\%  &/ &42.1\% & 24.7\%& 64.1\%&\textbf{79.2\%}  \\
\hline \\
\multirow{4}*{$\ell_2$ attacks}& PGD-$\ell_2$ & 24.1\%&61.2\%& 89.3\%& 30.7\%& 	68.2\%& 	62.0\%& 	73.2\%& 	73.7\% & /& 69.1\%& 76.8\%& 78.0\%&\textbf{97.4\%}\\
& C\&W-$\ell_2$ & 2.0\%&56.2\%& 42.1\%& 27.7\%& 	66.7\%& 	41.2\%& 	57.6\%& 	50.1\% & /& 71.8\%& 19.6\%& 66.7\%& \textbf{97.7\%}\\
& Gaussian Noise & 98.5\%&89.2\%& 98.1\%& 81.1\% & 98.8\% & 98.2\% & 98.7\%& 98.2\%& 97.6\%& 97.1\%&97.0\%& 98.6\%& \textbf{98.9\%} \\
& BA & 10.0\%&64.3\%& 12.1\%& 46.8\%& 	80.3\%& 	18.7\%& 	40.2\%& 	38.1\% & 83.0\%& 78.5\%& 24.1\%& 90.2\%&\textbf{97.5\%} \\
\hline \\
\multirow{6}*{$\ell_{\infty}$ attacks}& PGD-$\ell_{\infty}$ & 0.0\%&19.1\%& 95.8\%& 3.2\%& 	9.3\%& 	82.9\%& 	72.0\%& 	74.3\% & /& 51.1\%& 78.2\%& 79.8\%&\textbf{95.8\%}\\
& FGSM & 48.1\%&53.9\%& 96.1\%& 45.1\%& 	66.8\%& 	89.8\%& 	87.1\%& 	86.3\% & 34.0\%& 69.3\%&\textbf{97.1\%}& 75.2\%& 89.1\% \\
& MI-FGSM & 35.8\%&35.7\%& \textbf{96.6\%}& 14.8\%& 	41.6\%& 	85.9\%& 	81.4\%& 	80.7\% & 16.3\%& 60.3\%& 90.5\%& 39.9\%& 84.9\% \\
& SPSA & 37.2\%&6.4\%& 92.5\%& 23.3\%& 	39.0\%& 	83.5\%& 	64.0\%& 	70.6\% & /& 70.8\%& 77.7\%& 77.4\%&\textbf{98.0\%} \\
& NATTACK & 40.3\%&61.0\%& 91.3\%& 57.4\%& 	69.9\%& 	82.3\%& 	79.2\%& 	78.2\% & /& 84.9\%&89.8\%& 88.6\%& \textbf{97.0\%} \\
& AutoAttack & 0.0\%&1.0\%& {76.7\%}& 0.1\%& 	0.1\%& 	\textbf{78.7\%}& 	39.2\%& 	52.7\% & /& 38.5\%&0.0\%& 19.4\%& 71.5\% \\
\hline \\
All attacks & - &0.0\%&1.1\%& 9.2\%& 0.1\%& 	0.1\%& 	6.9\%& 	29.4\%& 	26.3\% & 16.3\%& 38.1\% &0.0\%& 18.4\%& \textbf{68.5\%}\\
\hline\\
Clean accuracy & - &\textbf{99.1\%}&  91.0\% & 98.9\% & 98.5\% & 98.7\% & 98.3\% & 98.6\%& 98.2\%& 99.0\%&97.1\% & 98.0\% & 98.4\%& 98.4\%\\
\bottomrule
\end{tabular}
\end{center}
\end{table*}

\begin{table*}[]
\caption{Model robustness of ResNet-34 on Tiny-ImageNet over each individual attack (the higher the better).}
\label{tab:tiny-breakdown}

\begin{center}
\footnotesize
\setlength{\tabcolsep}{1.2mm}
\begin{tabular}{clc|cccccccc|cc|c}
\toprule
& & Vanilla & PAT&TRADES & $P_1$ & $P_2$ & $P_{\infty}$ &AVG & MAX & MSD & MN & MBN & \textbf{GBN}\\
\hline \\
\multirow{2}*{$\ell_1$ attacks}& PGD-$\ell_1$ & 9.8\%&13.2\%& 27.3\% & 20.3\%& 	27.8\%& 	22.8\%& 	31.6\%& 	24.4\% & 8.8\%&12.3\%& 44.9\%& \textbf{55.0\%}\\
& BBA & 5.3\%&13.4\%& 24.8\% & 19.7\%& 	26.4\%& 	20.0\%& 	31.0\%& 	22.4\%  &7.2\% & 8.2\%& 36.9\%&\textbf{44.5\%}  \\
\hline \\
\multirow{4}*{$\ell_2$ attacks}& PGD-$\ell_2$ & 12.2\%&14.0\%& 33.6\% & 29.3\%& 	33.5\%& 	29.7\%& 	33.8 \%&	29.0\% & 13.5\%& 17.9\%& 31.2\%& \textbf{53.3\%}\\
& C\&W-$\ell_2$ & 10.1\%&13.3\%& 31.5\% & 30.5\%& 	32.0\%& 	28.2\%& 	33.1\%& 	28.0\% & 10.8\%& 18.5\%& 31.3\%& \textbf{44.6\%}\\
& Gaussian Noise & \textbf{53.7\%}&20.3\%& 42.7\% & 52.4\% & 53.6\% & 44.0\% & 41.2\%& 35.9\%& 26.1\%&39.9\%& 45.6\%& 42.4\% \\
& BA & 26.9\%&16.4\%& 36.3\% & 43.7 \%&	43.3\%& 	32.5\%& 	38.5\%& 	32.5\% & 13.3\%&29.6\%& 38.0\%& \textbf{43.4\%} \\
\hline \\
\multirow{6}*{$\ell_{\infty}$ attacks}& PGD-$\ell_{\infty}$ & 0.0\%&2.1\%& 12.8\% & 0.1\%& 	0.2\%& 	12.6\%& 	8.9\%& 	10.6\% & 7.5\%& 19.2\%& 39.7\%&\textbf{49.9\%}\\
& FGSM & 3.2\%&4.6\%& 15.7\% & 7.1\%& 	6.3\%& 	15.7\%& 	13.6\%& 	14.2\% & 8.7\%& 22.0\%& 41.3\%& \textbf{50.2\%} \\
& MI-FGSM & 1.3\%&3.9\%& 14.0\% & 2.3\%& 	3.8\%& 	14.1\%& 	11.5\%& 	12.8\% & 7.5\%& 14.8\%& 36.0\%&\textbf{53.1\%} \\
& SPSA & 0.3\%&3.0\%& 14.6\% & 0.4\%& 	2.3 \%&	14.0\%& 	8.8\%& 	11.8\% & 9.7\%& 29.6\%& 38.9\%& \textbf{51.5\%} \\
& NATTACK & 0.3\%&3.3\%& 15.0\% & 0.8\%& 	4.1\%&	15.0\%& 	9.9\%& 	12.5\% & 11.8\%& 29.8\%& 40.1\%& \textbf{48.0\%} \\
& AutoAttack & 0.0\%&1.8\%& 9.1\% & 0.3\%& 	0.9\%& 	8.9\%& 	6.2\%& 	7.8\% & 4.7\%& 6.6\%& 19.9\%&\textbf{39.3\%} \\
\hline \\
All attacks & - &0.0\%&1.9\%& 9.1\% & 0.0\%& 	1.3\%& 	8.9\%& 	6.4 \%&	7.5\% & 4.6\%& 6.5\%& 18.3\%& \textbf{37.7\%}\\
\hline\\
Clean accuracy & - &\textbf{54.0\%} &20.2\% & 43.0\% & 52.3\%& 54.1\%& 44.1\%& 41.2\%& 35.9\% &28.5\% & 46.3\% & 45.5\%& 43.2\% \\
\bottomrule
\end{tabular}
\end{center}
\end{table*}

\begin{table*}[]
\vspace{-0.1in}
\caption{Model robustness of VGG-16 on CIFAR-10 over each individual attack (the higher the better).}
\label{tab:cifar10-breakdown-vgg16}

\begin{center}
\footnotesize
\setlength{\tabcolsep}{1.2mm}
\begin{tabular}{clc|cccccccc|cc|c}
\toprule
& & Vanilla & PAT&TRADES & $P_1$& $P_2$ & $P_{\infty}$ &AVG & MAX & MSD & MN & MBN & \textbf{GBN}\\
\hline \\
\multirow{2}*{$\ell_1$ attacks}& PGD-$\ell_1$ & 0.5\%&45.4\%& 27.2\% & 20.3\%& 	31.7\%& 	24.8\%& 	53.2\%& 	50.6\% & 50.7\%& 45.7\%& 49.2\%& 59.6\%\\
& BBA & 0.0\% &39.4\%&18.0\% & 16.7\%& 	28.8\%& 	18.4\%& 	50.2\%& 	46.4\%  &44.3\% & 40.8\% &46.6\% &\textbf{69.0\%}  \\
\hline \\
\multirow{4}*{$\ell_2$ attacks}& PGD-$\ell_2$ & 0.1\%&62.9\%& 60.3\% & 54.4\%& 	61.0\%& 	59.0\%& 	64.0\%& 	62.0\% & 64.4\%& 49.6\%& 62.0\%& \textbf{69.9\%}\\
& C\&W-$\ell_2$ & 0.0\%&60.1\%& 60.6\% & 55.0\%& 	61.1\%& 	56.6\%& 	62.1\%& 	60.2\% & 63.5\%& 31.3\%& 21.6\%& \textbf{74.5\%}\\
& Gaussian Noise & \textbf{81.4\%}&80.2\%& 79.3\% & 69.6\% & 85.2\% & 82.5\% & 74.4\%& 62.2\%& 69.5\%& 61.4\%& 51.1\%& 79.1\% \\
& BA & 0.7\% &65.2\%& 61.2\% & 62.1\%& 	66.4\%& 	61.8\%& 	66.3\%& 	63.9\% & 61.1\% &39.4\%& 49.0\%& \textbf{74.5\%} \\
\hline \\

\multirow{6}*{$\ell_{\infty}$ attacks}& PGD-1000-$\ell_{\infty}$ & 0.0 \%&40.5\%& 49.6\% & 17.0\%& 	25.4\%& 	44.1\%& 	37.1\%& 	43.8\% & 42.2\%& 32.8\%& 56.0\%& \textbf{58.1\%}\\
& FGSM & 8.9\%&49.6\%& 55.5\% & 33.6\%& 	40.1\%& 	49.9\%& 	46.5\%& 	50.1\% & 47.1\%& 41.6\%& \textbf{63.4\%}& 57.0\% \\
& MI-FGSM & 0.3\%&46.6\%& 53.7\% & 27.2\%& 	35.2\%& 	48.5\%& 	43.2\%& 	47.8\% & 46.5\%& 32.6\%& 60.5\%& \textbf{73.3\%} \\
& SPSA & 0.4\%&40.3\%& 51.9\% & 20.1\%& 	26.5\%& 	44.6\%& 	37.6\%& 	43.5\% & 45.4\%& 39.4\%& 69.2\%& \textbf{70.7\%} \\
& NATTACK & 2.2\%&40.8\%& 49.8\% & 23.2\%& 	27.8\%& 	45.1\%& 	38.9\%& 	45.1\% & 44.8\%& 39.8\%& 45.3\%& \textbf{64.1\%} \\
& AutoAttack & 0.0\%&34.9\%&46.0\% & 11.4\%& 	21.6\%& 	40.1\%& 	33.2\%& 	37.9\% & 38.7\%& 13.6\%& 40.2\%& \textbf{51.2\%} \\
\hline \\
All attacks & - &0.0\%&33.4\%& 17.3\% & 11.2\%& 	21.1\%& 	18.8\%& 	33.1\%& 	37.6\% & 38.6\%&13.5\% & 21.0\%& \textbf{50.5\%}\\
\hline\\
Clean accuracy & - & \textbf{90.4\%}&80.4\%& 86.9\% & 84.2\% & 87.3\% & 84.6\% & 81.5\%& 78.7\%& 78.0\%& 79.8\%& 84.1\% &83.6\% \\
\bottomrule
\end{tabular}
\end{center}
\vspace{-0.1in}
\end{table*}

\begin{table*}[]
\caption{Model robustness of WideResNet-28-10 on CIFAR-10 over each individual attack (the higher the better).}
\label{tab:cifar10-breakdown-wrn}

\begin{center}
\footnotesize
\setlength{\tabcolsep}{1.2mm}
\begin{tabular}{clc|cccccccc|cc|c}
\toprule
& & Vanilla & PAT&TRADES & $P_1$& $P_2$ & $P_{\infty}$ &AVG & MAX & MSD & MN & MBN & \textbf{GBN}\\
\hline \\
\multirow{2}*{$\ell_1$ attacks}& PGD-$\ell_1$ & 0.3\%&31.4\%& 26.9\% & 19.8\%& 	33.3\%& 	15.7\%& 	56.1\%& 	52.9\% & 52.9\%& 45.2\%& 48.3\%& 61.4\%\\
& BBA & 0.0\% &36.4\%&16.3\% & 16.0\%& 	30.3\%& 	11.8\%& 	51.8\%& 	47.1\%  &44.8\% & 41.3\% &47.3\% &\textbf{71.1\%}  \\
\hline \\
\multirow{4}*{$\ell_2$ attacks}& PGD-$\ell_2$ & 0.6\%&44.5\%& 60.4\% & 56.5\%& 	63.2\%& 	56.7\%& 	66.4\%& 	65.6\% & 62.6\%& 50.7\%& 61.7\%& \textbf{70.4\%}\\
& C\&W-$\ell_2$ & 0.0\%&49.1\%& 59.8\% & 57.4\%& 	63.0\%& 	55.5\%& 	65.2\%& 	63.8\% & 63.9\%& 33.2\%& 24.2\%& \textbf{75.6\%}\\
& Gaussian Noise & \textbf{91.1\%}&61.3\%& 79.9\% & 69.6\% & 85.1\% & 82.5\% & 78.4\%& 69.8\%& 77.1\%& 63.8\%& 54.0\%& 77.4\% \\
& BA & 0.4\% &48.0\%& 60.5\% & 64.9\%& 	68.3\%& 	64.3\%& 	69.6\%& 	68.0\% & 65.1\% &41.3\%& 49.7\%& \textbf{75.4\%} \\
\hline \\

\multirow{6}*{$\ell_{\infty}$ attacks}& PGD-1000-$\ell_{\infty}$ & 0.2 \%&26.9\%& 49.9\% & 14.0\%& 	25.6\%& 	47.1\%& 	39.1\%& 	46.1\% & 46.7\%& 35.0\%& 56.2\%& \textbf{60.2\%}\\
& FGSM & 8.0\%&40.9\%& 57.8\% & 32.3\%& 	39.0\%& 	54.8\%& 	48.7\%& 	53.8\% & 51.6\%& 41.7\%& \textbf{64.6\%}& 57.9\% \\
& MI-FGSM & 0.0\%&37.2\%& 53.9\% & 24.9\%& 	33.8\%& 	52.9\%& 	45.0\%& 	51.5\% & 49.8\%& 34.9\%& 61.3\%& \textbf{70.1\%} \\
& SPSA & 0.2\%&26.4\%& 49.7\% & 14.3\%& 	26.1\%& 	49.9\%& 	40.1\%& 	47.5\% & 46.4\%& 39.5\%& 68.3\%& \textbf{70.2\%} \\
& NATTACK & 1.3\%&27.6\%& 50.1\% & 16.0\%& 	28.1\%& 	50.2\%& 	40.6\%& 	47.9\% & 45.7\%& 42.5\%& 47.2\%& \textbf{65.9\%} \\
& AutoAttack & 0.0\%&19.3\%&46.6\% & 10.5\%& 	23.1\%& 	45.5\%& 	36.1\%& 	42.2\% & 39.8\%& 15.6\%& 38.5\%& \textbf{51.8\%} \\
\hline \\
All attacks & - &0.0\%&18.2\%& 15.9\% & 10.3\%& 	23.1\%& 	11.6\%& 	35.4\%& 	42.0\% & 39.5\%&14.8\% & 24.0\%& \textbf{51.3\%}\\
\hline\\
Clean accuracy & - & \textbf{92.6\%}&61.2\%& 87.2\% & 83.2\% & 87.9\% & 84.4\% & 82.5\%& 78.6\%& 80.0\%& 83.5\%& 84.0\% &83.5\% \\
\bottomrule
\end{tabular}
\end{center}
\vspace{-0.1in}
\end{table*}

\begin{table*}[]
\caption{Model robustness (\%) of ResNet-20 on CIFAR-10 over each individual attack (the higher the better).}
\label{tab:cifar10-res20-break}

\begin{center}
\footnotesize
\setlength{\tabcolsep}{1.5mm}
\begin{tabular}{clc|cccccccc|cc|c}
\toprule
& & Vanilla & PAT&TRADES & $P_1$& $P_2$ & $P_{\infty}$ &AVG & MAX & MSD & MN & MBN & \textbf{GBN}\\
\hline \\
\multirow{2}*{$\ell_1$ attacks}& PGD-$\ell_1$ & 0.1&39.6& 25.2 & 22.2  & 34.1  & 22.6  & 47.8 & 43.9 & 49.3& 43.7& 47.3& 58.1\\
& BBA & 0.0 &34.1&15.7 & 19.3  & 32.0  & 16.3  &44.7  &41.2  &43.5 & 40.0 &45.6 &\textbf{68.6}  \\
\hline \\
\multirow{4}*{$\ell_2$ attacks}& PGD-$\ell_2$ & 0.0&50.9& 59.5 & 55.6  & 57.7  & 52.5  & 57.2 & 54.3 & 62.3& 48.9& 60.1& \textbf{68.9}\\
& C\&W-$\ell_2$ & 0.0&48.1& 58.9 & 54.7  & 56.1  &50.3  & 56.2 & 52.4 & 62.5& 29.6& 20.2& \textbf{74.9}\\
& Gaussian Noise & \textbf{81.1}&62.3& 79.3 & 68.4 & 84.7 & 81.1 & 76.2& 62.7& 68.7& 60.1& 49.1& 78.9 \\
& BA & 0.3 &51.3& 59.7 & 61.7  & 61.1  & 54.3  & 59.3 & 55.1 & 59.4 &38.1& 47.0& \textbf{73.0} \\
\hline \\

\multirow{6}*{$\ell_{\infty}$ attacks}& PGD-1000-$\ell_{\infty}$ & 0.0 &34.2& 47.5 & 14.1  & 26.7  & 41.6  &35.2 & 38.6 & 42.2& 32.1& 54.8& \textbf{58.0}\\
& FGSM & 9.2&43.1& 55.2 & 29.7  & 35.7  & 45.7  &40.5 & 41.8 & 46.1& 39.6& \textbf{62.0}& 56.0 \\
& MI-FGSM & 0.0&40.2& 53.3 & 22.4  & 32.7  & 43.6  &38.6 & 40.5 & 44.7& 30.7& 59.6& \textbf{72.2} \\
& C\&W-$\ell_{\infty}$ & 2.0&35.2& 53.3 & 19.6 & 40.2 & 50.7 &47.1& 43.8& 48.8& 41.6& 50.6& \textbf{67.1} \\
& SPSA & 0.4&34.6& 50.1 & 15.3  & 26.7  & 43.5  &34.9 & 39.0 & 44.3& 38.1& 67.5& \textbf{69.1} \\
& NATTACK & 1.8&36.7& 50.1 & 15.4  & 28.3  & 43.8  &35.5 & 40.0 & 44.6& 39.7& 45.1& \textbf{63.8} \\
& AutoAttack & 0.0&26.6&44.7 & 11.8  & 23.8  & 37.9  & 31.0 & 34.2 & 37.8& 13.0& 40.3& \textbf{50.2} \\
\hline \\
All attacks & - &0.0&24.5& 15.4 & 11.5  & 23.2  & 16.1  & 30.2 & 33.4 & 36.9&12.3 & 20.0& \textbf{48.2}\\
\hline\\
Clean accuracy & - & \textbf{89.4}&62.2& 86.2 & 83.7 & 87.2 & 83.6 & 80.0& 76.7& 78.4& 82.0& 79.1 &80.2 \\
\bottomrule
\end{tabular}
\end{center}
\end{table*}

\clearpage 
\newpage
\newpage
\newpage

\bibliographystyle{unsrt}
\bibliography{sn-bibliography}


%% file: math_commands.tex
\usepackage{amsmath,amsfonts,bm}

\def\eqref#1{equation~\ref{#1}}

\def\1{\bm{1}}

\def\rvg{{\mathbf{g}}}

\def\rvx{{\mathbf{x}}}
\def\rvy{{\mathbf{y}}}

\DeclareMathAlphabet{\mathsfit}{\encodingdefault}{\sfdefault}{m}{sl}
\SetMathAlphabet{\mathsfit}{bold}{\encodingdefault}{\sfdefault}{bx}{n}

\def\sD{{\mathbb{D}}}

\def\sR{{\mathbb{R}}}

\def\sY{{\mathbb{Y}}}



%% file: 1_introduction.tex
\section{Introduction}

Deep neural networks (DNNs) have achieved significant progress across a wide area of applications~\citep{Krizhevsky2012ImageNet,bahdanau2014neural,Hinton2012Deep}. However, they are susceptible to \emph{adversarial examples}~\citep{szegedy2013intriguing,goodfellow6572explaining}. \revise{These attacks are generated by adding human-imperceptible perturbations (often measured by $\ell_p$-norms such as $\ell_1$, $\ell_2$, and  $\ell_{\infty}$), which could easily mislead DNNs to wrong predictions leading to potential safety threats \citep{kurakin2016adversarial,Liu2019Perceptual,Liu2020Spatiotemporal}.}

To improve model robustness against adversarial perturbations, a long line of \emph{adversarial defense} methods have been proposed~\citep{Papernot2015Distillation,Engstrom2018Evaluating,goodfellow6572explaining,zhang2020interpreting}. 
Currently, most adversarial defenses are designed to counteract a single type of perturbation (\eg, small $\ell_{\infty}$-\revise{norm bounded} noise)~\citep{madry2017towards,alexey2017adversarialmachine,dong2017boosting}. These defenses offer no guarantees for other $\ell_p$-norm adversarial perturbations (\eg, $\ell_1$, $\ell_2$), and sometimes even increase model vulnerability to them~\citep{kang2019testing,Tramer2019Adversarial}. However, adversaries are not designated to generate individual-specific perturbation types and are likely to create multiple perturbations to attack the victim model in practice. To address this problem, other adversarial defense strategies have been proposed, with the goal of simultaneously achieving robustness against multiple $\ell_p$ bounded attacks, \ie, $\ell_{\infty}$, $\ell_1$, and $\ell_2$ attacks \cite{Tramer2019Adversarial,Maini2020adversarial}. Although these methods improve overall model robustness against adversarial attacks in multiple $\ell_p$ balls, the performance for each individual perturbation type is still far from satisfactory.



\begin{figure}[!t]
\begin{center}
\includegraphics[width=0.9\linewidth]{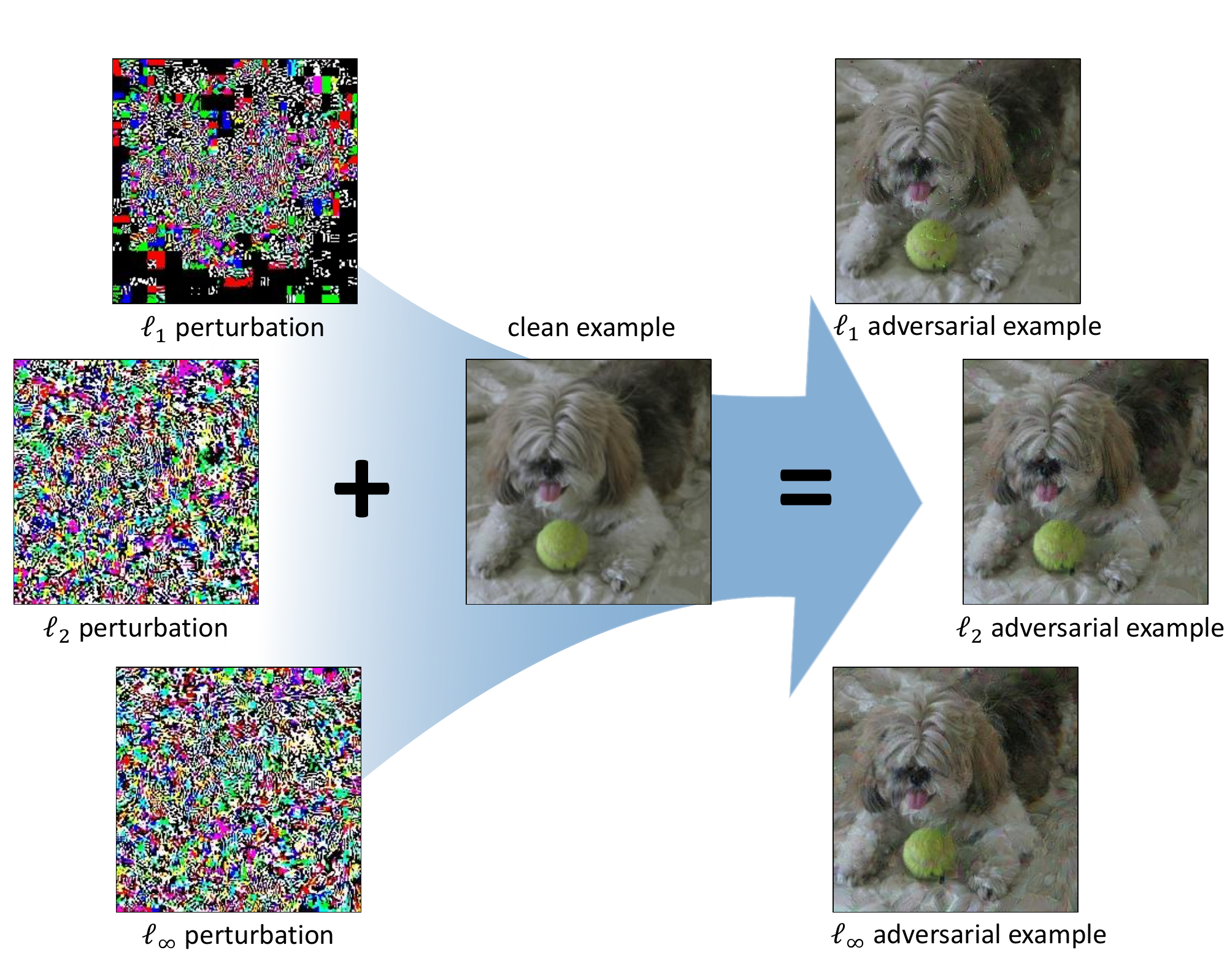}

\label{fig:multilp-front}

\end{center}
\caption{
{Visualization of clean examples and corresponding multiple types of adversarial examples ($\ell_1$, $\ell_2$, and $\ell_{\infty}$) on ImageNet. These adversarial examples are generated using PGD attacks \revise{and the perturbations are constrained by $\ell_1$, $\ell_2$, and $\ell_{\infty}$ norm metrics} (perturbations are amplified for better visualization). }}
\label{fig:BN_diff}
\end{figure}

In this work, we focus on improving model robustness against multiple $\ell_p$ bounded adversarial perturbations (such as $\ell_1$, $\ell_2$, and $\ell_{\infty}$ adversarial examples in Figure~\ref{fig:BN_diff}). Our primary observation is that different types of $\ell_p$-norm adversarial perturbations induce different batch normalization statistical properties ({\cf{} Figure~\ref{fig:append_multiBN}}) and have separable characteristics. This observation implies that different types of $\ell_p$-norm adversarial perturbations arise in different domains, which we refer to it as \emph{multi-domain hypothesis}. Through theoretical and empirical studies, we found that by simply adversarial training models using a mixture of perturbation types on a single BN, there inevitably exists a domain gap between the training and testing data, which would cause the performance degeneration on model robustness. We thus propose to defend multiple types of adversarial perturbation by designing a multiple-branch BN layer, where each BN branch is in charge of one corresponding perturbation type (\ie, domain) to ensure the normalized output are aligned towards learning perturbation-invariant (domain-invariant) representation.

One remaining problem is that the model does not know which type of perturbation is added during inference. During inference, the adversary would not tell the model which type of perturbations are generated. Therefore, we simultaneously train a gated sub-network, which aims to separate perturbation types on-the-fly. We combine the multi-branch BN layer and the gated sub-network as a building block for DNNs, referred to as \emph{Gated Batch Normalization} (GBN). GBN improves model robustness by separating perturbation-specific information for different perturbation types, and using BN layer statistics to better align data from the mixture distribution towards perturbation-invariant representations. We conduct extensive experiments on MNIST, CIFAR-10, and Tiny-ImageNet datasets, which demonstrate that our GBN approach outperforms previous defense strategies against multiple adversarial perturbations (\ie, $\ell_1$, $\ell_2$, and $\ell_{\infty}$) by large margins, \ie, 10-20\%.

Our \textbf{contributions} can be summarized as follows:

\begin{itemize}
  
    \item We for the first time observe that different types of $\ell_p$-norm adversarial perturbations induce different BN statistical properties and arise in different domains, which we refer to it as multi-domain hypothesis.
  
    \item Based on our hypothesis, we propose Gated Batch Normalization (GBN) consisting of a multi-branch BN layer and a gated sub-network, to adversarially train an invariant predictor for defending multiple $\ell_p$-norm adversarial examples.
    
    \item Extensive experiments are conducted to comprehensively evaluate the proposed GBN approach, which demonstrates that our GBN outperforms other baselines against multiple perturbation types by large margins (10-20\% \revise{absolute improvement}). 
\end{itemize}


%% file: 2_preliminaries.tex
\section{Background and Related work}
In this section, we first give the notation in this paper; then we provide a brief overview of existing work on adversarial attacks and defenses, as well as batch normalization techniques.

\subsection{Notation}
We use the following notation in this paper.

\textbf{Input space.} Let $\mathbb D$ $\subset$ $\sR^d$ be the input space. Let $\rvx$ be an input where $\rvx$ $\in$ $\mathbb D$, and $\rvy$ $\in$ $\sY$ represents the corresponding label for the instance. \revise{This paper considers the visual recognition problem and the input space should satisfy $\mathbb D$ = $[0, ..., 255]^d$.} 

\textbf{Deep learning model.} A DNN $f_{\Theta}$ is a function $\mathbb D \rightarrow \sY$ that maps the input space $\mathbb D$ to the labels $\sY$, where ${\Theta}$ denotes the parameters of the model. In this paper, we focus on the image classification task.

\textbf{Loss function.} We use $\ell(f_{\Theta}(\rvx),\rvy)$ to denote the loss function for the DNN $f_{\Theta}$ \wrt{} input $\rvx$ with ground truth label $\rvy$. \revise{Moreover, we use $\ell(\cdot)$ with subscripts to denote different loss functions in this paper.}

\textbf{Adversarial example.} We use $\rvx_{adv}=\rvx+\delta$ to denote an adversarial example. The added perturbation $\delta$ could make DNNs misclassify the input into wrong labels, \ie, $f_{\Theta}(\rvx+\delta) \neq f_{\Theta}(\rvx)$. \revise{Specifically, given a DNN $f_{\Theta}$ and an input image $\mathbf{x} \in \mathbb D$ with the ground truth label $\rvy \in \sY$, an adversarial example $\mathbf{x}_{adv}$ satisfies
\begin{equation}
\label{eqn:adv}
f_{\Theta}(\mathbf x_{adv}) \neq \mathbf y  \quad s.t. \quad \|\mathbf x-\mathbf x_{adv}\| \leq \epsilon,
\end{equation}
where $\|\cdot\|$ is a distance metric.}

\textbf{Different $\ell_p$-norm adversarial examples.} \revise{From Eqn~\ref{eqn:adv}, adversarial attacks can be often categorized based on the types of perturbations measured by different distance metrics $\|\cdot\|$ (\eg, $\ell_{p}$-norms bounded perturbations).  In this paper, we primarily consider $\ell_1$, $\ell_2$, and $\ell_{\infty}$ adversarial attacks by solving Eqn~\ref{eqn:adv}, where the perturbations are measured and constrained by $\ell_1$, $\ell_2$, and $\ell_{\infty}$ norms, respectively.} In particular, let $\mathbb D^0$ denote the set of clean examples, and $\mathbb D^{k} (k=1, ..., N)$  denote the set of adversarial examples generated by the $k$-th adversarial perturbation type. An adversarial example of the $k$-th type $\mathbf x^k_{adv}$ is generated by the pixel-wise addition of the perturbation $\delta^k$, \ie, $\mathbf x^k_{adv}=\mathbf x+\delta^k$. The perturbations $\delta^k$ are measured by different $\ell_p$-norms.

\subsection{Adversarial attacks}
\label{sec:adv}
Adversarial examples are perturbed inputs intentionally designed to mislead DNNs~\citep{szegedy2013intriguing,goodfellow6572explaining}. These generated perturbations are human imperceptible, but could easily lead DNNs to wrong predictions. 

Extensive studies have been proposed to optimize adversarial perturbations so that they could generate adversarial examples to attack deep learning models \cite{szegedy2013intriguing,goodfellow6572explaining,Liu2019Perceptual,liu2020bias,Croce2020scaling,Wei2022sparse}. These imperceptible perturbations could easily make DNNs misclassify the input images. Besides adversarial perturbations, adversarial patches are designed to attack DNNs by attaching additional stickers for their feasibility in the physical world \cite{brown2017adversarial,liu2020bias,Wang_2021_CVPR}, including patches \cite{brown2017adversarial}, camouflages \cite{Wang_2021_CVPR}, and light \cite{duan2021adversarial}. In general, adversarial attacks can be roughly categorized into \emph{white-box} and \emph{black-box} manners. For \emph{white-box} attacks, adversaries have complete knowledge of the target model and can fully access it \cite{goodfellow6572explaining,madry2017towards,liang2020efficient}; for \emph{black-box} attacks, adversaries have limited model knowledge and can not directly access the model \cite{Jonathan2018SPSA,li2019nattack,liu2023x,wei2018transferable,liang2022parallel}.

\subsection{Adversarial defenses}
Adversarial defenses aim to improve model robustness against adversarial attacks, which play important roles in increasing the availability of DNNs. In the adversarial machine learning literature, various defense approaches have been proposed to improve model robustness against adversarial examples~\citep{Papernot2015Distillation,xie2018mitigating,madry2017towards,liao2018defense,pmlr-v70-cisse17a,Goswami2019Detecting,Shao2022Openset}. Among them, adversarial training has been widely studied and demonstrated to be the most effective defense strategy ~\citep{goodfellow6572explaining,madry2017towards,liu2019training}. Specifically, adversarial training minimizes the worst case loss within some perturbation regions for classifiers, by augmenting the training set with adversarial examples. Formally, given a deep neural network $f_{\Theta}$, training data $\{\mathbf x^{(i)},\mathbf y^{(i)}\}_{i=1...n}$, the robust optimization problem is to solve the min-max problem as

\begin{align*}
\min_\Theta \sum_i \max_\delta \ell(f_{\Theta}(\mathbf x^{(i)}+\delta),\mathbf y^{(i)}),
\end{align*}

\noindent where $\delta$ is bounded by $\ell_p$ perturbations with radius $\epsilon$. To solve the inner optimization, the typical solution is addressed by using a form of adversarial attacks \citep{goodfellow6572explaining,madry2017towards}.  Specifically, adversarial training with PGD attacks, which incorporates adversarial examples generated by PGD attack into training, has so far remained empirically robust \citep{madry2017towards}.

However, these defenses only improve model robustness for specific $\ell_p$-norm perturbation (\eg, $\ell_{\infty}$) and typically offer no robustness guarantees against other $\ell_p$-norm attacks~\citep{kang2019testing,Tramer2019Adversarial,Schott2019Towards}. To address this problem, recent works have attempted to improve the robustness against several types of $\ell_p$-norm perturbation. \cite{Schott2019Towards} proposed Analysis by Synthesis (ABS), which used multiple variational autoencoders to defend $\ell_0$, $\ell_2$, and $\ell_{\infty}$ adversaries. However, ABS only works on the MNIST dataset. \cite{Croce2020Provable} proposed a provable adversarial defense against all $\ell_p$ norms for $p\geq$1 using a regularization term. However, it is not applicable to the empirical setting, since it only guarantees robustness for very small perturbations (\eg, 0.1 and $2/255$ for $\ell_2$ and $\ell_{\infty}$ on CIFAR-10). \cite{Tramer2019Adversarial} tried to defend against multiple perturbation types ($\ell_1$, $\ell_2$, and $\ell_{\infty}$) by combining different types of adversarial examples for adversarial training. Specifically, they introduced two training strategies, ``MAX'' and ``AVG'', where for each input image, the model is either trained on its strongest adversarial example or all types of perturbations. More recently, \cite{Maini2020adversarial} proposed multi steepest descent (MSD), and showed that a simple modification to standard PGD adversarial training improves robustness to $\ell_1$, $\ell_2$, and $\ell_{\infty}$ adversaries. There also exist another line of work, which focuses on improving model robustness against multiple non-$\ell_p$ adversarial attacks (\eg, adversarial patches) and unseen perturbations (\eg, corruptions)~\citep{Laidlaw2021perceptual,Lin2020dual,kang2019testing}. In this work, we follow~\citep{Tramer2019Adversarial,Maini2020adversarial} to focus on defense against $\ell_1$, $\ell_2$, and $\ell_{\infty}$ adversarial perturbations. \revise{In contrast to the previous studies, our paper proposes a different solution as (1) we propose the multi-domain hypothesis and treat different $\ell_p$-norm adversarial examples as different data domains for better defense, while previous studies treat them similarly; (2) we propose GBN as a defense solution that first separates different perturbations on-the-fly and then uses different BN branches to normalize them into aligned features, while others directly feed different attacks into models for adversarial training; and (3) we outperform others against multiple perturbation types by large margins (10-20\% absolute improvement).}

Another concurrent work~\citep{maini2021perturbation} proposed a two-stage pipeline with multiple models by categorizing the input and then sending them to the corresponding constitute model. Our paper differs from ~\citep{maini2021perturbation} significantly: (1) Motivation. \cite{maini2021perturbation} tries to defend multiple perturbations by employing several $\ell_p$ adversarially-trained models, while our GBN aims to train a (single) perturbation-invariant model for tackling multiple perturbations. (2) Technical implementation. \cite{maini2021perturbation} first categorizes input examples and then feeds them to corresponding specialized robust predictors (different $\ell_p$ adversarially-trained models). In contrast, at each layer of the model, our GBN first uses a gated subnetwork to separate the perturbations on-the-fly and then uses different BN branches to normalize them into aligned features for the learning of subsequent convolutional layers. (3) Usability. Our GBN could be embedded into any networks with batch normalization and can be combined with other defenses, while \cite{maini2021perturbation} relies on other adversarially trained models.


\subsection{Batch normalization}
\label{sec:BN_works}
BN~\citep{Ioffe2015BN} is typically used to stabilize and accelerate DNN training. Besides, a number of normalization techniques have been proposed to improve BN for style-transfer~\citep{huang2017adain}, domain adaption~\citep{Li2017revisiting,chang2019domain,deecke2018mode,huang2020normalization}, training of GANs~\citep{NIPS2017_6fab6e3a}, and building efficient DNNs~\citep{Li2020eagleeye}. Let $\rvx \in \mathbb{R}^d$ denote the input to a neural network layer. During training, BN normalizes each neuron/channel within $m$ mini-batch data by
\begin{equation}
\label{eqn:BN}
\hat{\rvx}_j=BN(\rvx_j)=\gamma_j \frac{\rvx_j - \mu_j}{\sqrt{\sigma^2_j +\xi}} + \beta_j, ~~j=1,2, ..., d,
\end{equation}
where $\mu_j=\frac{1}{m}  \sum_{i=1}^{m}  \rvx_j^{(i)}$ and $\sigma^2_j = \frac{1}{m}   \sum_{i=1}^{m} (\rvx^{(i)}_j-\mu_j)^2 $ are the mini-batch mean and variance for each neuron, respectively, and $\xi$ is a small number to prevent numerical instability. The learnable parameters $\gamma$ and $\beta$ are used to recover the representation capacity.
During inference, the population statistics of mean $\hat{\mu}$ and variance $\hat{\sigma}^2$ are used in Eqn.~\ref{eqn:BN}, which are usually calculated as the running average over different  training iterations $t$ with update factor $\alpha$:
\begin{equation}
    \label{eqn:running-average}
\begin{aligned}
\begin{cases}
\quad \hat{\mu}^t=  (1-\alpha) \hat{\mu}^{\revise{(t-1)}} + \alpha \mu^{t-1},\\
\quad (\hat{\sigma}^t)^2=  (1-\alpha) (\hat{\sigma}^{\revise{(t-1)}})^2 + \alpha (\sigma^{\revise{(t-1)}})^2.
\end{cases}
\end{aligned}
\end{equation}

\revise{For the convolutional input $\mathbf{X}\in \mathbb{R}^{d\times h \times w}$, where $h$ and $w$ are the height and width of the feature map, BN jointly normalize all the activations in a batch, over all locations~\citep{Ioffe2015BN}. That is, the normalization operation in Eqn.~\ref{eqn:BN} is performed within the effective mini-batch of size $mhw$.} 

We will provide comparisons of our GBN and other normalization methods in Section~\ref{sec:apprach_discussion} and \ref{sec:main_exp}.

%

%% file: 3_method.tex
\section{Gated Batch Normalization}

In this section, we first provide the problem definition, and then briefly illustrate our multi-domain hypothesis, which states that different $\ell_p$ bounded adversarial examples are drawn from different domains. Motivated by that, we propose Gated Batch Normalization (GBN), which improves model robustness against multiple adversarial perturbations.

\subsection{Problem Definition}

{\textbf{Goal}.} For an image classification model, given input image $\rvx$, a model $f_\Theta$ is designed to predict the ground truth label $\rvy$ by solving the following problem:
\begin{equation}
    \min_\Theta{\sum_{i=1}^{n}{\ell(f_{\Theta}(\rvx^{(i)},\rvy^{(i)})\revise{)}}},
\end{equation}
where $n$ denotes the number of input samples.

The purpose of our proposed GBN approach is to defend against multiple $\ell_p$-norm adversarial perturbations, \ie, provide correct predictions given adversarially perturbed inputs as

\begin{equation}
    \min_\Theta{\sum_{i=1}^{n}\sum_{k=1}^{N}{\ell(f_{\Theta}(\rvx^{(i)}+\delta^k,\rvy^{(i)})\revise{)}}},
\end{equation}
where $n$ denotes the number of input samples, and $N$ represents perturbation type numbers. Formally, we aim to build robust DNNs through adversarial training scheme by solving the min-max optimization as
\begin{equation}
    \min_\Theta{\sum_{i=1}^{n}\sum_{k=1}^{N}\max_{\delta^k}{\ell(f_{\Theta}(\rvx^{(i)}+\delta^k,\rvy^{(i)})\revise{)}}}.
\end{equation}

{\textbf{Threat model}.} In this paper, we consider the two commonly-adopted scenarios and assumptions, \ie, white-box and black-box. We first consider the basic \emph{white-box} threat model, where adversaries have direct access to the model including architecture and parameters. Meanwhile, we also assume that adversaries do not have access to the model, and need to perform \emph{black-box} attacks. Our defense aims to provide model robustness against multiple $\ell_p$-norm adversarial examples in both white-box and black-box scenarios. To verify our defense ability, for white-box assumption, we employ commonly-used gradient-based attacks (\cf{} Section \ref{sec:main_exp}) and adaptive attacks that are specially designed to break our GBN block (\cf{} Section \ref{sec:adaptive-attack}); for black-box assumption, we adopt several gradient-free and black-box attacks (\cf{} Section \ref{sec:main_exp}).

\subsection{Motivation: Multi-domain hypothesis}
\label{sec:multi-domain}

We assume $N$ \emph{adversarial perturbation types}\footnote{In this work, we consider $N=3$ adversarial perturbation types: $\ell_1, \ell_2$, and $\ell_{\infty}$. }, each characterized by a set $\mathbb S^k$ of perturbations for input $\mathbf x$. Our hypothesis states that different $\ell_{p}$ norm bounded adversarial perturbations $\mathbb D^k$ (for all $k\geq 0$, including clean examples) are drawn from different domains.

\begin{figure}[!htb]
\vspace{-0.15in}
\begin{center}

\subfigure[\revise{multiple BN branches}]{
\includegraphics[width=0.48\linewidth]{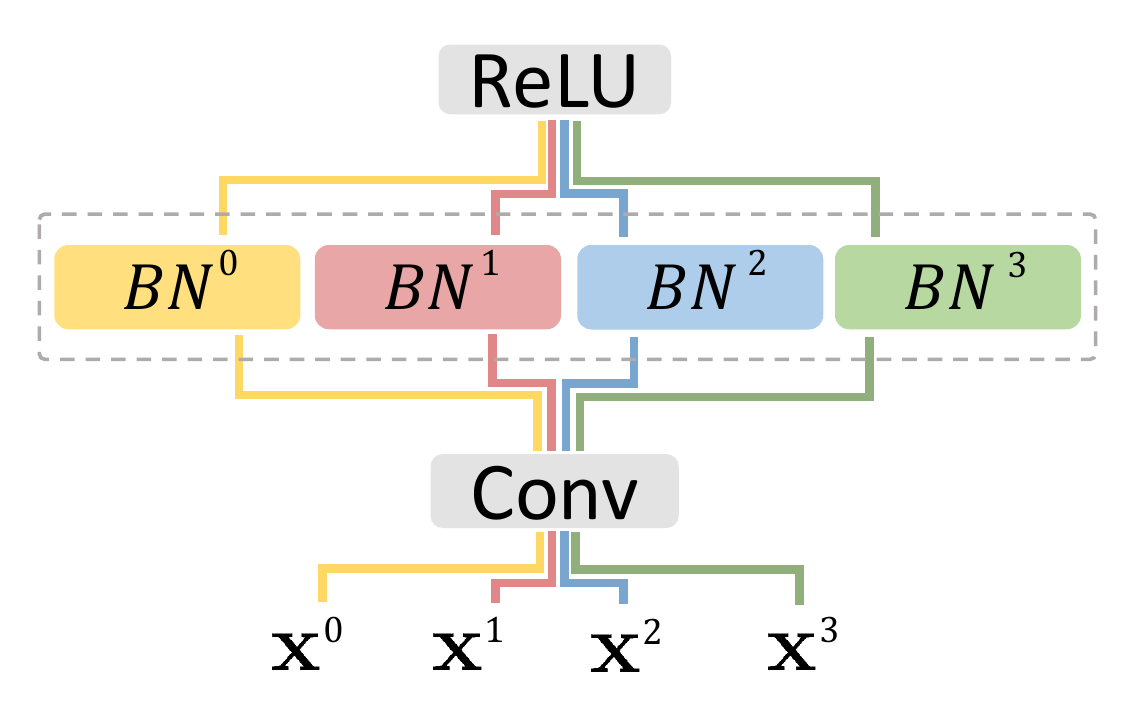}
\label{fig:res-multiple-bn1}
}
\subfigure[running means]{
\includegraphics[width=0.46\linewidth]{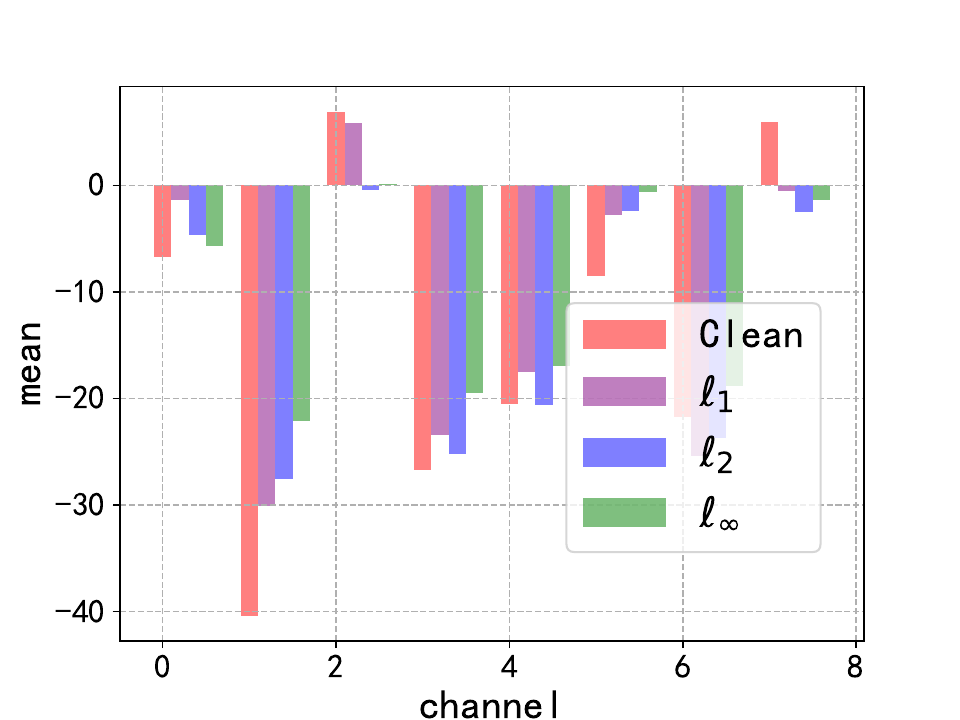}
\label{fig:res-multiple-motivate}
}

\end{center}
\caption{
\revision{(a): the structure with 4 BN branches, which use different mini-batches from different $\sD^k$ to estimate the normalization statistics. \revision{(b): running means of multiple BN branches on 8 randomly sampled channels in a VGG-16's $conv2\_1$ layer.}}}
\vspace{-0.1in}
\label{fig:append_multiBN}
\end{figure}

We first empirically verify the hypothesis by training a DNN with each BN layer composed of several BN branches. Specifically, we train a DNN on CIFAR-10 with each BN layer composed of 4 branches (\ie, we replace each BN with 4 BN branches, \revision{the structure multiple BN branches is shown in Figure \ref{fig:res-multiple-bn1}}). During training, we construct different mini-batches for clean, $\ell_1$, $\ell_2$, and $\ell_{\infty}$ adversarial images (\ie, $\sD^0$, $\sD^1$, $\sD^2$, and $\sD^3$) to estimate the normalization statistics of each branch of BN; during inference, given a mini-batch of data from a specific domain, we manually activate the corresponding BN branch and deactivate the others. We update the running statistics of each BN branch separately but optimize the model parameters using the sum of the four losses (\ie, clean, $\ell_1$, $\ell_2$, and $\ell_{\infty}$ adversarial examples). The results in \revision{Figure \ref{fig:res-multiple-motivate}} have shown that different $\sD^k$ induce significantly different running statistics, according to the different running means and variances. More studies including running statistics results and Fourier analysis can be found in Section \ref{sec:multi-abla}. Thus, training on a single perturbation type (domain) is insufficient for achieving the robustness against other types of perturbations (domains).  

Based on the above analysis, an intuitive solution is to training on a mixture of different perturbation types. However, during inference, adversaries are likely to generate different perturbation types with different ratios. According to our Theorem~\ref{th:dist} (\emph{c.f.} \emph{proof in supplementary material}), by training with a mixture of perturbation types on a single BN, there inevitably exists a domain gap between training and testing data, which causes the performance degeneration~\citep{Benz2021revisiting}.



\begin{theorem}
\label{th:dist}
\revise{For a specific batch normalization layer, the corresponding input feature set of the dataset with $N$ adversarial perturbation types can be expressed as $\mathbb F=\left(\mathbb F_1, ..., \mathbb F_N\right)$.
When we assume that the feature $\mathbb F_k$ (generated $k$-th type data in feature) inputted to the batch normalization layer follows a Gaussian distribution $\mathcal{N}\left(\mu_k, \sigma_k\right)$ and the sampling probability of $\mathbb F$ is $\mathbf{w};\|\mathbf{w}\|_{\ell1}=1$ and $\mathbf{w}'=\mathbf{w}+\mathbf{e}_w;\|\mathbf{w'}\|_{\ell1}=1$ in two different sets $\mathbb{S}_1$ and $\mathbb{S}_2$, respectively, the difference of the mixture distribution statistics between  $\mathbb{S}_1$ and $\mathbb{S}_2$ can be expressed as $\Delta_\mu=\mathbf{e}_w\bm{\mu}^T$ and $\Delta_\sigma=\mathbf{e}_w\mathbf{t}^T-2\left(\mathbf{w}\bm{\mu}^T\right)\left(\mathbf{e}_w\bm{\mu}^T\right)-\left(\mathbf{e}_w\bm{\mu}^T\right)^2$, where $\bm{\mu}=\left(\mu_1, ..., \mu_k\right)$ and $\mathbf{t}=\left(\mu_1^2+\sigma_1^2, ..., \mu_k^2+\sigma_k^2\right)$, $\Delta_\mu=0 \ and \ \Delta_\sigma=0$ $\iff$ $\mathbf{e}_w=\mathbf{0}$.}
\end{theorem}


To verify this, we train several individual models with different ratios of perturbation types for the estimation of the single BN layer, and we spot the obvious performance degeneration when the proportion of different perturbations types in the training set is different to that in the test set. This scenario would break the independently identically distribution assumption and make the single BN fail~\citep{Ioffe2015BN}. See Section \ref{sec:multi-abla} for experimental results.

Thus, instead of learning a mixture distributions with a single BN, \textbf{we utilize a multi-branch BN layer, in which each BN branch is in charge of one corresponding perturbation type (\ie, domain)}. Besides, the output of each perturbation is normalized by corresponding BN towards learning domain-invariant representation. One remaining problem is that the model does not know the input domain during inference. Existing work suggested that adversarial examples are separable from clean examples~\citep{Metzen2018On}, and clean/adversarial examples are drawn from different domains~\citep{Xie2020intriguing}. Therefore, \textbf{we introduce a gated sub-network to separate perturbations from different domains during inference, \ie, separates domain-specific (perturbation-specific) information.}

Thus, the multi-domain hypothesis motivates the design of GBN, which consists of a gated sub-network and a multi-branch BN layer. 




\begin{figure}[]
\vspace{-0.1in}
\begin{center}

\subfigure[\revise{Training of GBN}]{
\includegraphics[width=1.05\linewidth]{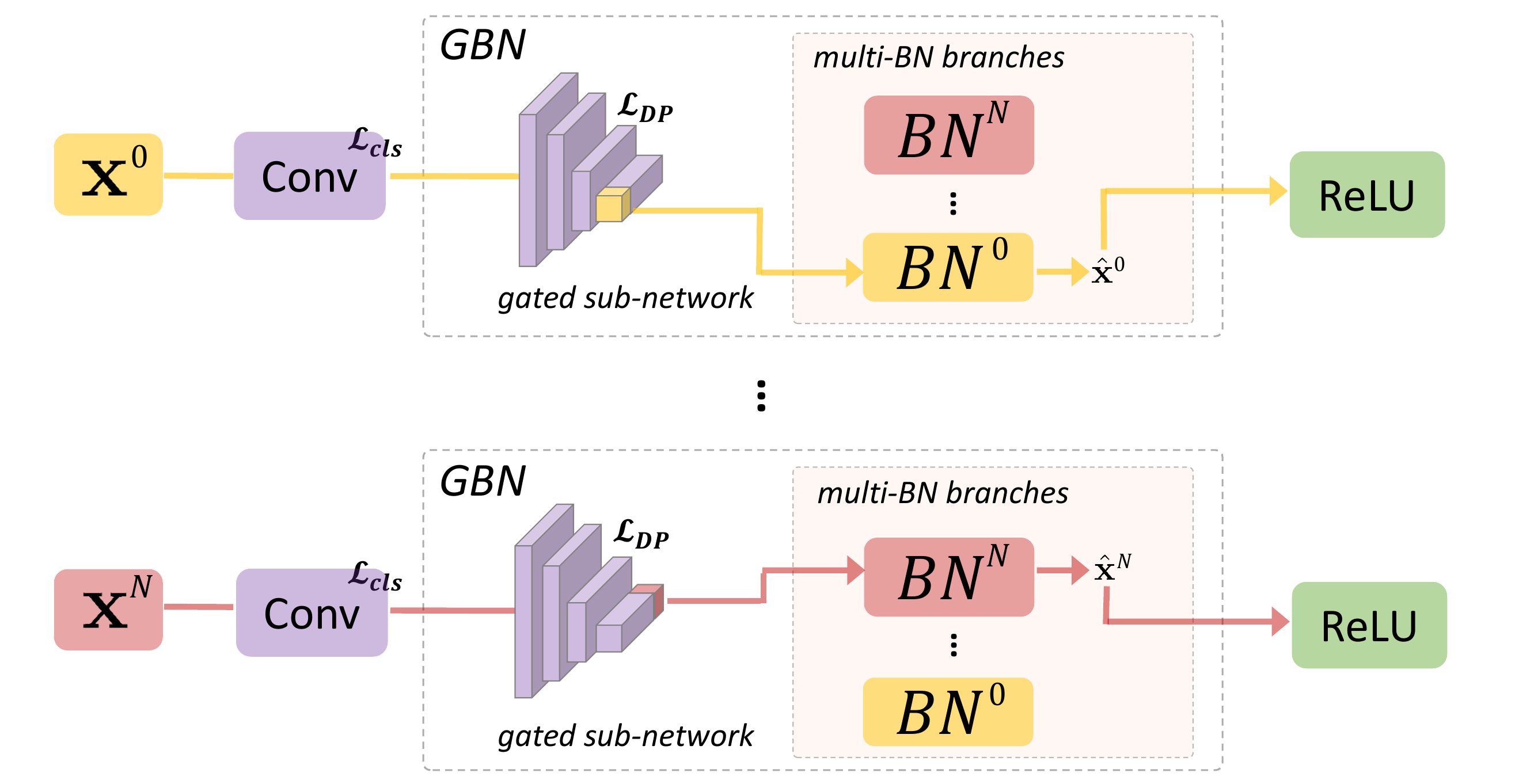}
\label{fig:gbn-training}
}
\subfigure[\revise{Inference of GBN}]{
\includegraphics[width=1.05\linewidth]{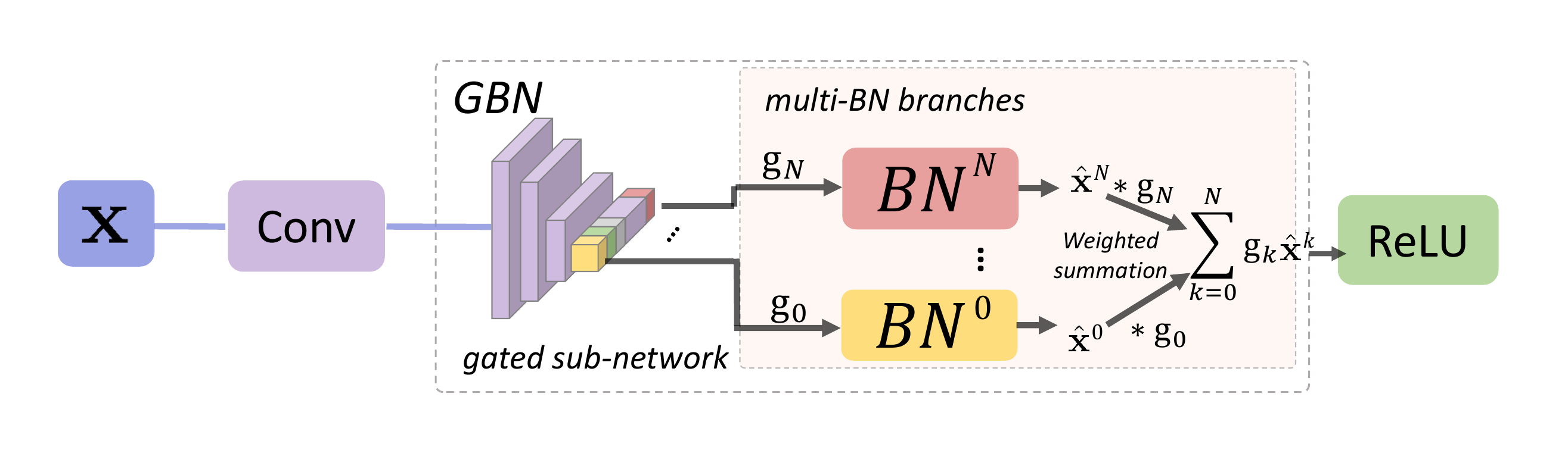}
\label{fig:gbn-inference}
}

\end{center}

\caption{The training and inference procedures of GBN. During training, samples from different domains (denoted by different colors) are fed into corresponding BN branches to update the running statistics. \revise{During inference, the calculation process works in a sequential manner. Given input data from an unknown domain, the gated sub-network first predicts the domain of input $\rvx$; GBN then lets $\rvx$ go through each BN branch $BN^k(\cdot)$ and obtain the normalized output $\hat{\rvx}^{k}$; finally, we jointly normalize the output by calculating a weighted summation as $\sum_{k=0}^{N}{\rvg_k \hat{\rvx}^{k}}$.}}
\label{fig:gbn-block}
\vspace{-0.1in}
\end{figure}



\subsection{Gated Batch Normalization architecture}

To train a perturbation-invariant predictor and obtain domain-invariant representations for multiple $\ell_p$ bounded adversarial perturbations, we use the BN layer statistics to align data from the mixture distribution. Specifically, GBN uses a BN layer with multiple branches $\Psi=\{BN^k(\cdot), k=0,..., N\}$ during training, where each branch $BN^k(\cdot)$ is exclusively in charge of the domain $\sD^k$. The aligned data are then aggregated as the input to subsequent layers to train a classifier, which achieves the minimum risk across different domains (\ie, the best robustness to different perturbations with a high accuracy on clean data).

To separate the input domain during inference and calculate the normalized output, GBN utilizes a gated sub-network $\Phi_{\theta}(\rvx)$ to predict the domain for each layer input $\rvx$, and we will illustrate how to train $\Phi_{\theta}(\rvx)$ in Section~\ref{sec:training}. Given the prediction of sub-network $\rvg=\Phi_{\theta}(\rvx)$ \revise{that outputs the confidence over $N+1$ domains}, we calculate the normalized output in a soft-gated way:
\begin{equation}
\begin{aligned}
\label{eqn:GBN}
\hat{\rvx} = GBN (\rvx)= \sum_{k=0}^{N}{\rvg_k \hat{\rvx}^{k}},
\end{aligned}
\end{equation}
where $\rvg_k$ represents the confidence of $\rvx$ belonging to the $k$-th domain; $\hat{\rvx}^{k}$ is the normalized output of $BN^{k}(\cdot)$, which uses the population statistics of the domain $\sD^k$ to perform the normalization. 

 We provide an overview of the inference procedure in Figure~\ref{fig:gbn-inference} and Algorithm~\ref{alg:infer_GBN}. In the \emph{supplementary material}, 
 we discussed an alternative approach that takes the top-1 prediction of $\rvg$ (the hard label), and show that our soft-label version achieves slightly better results than the hard-label one.
 
 Our GBN aims to disentangle domain-invariant features and domain-specific features: (1) the distributions of different domains are aligned by their normalized outputs (all are standardized distributions), which ensures that the following linear layers learn domain-invariant representations; and (2) the domain-specific features for each domain $\sD^k$ are obtained by the population statistics $\{\hat{\mu}^k, \hat{\sigma}^k\}$ of its corresponding BN branch  $BN^k(\cdot)$. 
 
\begin{algorithm}[thb]
   \caption{Inference of Gated Batch Normalization}
   \label{alg:infer_GBN}

\leftline{{\bf Input:} Layer input $\rvx$} 
{\bf Output:} Normalized output by GBN

  \begin{algorithmic}[1]
    \State Compute the output of gated sub-network $\rvg=\Phi_{\theta}(\rvx)$
   \For{$k$ in $N$+1 domains}
   \State Let $\rvx$ go through $BN^k(\cdot)$ and obtain the normalized output $\hat{\rvx}^{k}$ based on Eqn.~\ref{eqn:BN}.
   \EndFor
    \State Compute GBN output based on Eqn.~\ref{eqn:GBN}.
\end{algorithmic}
\end{algorithm}

\subsection{Training}
\label{sec:training}
We provide an overview of the training procedure in Figure~\ref{fig:gbn-training}, and the details are in Algorithm~\ref{alg:train_GBN}. Specifically, for each mini-batch $\mathcal{B}^0$ consisting of clean samples, we use PGD~\citep{madry2017towards} to generate batches of adversarial examples $\mathcal{B}^k$ ($k \in \{1,...,N\}$) for each domain $\sD^k$. To capture the domain-specific statistics of different perturbation types, given the mini-batch data $\mathcal{B}^{k}$ from domain $\sD^k$, we ensure that $\mathcal{B}^{k}$ goes through its corresponding BN branch $BN^k(\cdot)$, and we use Eqn.~\ref{eqn:BN} to compute the normalized output $\widehat{\mathcal{B}}^{k}$. The population statistics $\{\hat{\mu}^k, \hat{\sigma}^k\}$ of $BN^k(\cdot)$ are updated based on Eqn.~\ref{eqn:running-average}. In other words, we disentangle the mixture distribution for normalization and apply separate BN branches to different perturbation types for statistics estimation.

To train the gated sub-network $\rvg$, we provide the supervision of the input domain for each training sample. Specifically, we introduce the domain prediction loss $\mathcal{L}_{DP}$:

\begin{equation}
\begin{aligned}
\label{eqn:da_loss}
\mathcal{L}_{DP}= \sum_{k=0}^{N}\sum_{\revise{\rvx \sim \sD^{k}}} \ell(\Phi_{\theta}(\rvx),k).
\end{aligned}
\end{equation}

Finally, we optimize the parameters $\Theta$ of the entire neural network (\eg, the weight matrices of the convolutional layers, except for the gated sub-network) using the classification loss $\mathcal{L}_{cls}$:
\begin{equation}
\begin{aligned}
\label{eqn:cls_loss}
\mathcal{L}_{cls}= \sum_{k=0}^{N}\sum_{(\rvx,\rvy)\in \sD^{k}} \ell(f_{\Theta}(\rvx;BN^k(\cdot)),\rvy).
\end{aligned}
\end{equation}
\noindent \revise{where $BN^k(\cdot)$ is the corresponding BN branch.}

\begin{algorithm}[thb]

   \caption{Training of Gated Batch Normalization (GBN) for Each Iteration.}
   \label{alg:train_GBN}
\leftline{{\bf Input:} Network $f$ with GBN} 
\leftline{{\bf Output:} Model parameters $\Theta$ and $\theta$} 

\begin{algorithmic}[1]
   \State Given the mini-batch data $\mathcal{B}^0$, use PGD algorithm to generate batches of adversarial examples $\mathcal{B}^k$ ($k \in \{1,...,N\}$) of different perturbation types
   \For{$k$ in $N$+1 domains}
   \State Let $\mathcal{B}^{k}$ go through $BN^k(\cdot)$ at each layer to obtain normalized outputs $\widehat{\mathcal{B}}^{k}$, using Eqn.~\ref{eqn:BN}.
   \State Update the population statistics $\{\hat{\mu}^k, \hat{\sigma}^k\}$ of $BN^k(\cdot)$ at each layer, using Eqn.~\ref{eqn:running-average}.
   \EndFor
    \State Update $\theta$ (parameters of the gated sub-network at each layer) based on Eqn.~\ref{eqn:da_loss}.
    \State Update $\Theta$ (parameters of the whole network) based on Eqn.~\ref{eqn:cls_loss}.
\end{algorithmic}
\end{algorithm}

\subsection{Discussion over other normalization techniques}
\label{sec:apprach_discussion}
Since our GBN block is based on the batch normalization technique, we further compare the differences between our GBN and other related normalization techniques including Mode Normalization (MN)~\citep{deecke2018mode} and MBN~\citep{Xie2020intriguing,xie2020adversarial}. 

\revise{Our GBN differs from them to a large extent in the following aspects: (1) Goal. The goal of MBN is to improve image recognition or analyze the properties of BN for adversarial training, and MN aims to tackle the problem of multi-task/multi-distribution learning. In contrast, our GBN is the first to defend against adversarial attacks using multiple BNs on adversarially trained models and further generalize to multiple perturbations. (2) Motivation and implementation. For MN, though also containing gated networks, it only adopts two BN branches (without direct correspondence to specific domains) regardless of the domain numbers. In other words, MN aims to learn a mixture distribution for several domains and normalizes the input with different weights from the two mixed distributions. In contrast, our GBN separates domain-specific statistics for different domains (each BN captures the statistics for each domain) and then aligns data to build domain-invariant representations. For MBN, it applies two BNs to clean and adversarial examples for statistics estimation, and manually selects the BN branch for each input image. This is impractical since it requires prior knowledge of the input domain during inference. However, our GBN can predict the domain of input and route the corresponding BN branches during inference. (3) Performance. The experimental results in Table \ref{tab:adv_result} and Section \ref{sec:ablation_study} verify the effectiveness of our motivation (+20\% absolute robustness improvement). In addition, these results indicate that simply normalizing inputs with mixture distribution statistics (MN) or using fixed unrelated statistics (MBN) is less effective for defending multiple types of perturbations.}





%% file: 4_experiments.tex
\section{Experiments}
We evaluate the effectiveness of our GBN block to simultaneously defend against $\ell_1$, $\ell_2$, and $\ell_{\infty}$ perturbations, which are the most representative and commonly used adversarial perturbation types. 

\subsection{Experimental setup}

{\textbf{Datasets.}} In this paper, we conduct experiments on image classification benchmarks, including MNIST~\citep{lecun1998mnist}, CIFAR-10~\citep{krizhevsky2009learning}, and Tiny-ImageNet~\citep{wu2017tiny}. MNIST is a dataset containing 10 classes of handwritten digits of size 28 $\times$ 28; CIFAR-10 consists of 60,000 natural scene color images in 10 classes of size 32 $\times$ 32; Tiny-Imagenet has 200 classes of size 64 $\times$ 64. {For our experiments, we use the given training set in each datasets to train our models, and divide the given test set in each datasets into validation set and test set (the ratio between validation set and test set is 2:8). In each training epoch we use the validation set to validate the performance of the trained models. After finishing training, we use test set to evaluate the final performance of the models. }

{\textbf{Architectures and hyperparameters.}} We use LeNet~\citep{lecun1998gradient} for MNIST; ResNet-20~\citep{he2016deep}, VGG-16~\citep{Simonyan2015Very}, and WRN-28-10~\citep{Zagoruyko2016WRN} for CIFAR-10; and ResNet-34~\citep{he2016deep} for Tiny-ImageNet. For fair comparisons, we keep the architecture and main hyper-parameters the same for GBN and other baselines. Specifically, all the models for each method on MNIST, CIFAR-10, and Tiny-ImageNet are trained for 40, 40, and 20 epochs, respectively. We set the mini-batch size=64, use the SGD optimizer with weight decay 0.0005 for Tiny-ImageNet and use no weight decay for MNIST and CIFAR-10. We set the learning rate as 0.1 for MNIST and 0.01 for CIFAR-10 and Tiny-ImageNet. Unless otherwise specified, we add GBN in all layers.


{\textbf{Adversarial attacks.}}\label{sec:attack} {We follow existing guidelines~\citep{Schott2019Towards,Tramer2019Adversarial,Maini2020adversarial} where we incorporate multiple adversarial attacks with different perturbation types and adopt their commonly-used settings for attacks.} For MNIST, the magnitude of perturbation for $\ell_1$, $\ell_2$, and $\ell_{\infty}$ is $\epsilon=$ 10, 2, 0.3. For CIFAR-10 and Tiny-ImageNet, the perturbation magnitude for $\ell_1$, $\ell_2$, and $\ell_{\infty}$ is $\epsilon=$ 12, 0.5, 0.03. For $\ell_1$ attacks, we adopt PGD~\citep{madry2017towards}, and BBA~\citep{NIPS2019_9446}; for $\ell_2$ attacks, we use PGD, C\&W~\citep{carlini2017towards}, Gaussian noise~\citep{Rauber2017Foolbox}, and boundary attack (BA)~\citep{Brendel2018Decision}; For $\ell_{\infty}$ attacks, we use PGD, FGSM~\citep{goodfellow6572explaining}, SPSA~\citep{Jonathan2018SPSA}, Nattack~\citep{li2019nattack}, Momentum Iterative Method (MI-FGSM)~\citep{dong2017boosting}, C\&W, and AutoAttack~\citep{croce2020autoattack}. To demonstrate that our GBN does not introduce obfuscated gradients, we use white-box and black-box adversarial examples, generated with both gradient-based and gradient-free algorithms.

{\textbf{Adversarial defenses.}} We compare with existing defenses against the union of $\ell_1$, $\ell_2$, $\ell_{\infty}$ adversaries. We compare with ABS~\citep{Schott2019Towards}\footnote{\revise{ABS} considers the $\ell_0$ perturbations, which are subsumed within the $\ell_1$ ball of the same radius. Meanwhile, ABS is only designed for MNIST. Due to the limited code, we use the results reported in~\citep{Schott2019Towards}.}, MAX, AVG~\citep{Tramer2019Adversarial}, and MSD~\citep{Maini2020adversarial}. For completeness, we also compare with TRADES~\citep{zhang2019theoretically}, PAT~\citep{Laidlaw2021perceptual}, and PGD adversarial training~\citep{madry2017towards} with a single perturbation type (\ie, the model is trained on $\ell_1$, $\ell_2$, or $\ell_{\infty}$, denoted as $P_1$, $P_2$, and $P_{\infty}$ respectively). 

{\textbf{Normalization techniques.}} We further compare our methods to existing normalization techniques, including MN and MBN. MN extends BN to more than a single mean and variance, detects the modes of the data, and then normalizes samples that share common features. MBN includes two BN branches for clean and adversarial examples respectively. Since manually selecting the BN branch is infeasible in the adversarial setting, we add a 2-way gated sub-network to each MBN block. \revise{We also evaluate the original MBN that only uses the clean BN during inference, which shows weak robustness (see Section \ref{sec:ablation_study}).}

\begin{table*}[]

\caption{Accuracies on different datasets (\%). \revise{We emphasize both the best and second-best values for each type of attack for better visualization.} We also provide the Standard Deviation for \emph{All attacks} of each method.}
\label{tab:adv_result}
\vspace{-0.1in}
\begin{center}
\small
\subtable[LeNet on MNIST]{
\label{tab:mnist_sum}
\setlength{\tabcolsep}{0.6mm}
\begin{tabular}{cc|ccccccccc|cc|c}
\toprule
& Vanilla & {PAT}& {TRADES} & $P_1$ & $P_2$ & $P_{\infty}$ & {AVG} & {MAX} & {ABS} & {MSD} & {MN} & {MBN}& {\textbf{GBN(ours)}}\\
\hline
$\ell_1$ attacks&  5.1& 50.3 & 9.6 & 32.6 & 67.3  & 7.1 &30.2 & 26.6 & /& 41.2& 23.1 & \textbf{\revise{60.5}}& \textbf{72.8}\\

$\ell_2$ attacks& 1.2&  54.2 & 11.5 & 27.1  & 65.8  & 18.2  &39.6 & 37.7 & \textbf{\revise{82.3}} & 71.1 & 18.7& 65.0& \textbf{92.3}\\

$\ell_{\infty}$ attacks&  0.0&  1.1 & \textbf{\revise{76.5}} & 0.3  & 0.1  & \textbf{78.6}  & 38.9 & 52.2& 16.7& 38.4&0.0 &18.7& 68.5\\

\multirow{2}*{All attacks} &0.0&  1.0 &9.2 & 0.1  & 0.1  & 6.9  & 29.4 & 26.3 & 16.3& \textbf{\revise{38.1}} & 0.0 & 18.4& \textbf{68.5}\\

& $\pm{}$0.04 & $\pm{}$0.06  & $\pm{}$0.11 & $\pm{}$0.16 & $\pm{}$0.17 & $\pm{}$0.09 & $\pm{}$0.24 & $\pm{}$0.22 & $\pm{}$0.00 & $\pm{}$0.25 & $\pm{}$0.03 & $\pm{}$0.34 & $\pm{}$0.31\\

Clean &\textbf{99.1}&  91.0 & 98.9 & 98.5 & 98.7 & 98.3 & 98.6& 98.2& 99.0&97.1 & 98.0 & 98.4& 98.4 \\
\bottomrule
\end{tabular}
}

\subtable[ResNet-20 on CIFAR-10]{
\label{tab:cifar_sum}
\setlength{\tabcolsep}{1.0mm}
\begin{tabular}{cc|cccccccc|cc|c}
\toprule
& Vanilla & PAT & TRADES & $P_1$ & $P_2$ & $P_{\infty}$ &AVG & MAX & MSD &MN& MBN& \textbf{GBN(ours)}\\
\hline

$\ell_1$ attacks & 0.0 & 33.7& 15.7 & 19.1  & 31.9  & 16.3  & \textbf{\revise{44.4}} & 41.1 & 42.5& 39.1 & 44.2& \textbf{56.9}\\

$\ell_2$ attacks &  0.0& 47.7& \textbf{\revise{59.1}} & 54.7  & 55.8  & 50.1  & 55.7 & 51.6 & 58.2& 29.5 & 20.0& \textbf{68.5}\\

$\ell_{\infty}$ attacks &  0.0& 24.8& 44.6 & 11.7  & 23.5  & 37.6  & 30.2 & 33.7 & 37.6& 12.7 & \textbf{\revise{39.7}}& \textbf{49.2}\\

\multirow{2}*{All attacks} &0.0& 24.5& 15.4 & 11.5  & 23.2  & 16.1  &30.2 & 33.4 & \textbf{\revise{36.9}} & 12.3 & 20.0& \textbf{48.2}\\

& $\pm{}$0.03 & $\pm{}$0.36 & $\pm{}$0.35 & $\pm{}$0.05 & $\pm{}$0.41 & $\pm{}$0.36 & $\pm{}$0.50 & $\pm{}$0.42 & $\pm{}$0.56 & $\pm{}$0.51 & $\pm{}$0.61 & $\pm{}$0.64\\

Clean &\textbf{89.4}& 62.2& 86.2 & 83.7 & 87.2 & 83.6 & 80.0& 76.7& 78.4 & 82.0 & 79.1&80.2 \\
\bottomrule
\end{tabular}
}

\subtable[ResNet-34 on Tiny-ImageNet]{
\label{tab:imagenet_sum}
\setlength{\tabcolsep}{1.0mm}
\begin{tabular}{cc|cccccccc|cc|cc}
\toprule
& Vanilla & PAT & TRADES & $P_1$ & $P_2$ & $P_{\infty}$ & AVG & MAX & MSD & MN & MBN & \textbf{GBN(ours)}\\
\hline

$\ell_1$ attacks &  5.6 &13.1 & 25.2 & 19.5 & 26.1 & 19.7 & 30.7 & 22.7 & 7.1& 7.9 & \textbf{\revise{36.1}}& \textbf{43.7}\\

$\ell_2$ attacks &  10.1 &12.7 & 31.2 & 29.6 & \textbf{\revise{32.5}} & 27.9 & 32.2 & 27.5 & 11.0& 17.4 & 30.4&\textbf{41.6}\\

$\ell_{\infty}$ attacks &  0.0 &2.2 & 9.3 & 0.3 & 0.5 & 9.0 & 6.5 & 8.1 & 5.0& 6.4 & \textbf{\revise{19.3}}& \textbf{38.2}\\

\multirow{2}*{All attacks} &0.0 &1.9 & 9.0 & 0.0 & 	0.5 & 	8.8 & 	6.2  &	7.4 & 4.6 & 6.5 & \textbf{\revise{18.3}}& \textbf{37.7}\\

& $\pm{}$0.05 & $\pm{}$0.35   & $\pm{}$0.44 & $\pm{}$0.21 & $\pm{}$0.37 & $\pm{}$0.52 & $\pm{}$0.56 & $\pm{}$0.58 & $\pm{}$0.67 & $\pm{}$0.62 & $\pm{}$0.52 & $\pm{}$0.63\\

Clean &\textbf{54.0} &20.2 & 43.0 & 52.3& 54.1& 44.1& 41.2& 35.9 &28.5 & 46.3 & 45.5& 43.2 \\
\bottomrule
\end{tabular}
}
\end{center}

\end{table*}

{\textbf{{Evaluation metrics.}}} {To evaluate model robustness against attacks, we report the \emph{\textbf{top-1 classification accuracy}} for each adversarial attack method and clean examples (the higher the better). Since we use several adversarial attack methods, for each perturbation type (\eg, $\ell_1$, $\ell_2$, $\ell_{\infty}$), we measure the \emph{\textbf{worst-case}} top-1 classification accuracy (the higher the better), where we report the worst result among various attacks within the perturbation type. In other words, given an input, if any attack within specific norm (\eg, $\ell_1$) generates an adversarial example that leads to the wrong prediction, we mark it as a failure. Moreover, we also report \emph{\textbf{all attacks}}, which represents the worst-case performance over all attacks of different perturbations for each input~\citep{Maini2020adversarial}. Similarly, given an image, if any attack generates an adversarial example that leads to the wrong prediction, we mark it as a failure.}

\begin{figure}[]
\centering
\vspace{-0.1in}
\subfigure[Conv gate]{
\includegraphics[width=0.35\linewidth]{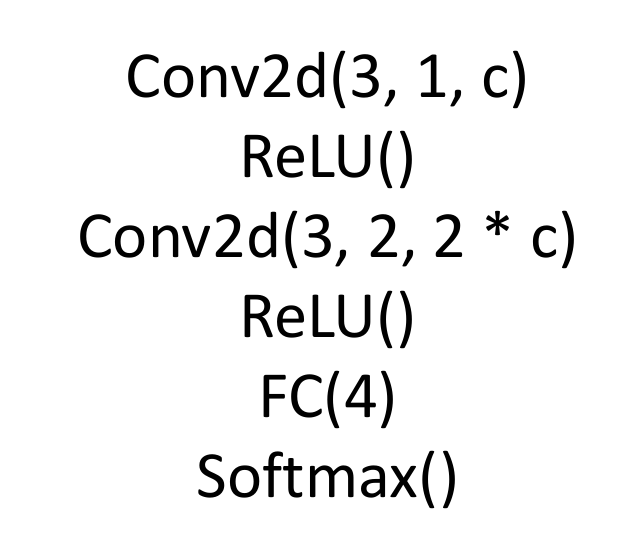}
}
\subfigure[FC gate]{
\includegraphics[width=0.35\linewidth]{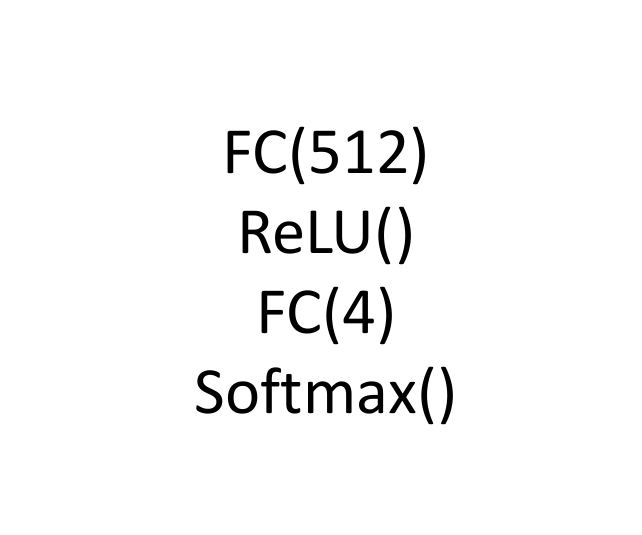}
}
\caption{\revise{The architecture of the gated sub-network.}}
\label{fig:gate_net}
\end{figure}

\revise{\textbf{GBN architectures.}
We here illustrate the architecture of the gated sub-network $\rvg$ in GBN. As shown in Figure \ref{fig:gate_net}, we devise two types of gated sub-network, namely \emph{Conv gate} and \emph{FC gate}.} \revise{Here, the notation Conv2d($d, s, k$) refers to a convolutional layer with $k$ filters whose size are $d \times d$ convolved with stride $s$; ReLU($\cdot$) denotes the rectified linear unit used as the activation function in the network; and FC($m$) denotes a fully connected layer with output size $m$. Specifically, \emph{Conv gate} is a convolutional neural network consisting of two convolutional layers, two ReLU layers, and one fully-connected layer; \emph{FC gate} is a simple fully-connected neural network that includes two fully-connected layers and one ReLU layer.}

\textbf{Data availability statement.} All our experiments are conducted on commonly adopted open datasets MNIST, CIFAR-10, and Tiny-ImageNet. For the implementation of defenses, we use the published codes for ABS\footnote{\url{https://github.com/bethgelab/AnalysisBySynthesis}}, MAX/AVG\footnote{\url{https://github.com/ftramer/MultiRobustness}}, MSD\footnote{\url{https://github:com/locuslab/robust_union}}, TRADES\footnote{\url{https://github.com/yaodongyu/TRADES}}, and PAT\footnote{\url{https://github.com/cassidylaidlaw/perceptual-advex}}. For attacks, we use the original codes\footnote{\url{https://github.com/fra31/auto-attack}} for AutoAttack, ART\footnote{\url{https://github.com/Trusted-AI/adversarial-robustness-toolbox}} for C\&W-$\ell_{\infty}$, and ARES\footnote{\url{https://github.com/thu-ml/ares}} for SPSA and NATTACK; for other attacks we adopt FoolBox~\citep{Rauber2017Foolbox}. The code of our GBN can be found at \url{https://github.com/liuaishan/GBN}.

\emph{We defer more detailed implementation of GBN and hyper-parameters of our experiments to the supplementary material.}

\subsection{Comparison with other baselines}
\label{sec:main_exp}

In this part, we first evaluate model robustness against multiple perturbation types (\ie, $\ell_1$, $\ell_2$, and $\ell_{\infty}$) in both the white-box and black-box settings. For all attacks except FGSM and APGD$_{CE}$ (we use the default setting in AutoAttack), we ran 5 random restarts for each input. For GBN, we trained each model 3 times, and the results are similar. 

The results on MNIST using LeNet, CIFAR-10 using ResNet-20, and Tiny-ImageNet using ResNet-34 are shown in Table~\ref{tab:adv_result}. We also report the breakdown results for each individual attacks in the supplementary material. \emph{Please refer to the supplementary materials for more experimental results on different datasets with different models.} Note that the results in Table~\ref{tab:adv_result} of each perturbation type indicate the sample-wise worst-case robustness, which would be slightly less than the worst robustness accuracy of these individual attacks in the breakdown results due to the computation strategy of the evaluation metrics. From these results, we can draw the following observations.

(1) For defending multiple perturbations (\ie, $\ell_1$, $\ell_2$, $\ell_{\infty}$) in both white-box and black-box scenarios, our GBN consistently outperforms others by large margins over 10\% \revise{absolute improvement} on all 3 datasets. This demonstrates the superiority of our GBN approach to improving model robustness against multiple perturbations.

(2) Due to the trade-off between adversarial robustness and standard accuracy~\citep{tsipras2018robustness}, our clean accuracy is lower than the vanilla model. However, our GBN maintains a comparatively high clean accuracy compared to other adversarial defense strategies. 

(3) GBN is easy to train on different datasets, while other defenses for multiple perturbation types (\eg, MSD) are hard to converge on large datasets such as Tiny-ImageNet. This further indicates the training advantage of our GBN.

(4) Based on the results of black-box and gradient-free adversarial attacks in our supplementary material (\ie, BA, SPSA, NATTACK), GBN outperforms the others by large margins, confirming that no gradient masking~\citep{athalye2018obfuscated} has been introduced.

(5) In contrast to other normalization methods, as shown in Table \ref{tab:adv_result}, GBN outperforms MN and MBN by large margins (up to 60\% \revise{absolute improvement}). We conjecture that the key factor behind is that MN and MBN do not align the feature representations for different data distributions (different $\ell_p$-norm perturbations). Specifically, MN aims to preserve the diversity among different distributions, by learning the mixture of Gaussian; as for the modified MBN, the data from different perturbation types are mixed in the same BN. For our GBN, it first aligns the distribution among different perturbations by using one BN branch to handle one perturbation type (learn domain-specific features), and then the aligned distribution contributes to the subsequent modules or layers (\eg, convolutional layer) for learning domain-invariant representations. More discussions can be found in Section \ref{sec:t-sne}.


\subsection{Adaptive white-box attacks for GBN} 
\label{sec:adaptive-attack}

Besides the commonly-adopted adversarial attacks, we here examine the performance of GBN against white-box attacks specially designed for GBN, aiming to provide a more rigorous analysis. In the following {cases, the adversary knows every detail of our GBN and could directly attack the GBN block.}

In particular, we conduct the following experiments: (1) attacking the gated sub-networks, \ie, generating adversarial examples to fool the classification of the gated sub-networks; (2) manually selecting the BN branch to generate adversarial examples; (3) generate adversarial attacks by optimizing jointly the model final prediction and the gated sub-network; \revise{and (4) generate adversarial attacks by randomly attacking with $\ell_1$, $\ell_2$, and $\ell_{\infty}$-norm perturbations.}

\textbf{First type of adaptive attack}. Given clean, $\ell_1$, $\ell_2$, and $\ell_{\infty}$ adversarial examples, we then perturb these samples to fool the gated sub-networks at all layers of ResNet-20 on CIFAR-10 using PGD attacks. Our whole model achieves 73.2\%, 71.4\%, 72.4\%, and 71.8\% classification accuracy for fooling the gated sub-networks to wrong predictions towards clean, $\ell_1$, $\ell_2$, and $\ell_{\infty}$ domains.

\textbf{Second type of adaptive attack}. Here, the adversary generates adversarial examples by manually selecting the BN branch of the GBN module to perform PGD attacks. In Table \ref{tab:var_attack}, GBN remains the same level of robustness on both MNIST and CIFAR-10 dataset compared to our main white-box attacks.

\textbf{{Third type of adaptive attack}}. {Here, we choose ResNet-20 and the CIFAR-10 dataset, where we generate adversarial attacks by optimizing jointly the model final prediction and the gated sub-network. In other words, we attack the final outputs of the model to generate $\ell_1$, $\ell_2$, and $\ell_{\infty}$ adversarial examples; meanwhile we also force the outputs of all sub-networks to wrong branch predictions, which is inconsistent with the branch of the adversarial examples. As shown in Table \ref{tab:third-attack}, the third type of adaptive attack achieves the strongest attacking ability among the three attacks and could attack our GBN to some extent. However, the robust accuracy of GBN under this type of attack does not decrease too much.} 

\revise{\textbf{Fourth type of adaptive attack}. Given clean examples, we set the overall PGD attack step number as 100. At each iteration step, we generate attacks by randomly choosing $\ell_1$, $\ell_2$, and $\ell_{\infty}$ bounded perturbations and perform clip operation based on the average constraint among the three types. The ResNet-20 model with GBN on CIFAR-10 achieves 57.1\% accuracy in this scenario.}

\begin{table}[]

\caption{{Second type of adaptive attack.} White-box attacks on MNIST with LeNet and CIFAR-10 with ResNet-20 by manually selecting BN branches to generate adversarial examples (BN$^0$, BN$^1$, BN$^2$, and BN$^3$ denotes the BN branch for clean, $\ell_1$, $\ell_2$, and $\ell_{\infty}$ adversarial examples in GBN). Results are shown in model accuracy (\%).}

\label{tab:var_attack}
\begin{center}
\footnotesize

\label{tab:var_attack_mnist}
\setlength{\tabcolsep}{1.0mm}
\begin{tabular}{c|cccc|cccc}

\toprule
& \multicolumn{4}{c|}{MNIST} & \multicolumn{4}{c}{CIFAR-10}\\
&  BN$^0$ & BN$^1$ & BN$^2$ & BN$^3$ &  BN$^0$ & BN$^1$ & BN$^2$ & BN$^3$ \\
\hline
PGD-$\ell_1$&  86.7 & 86.3 & 86.5 & 96.7 &  58.2 & 58.0 & 58.7 & 59.7 \\
PGD-$\ell_2$&  97.6& 98.0 & 97.2 & 97.8 &  69.2& 69.3 & 68.8 & 69.2\\
PGD-$\ell_{\infty}$ &  96.8 &97.8 &96.2 & 95.8 &  60.1 &60.7 &60.7 & 59.5\\

Clean accuracy & 98.4 & 98.4 & 98.4 & 98.4 & 80.2 & 80.2 & 80.2 & 80.2  \\
\bottomrule
\end{tabular}
\end{center}
\end{table}

\begin{table}[]
\caption{{Third type of adaptive attack. (a) shows the classification accuracy (\%) of ResNet-20 on CIFAR-10 against attacks that jointly fool model prediction and gated sub-network. \revise{(b) shows the results of standard white-box PGD attacks.}}}
\label{tab:third-attack}
\begin{center}
\footnotesize

\subtable[Third type of adaptive attack using LeNet on MNIST]{
\setlength{\tabcolsep}{1.0mm}
\begin{tabular}{ccccc}
\toprule
 & {Clean} & {Adaptive PGD-$\ell_1$} & {Adaptive PGD-$\ell_2$} & {Adaptive PGD-$\ell_{\infty}$} \\
\hline \\
\textbf{{GBN}} & {80.2} & {56.5} & {50.6} & {55.4} \\
\bottomrule
\end{tabular}
}

\subtable[\revise{Standard white-box attack on MNIST and CIFAR-10}]{
\setlength{\tabcolsep}{0.9mm}
\begin{tabular}{cccc|ccc}
\toprule

& \multicolumn{3}{c|}{\revise{MNIST}} & \multicolumn{3}{c}{\revise{CIFAR-10}}\\
&  \revise{PGD-$\ell_1$} & \revise{PGD-$\ell_2$} & \revise{PGD-$\ell_{\infty}$} & \revise{PGD-$\ell_1$} & \revise{PGD-$\ell_2$} & \revise{PGD-$\ell_{\infty}$} \\
\hline
\\
\textbf{\revise{GBN}} & \revise{86.1} & \revise{97.4} & \revise{95.8} & \revise{58.1} & \revise{68.9} & \revise{58.0} \\
\bottomrule
\end{tabular}
}
\end{center}
\end{table}

\begin{figure}[]
\vspace{-0.1in}
\begin{center}

\subfigure[attack gated sub-net]{
\includegraphics[width=0.47\linewidth]{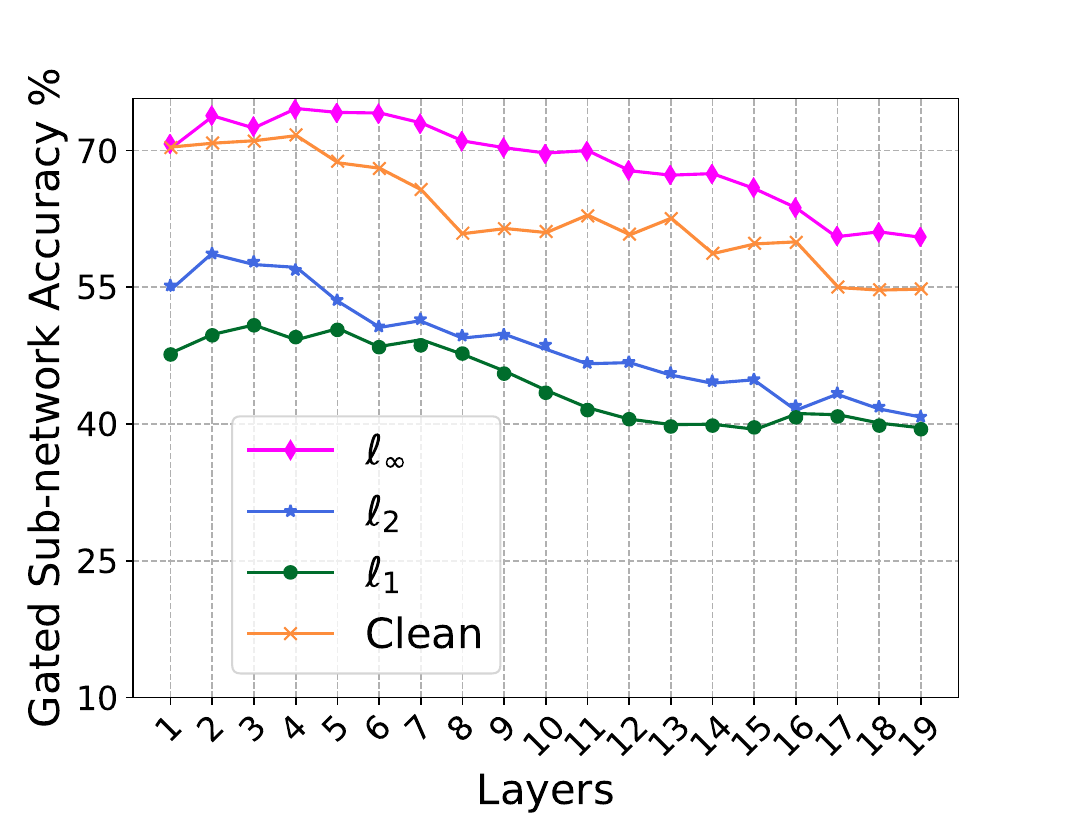}
\label{fig:type1_attack}
}
\subfigure[attack whole network]{
\includegraphics[width=0.47\linewidth]{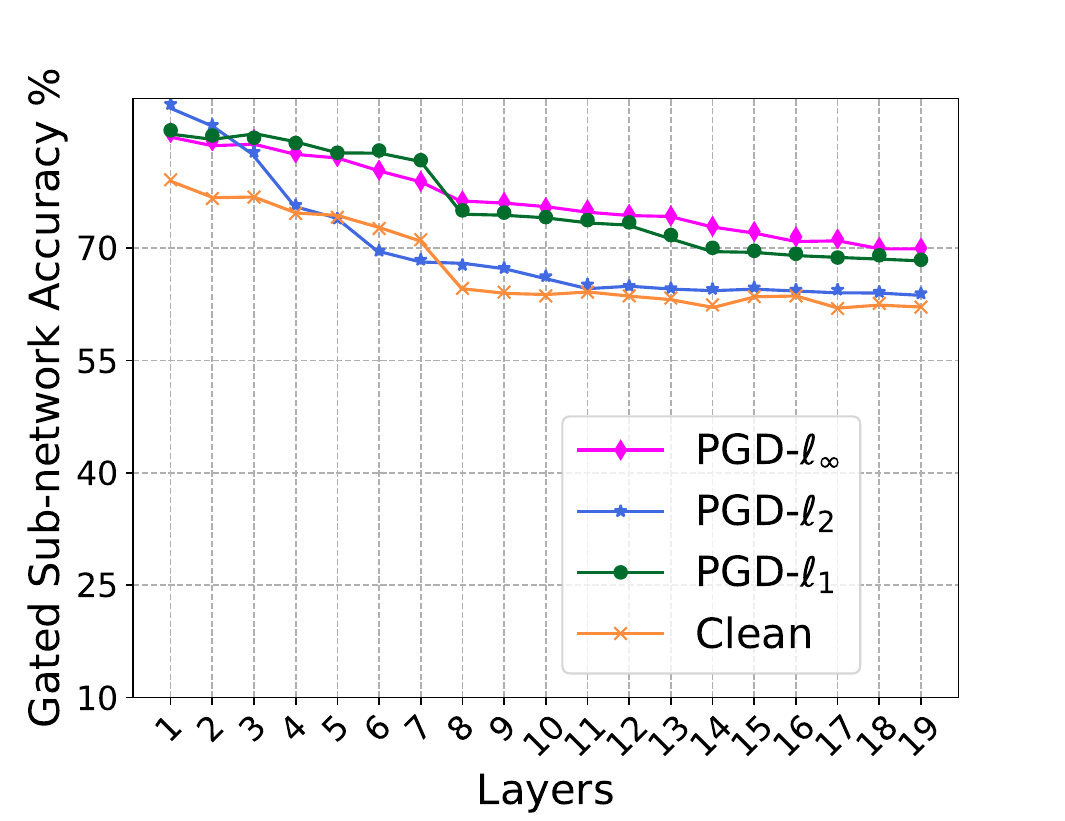}
\label{fig:gate_acc}
}

\end{center}
\caption{Gated sub-network prediction accuracy in different scenarios (ResNet-20 on CIFAR-10). \revise{The x-axis denotes the serial number of layers in the ResNet-20 model, and each point in the figure represents the prediction value of the gated sub-network at specific layer of ResNet-20.} (a) shows the results of fooling gated sub-networks in all GBN layers ({the first type of adaptive attack}); (b) shows the results of attacking whole network (common white-box attacks for the model).}

\end{figure}

\textbf{{Discussions.}} {Based on the above results, we can draw several conclusions as follows.}

{(1) Our GBN does not rely on obfuscated gradient to provide robustness to models~\citep{athalye2018obfuscated}. Obfuscated gradient or gradient masking has been demonstrated to cause a false sense of security, which often shows high accuracy on standard white-box attacks while low performance on adaptive attacks. The robustness of GBN decreases on adaptive attacks but still remains comparatively stable, which specifies that our GBN does not introduce obfuscated gradient.}

{(2) Compared to attacking the full model, as done in Table \ref{tab:adv_result}, the model accuracy towards the first and second adaptive attacks are at the same level or comparatively higher. We conjecture that the reasons might be listed as follows. For the \emph{first adaptive attack}, different GBN layers use different features to predict the domains, thus it is difficult to simultaneously fool all GBN layers under the magnitude constraint (see Figure~\ref{fig:type1_attack} for gated sub-networks prediction accuracy using the first type of attack). \revise{For example, the prediction accuracies of the gated sub-networks at different layers (each point in  Figure~\ref{fig:type1_attack}) are almost all higher than 40\% after being attacked.} Even if being able to fool all the gated sub-networks and force the adversarial examples to wrong BN branches during inference, the attack is similar to generating and feeding white-box adversarial examples for adversarially-trained models with single BN (such as AVG), since the adversarial examples are all normalized by the ``wrong'' BN branch for AVG. According to the results in the supplementary materials, the results of the first adaptive attack on GBN and common white-box PGD attacks on AVG are also similar. \emph{As for the second type of adaptive attack}, we conjecture that the BN branches during adversarial examples generation might not be critical for the adversarial attacking ability. The common white-box attacks (our main experiments) generate adversarial examples based on gradients via test mode BN, which would go through the corresponding BN branches that are most likely to maximize the model loss. Thus, the second adaptive attack shows relatively lower attacking performance than common white-box attacks.}

\revise{(3) The third type of attack achieves the best-attacking performance and outperforms most of the common white-box attacks. Though showing weaker performance than the third one, the fourth type of adaptive attack achieves comparatively high attacking performance. Similarly, the fourth type of adaptive attack aims to mislead the overall model output, meanwhile, it also achieves the effect to fool the gated sub-network since the combination of perturbation types might be hard for GBN to classify. Therefore, it suggests that the weak point of our GBN might be attacked and exploited by adversaries that simultaneously fool the GBN prediction and model prediction.}

\subsection{Ablation study}
\label{sec:ablation_study}
In this section, we provide some ablation studies to further investigate our GBN approach.

\textbf{{Different attack steps and budgets}.} {In this part, we evaluate and show the behavior of our GBN towards adversarial attacks with different iteration steps and perturbation budgets.}

Firstly, we provide the experimental results of PGD adversarial attacks using different iteration step numbers (\ie, 50, 100, 200, 1000) in terms of $\ell_{\infty}$ norm on CIFAR-10. For PGD attacks, we set the perturbation budgets $\epsilon$=0.03, step size $\alpha$=$\epsilon$/10, and different iteration steps $k$. In general, with the increasing of the iteration steps, the attacks become stronger. As shown in Table~\ref{tab:pgd-k}, our GBN still outperforms other methods on PGD attacks using different iteration steps and remains a high level of robustness.

{Moreover, we conduct experiments using PGD attacks of different perturbation budgets on CIFAR-10. Specifically, we choose ResNet-20 and use white-box PGD-$\ell_{1}$, PGD-$\ell_{2}$, and PGD-$\ell_{\infty}$ attacks with 7 different perturbation budgets. The specific budgets and results are shown in Table \ref{tab:magnitude}. From the results we can observe that the robustness of GBN decreases with the increasing of perturbation budges but it could still keep robustness to a certain extent. }

\begin{table}[]
\caption{Model accuracy (\%) of ResNet-20 on CIFAR-10 on different PGD-$k$ attacks ($k$ is the iteration steps).}
\label{tab:pgd-k}
\begin{center}
\footnotesize
\setlength{\tabcolsep}{0.7mm}
\begin{tabular}{cc|cccc|cccc}

\toprule
& Vanilla & TRADES & AVG & MAX & MSD & MN & MBN & \textbf{GBN(ours)} \\
\hline \\
PGD-50 & 2.3 & 49.5 & 37.1 & 40.4 & 44.1 & 34.1 & 58.4 & \textbf{59.5} \\
PGD-100 & 0.0 & 49.4 & 36.5 & 39.8 & 43.5 & 33.8 & 57.7 & \textbf{59.2} \\
PGD-200 & 0.0 & 48.8 & 36.1 & 38.9 & 43.0 & 33.2 & 57.1 & \textbf{58.6} \\
PGD-1000 & 0.0 & 47.5 & 35.2 & 38.6 & 42.2 & 32.1 & 54.8 & \textbf{58.0} \\
\bottomrule

\end{tabular}
\end{center}
\end{table}

\begin{table}[]

\caption{{Accuracy (\%) of GBN under white-box PGD-$\ell_{1}$, PGD-$\ell_{2}$, and PGD-$\ell_{\infty}$ attacks with different perturbation budgets.}}
\label{tab:magnitude}
\small
\subtable[{PGD-$\ell_{1}$}]{
\resizebox{\linewidth}{!}{
\begin{tabular}{c|ccccccc}
\toprule
 {$\varepsilon$}              & {3}     & {6}     & {12}   & {24}    & {36}    & {48}    & {60}    \\ \midrule
{PGD-$\ell_{1}$} & {74.7} & {66.1} & {58.1} & {46.5} & {37.5} & {25.5} & {16.5} \\ \bottomrule
\end{tabular}}}

\subtable[{PGD-$\ell_{2}$}]{
\resizebox{\linewidth}{!}{
\begin{tabular}{c|ccccccc}
\toprule
 {$\varepsilon$}              & {0.125} & {0.25}  & {0.5}  & {1} & {2}     & {3}    & {4}     \\ \midrule
{PGD-$\ell_{2}$} & {76.4} & {72.1} & {68.9} & {58.5} & {48.2} & {36.4} & {24.3} \\ \bottomrule
\end{tabular}}}

\subtable[{PGD-$\ell_{\infty}$}]{
\resizebox{\linewidth}{!}{
\begin{tabular}{c|ccccccc}
\toprule
  {$\varepsilon$}                  & {2/255} & {4/255} & {8/255} & {16/255} & {24/255} & {32/255} & {40/255} \\ \midrule
{PGD-$\ell_{\infty}$} & {70.5} & {62.2} & {58.0}  & {47.9}   & {35.8}  & {22.4}  & {15.7}   \\ \bottomrule
\end{tabular}}}
\end{table}

\begin{figure}[]
\vspace{-0.15in}
\begin{center}

\subfigure[single layer study]{
\includegraphics[width=0.47\linewidth]{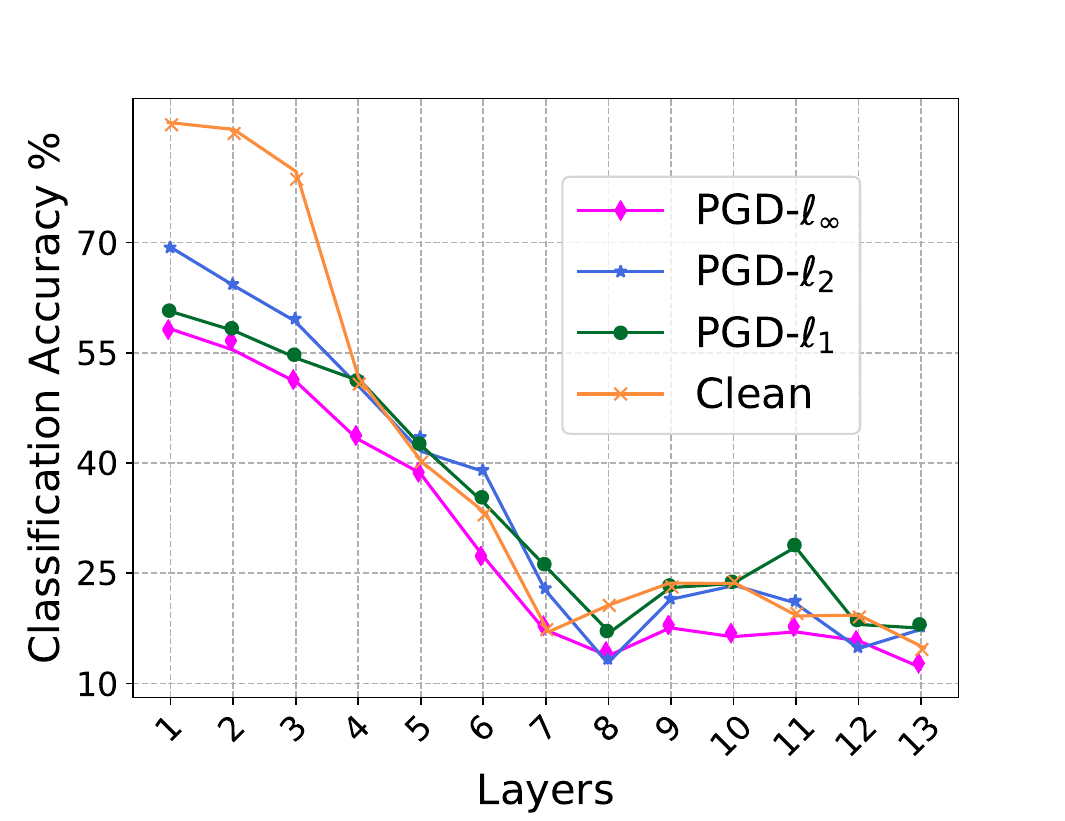}
\label{fig:single_layer}
}
\subfigure[top-m layers study]{
\includegraphics[width=0.47\linewidth]{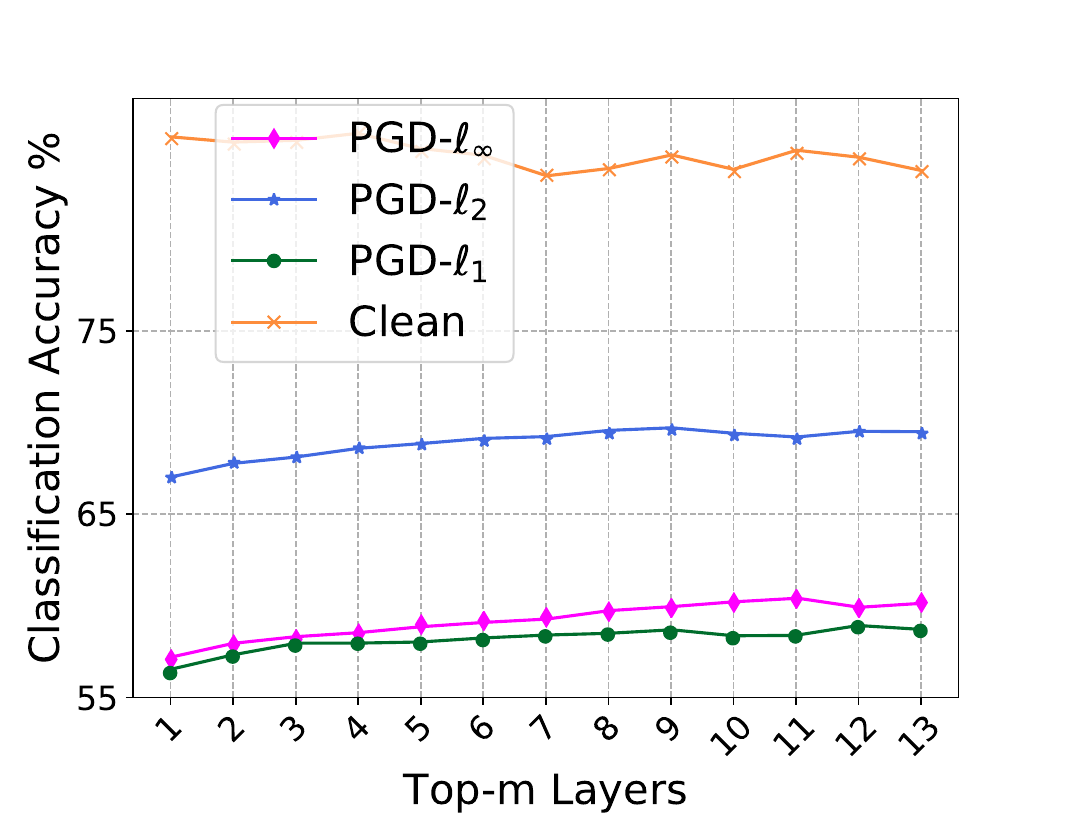}
\label{fig:layer_group}
}

\end{center}
\caption{Adding GBN to different layers (VGG-16 on CIFAR-10): (a) shows the results of adding GBN to single layers; (b) shows the results of adding GBN to top-$m$ layers.}
\end{figure}

\textbf{Effectiveness of the gated sub-network.} As a core part of our GBN block, the performance of the gated sub-network plays a critical role in adversarial defense. Therefore, in this part, we provide a deeper understanding of GBN by investigating the gated sub-networks performance. \revise{Specifically, we conduct two experiments as (1) we report classification accuracies of the gated sub-network of our GBN for different perturbation types (domains), and (2) we remove the 2-way gated sub-network of MBN (denoted as ``vanilla MBN'') and follow the original implementation \cite{Xie2020intriguing} by only using the clean BNs for inference.} For the \emph{first experiment}, against common white-box attacks (Figure~\ref{fig:gate_acc}), the prediction accuracy of the gated sub-network drops with the increasing of the layer depth. The reason might be that low-level statistics learned in shallow layers are easier to train~\citep{Asano2020A}, while the features from different domains are more and more entangled as the layer depth increases, making it harder for models to separate them. \revise{For the \emph{second experiment}, the accuracies of vanilla MBN on CIFAR-10 with ResNet-20 against white-box PGD-$\ell_1$, PGD-$\ell_2$, and PGD-$\ell_{\infty}$ attacks are 21.3\%, 44.8\%, 33.7\%, which show significantly worse robustness compared to MBN with gated sub-network and our GBN (see Table \ref{tab:cifar_sum}). To sum up, the above experiments demonstrate the efficacy of the gated sub-network and we suggest improving the gated sub-network prediction accuracy to further improve model robustness.} 






{\textbf{Adding GBN to different layers.}} Here, we try to add GBN blocks at different layers of the model. Firstly, we add GBN to different single layers. In other words, we only add GBN to a single layer and then adversarially train the model. As shown in Figure ~\ref{fig:single_layer}, the standard performance and adversarial robustness decrease as we only add GBN into the deeper layers. The reasons might be: (1) shallow layers are more critical to model robustness~\citep{liu2019training} and \revise{(2) the features from different domains (\ie, different $\ell_p$-norm adversarial examples and clean examples) are highly entangled in the deeper layers, which
might be harder to separate (\emph{see Supplementary Materials for the running statistics at deeper layers of the model}).}

We then add GBN to layer groups (\ie, top-$m$ layers). As shown in Figure ~\ref{fig:layer_group}, the model robustness improves as more layers are involved, while the clean accuracy remains comparatively stable. Therefore, we add GBN in all layers by default. 

{\textbf{Using gated sub-network only in the first GBN}.}
\label{sec:gate-in-first-gbn-only}
After investigating the performance of GBN on different model layers, we conduct extra experiments by only using a gated sub-network in the first GBN block. In other words, we only include gated sub-network in the first GBN layer, and then send the soft-labels to the following layers. Specifically, the following GBN layers do not contain gated sub-networks and only depend on the prediction results of the first GBN layer to classify the input domains.

The accuracy of a
ResNet-20 on CIFAR-10 against PGD-$\ell_1$, PGD-$\ell_2$, PGD-$\ell_{\infty}$ adversarial examples, and clean examples are 57.3\%, 67.1\%, 57.7\%, and 83.5\%, which is slightly weaker than our implementation. Thus, we use our original implementations in our main experiments (adding gated sub-networks in each GBN layer). \revise{We conjecture the reasons as (1) Model parameter sizes. The model parameter sizes for models with gated sub-networks in the first blocks and in all blocks are 0.3M and 8.8M, respectively. Since larger parameter sizes within the same model family often indicate better robustness \cite{tang2021robustart}, models added GBNs in all layers contain larger parameter sizes and in turn show better robustness. However, we should also notice that these additional parameters contribute primarily to the domain prediction of gated sub-network than the final model classification. (2) Prediction differs at different layers. Given an input, the weights for each domain of the gated sub-network in each layer are not the same for the best defense performance. For more studies, we put them as future work.}

{\textbf{{Replacing BN branches with MLP branches in GBN}.}}
\label{sec:replacement}
{In addition, we try to investigate the critical role of BN branches in our GBN, where we conduct ablations using ResNet-20 on the CIFAR-10 dataset by replacing the BN branches with MLP branches. In particular, we replace the BN branches after the gated subnetwork with a convolution layer (we set the kernel size as 3*3, stride as 1, and the input channel the same as the output channel). However, since there are no BN layers, the training is unstable and could not converge, which further demonstrate the importance of BN in our GBN approach.}

%% file: 5_discussion.tex
\section{Discussion and Analysis}

\subsection{{Multi-domain hypothesis}}
\label{sec:multi-abla}

Here, we provide deeper understanding and verification of our multi-domain hypothesis.

\begin{figure}[]
\vspace{-0.1in}
\centering


\subfigure[running means]{
\includegraphics[width=0.46\linewidth]{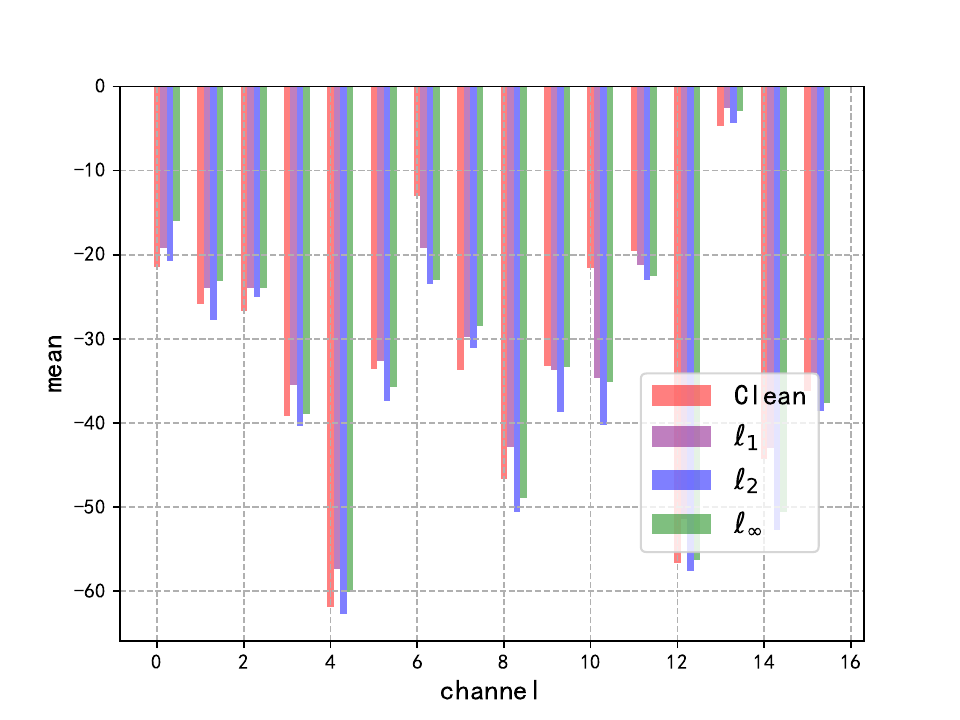}
}
\subfigure[running variances]{
\includegraphics[width=0.46\linewidth]{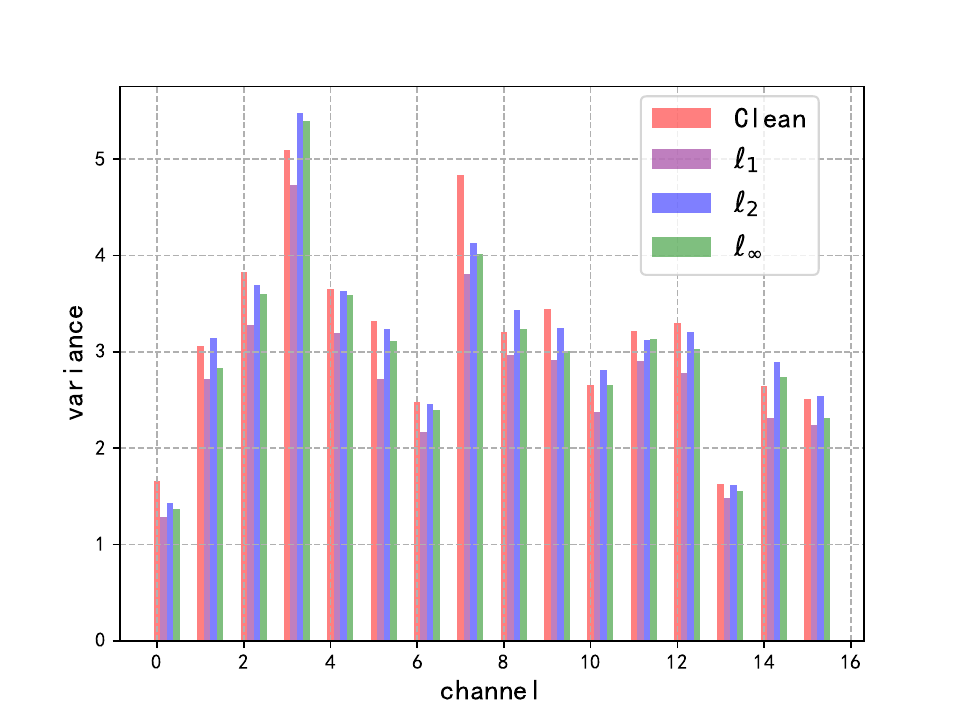}
}

\subfigure[running means]{
\includegraphics[width=0.46\linewidth]{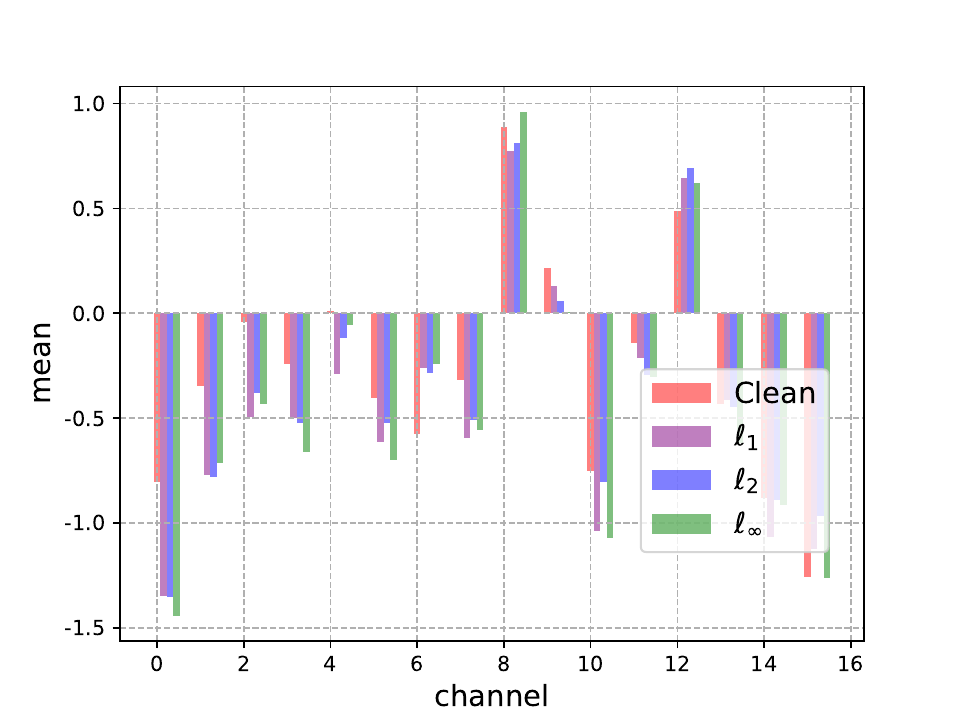}
}
\subfigure[running variances]{
\includegraphics[width=0.46\linewidth]{layer2.2.bn1_var-used.pdf}
}

\subfigure[\revise{running means}]{
\includegraphics[width=0.46\linewidth]{bn31_mean.pdf}
}
\subfigure[\revise{running variances}]{
\includegraphics[width=0.46\linewidth]{bn31_var.pdf}
}

\subfigure[\revise{running means}]{
\includegraphics[width=0.46\linewidth]{layer2.2.bn1_mean-used.pdf}
}
\subfigure[\revise{running variances}]{
\includegraphics[width=0.46\linewidth]{layer3.2.bn1_var-used.pdf}
}
\caption{Running statistics (running mean and variance) of each BN in the multiple BN branches at different layers on models trained on CIFAR-10. \revise{Here, we randomly choose 16 channels for visualization.} (a) and (b) show the layer $conv3\_1$ on VGG-16; (c) and (d) denote the layer $block1.layer1.bn2$ on WideResNet-28-10; \revise{(e) and (f) denote the layer $layer1.1.bn1$ on ResNet-20; \revise{(g) and (h) denote the layer $block1.layer1.bn2$ on WideResNet-34-10.}}}
\label{fig:running_vgg}
\end{figure}

{\textbf{Visualization of running statistics on different models}.}\label{sec:visual-running} In this part, we provide more running statistics of different deep models to further verify our multi-domain hypothesis. \revise{In particular, we train a VGG-16, WideResNet-28-10, ResNet-20, and WideResNet-34-10 model with the multiple BN branch structure.} {The running statistics of each BN branch at different layers \revise{(16 randomly selected channels)} are shown in Figure \ref{fig:running_vgg}. From the results, we can further observe that different perturbation types induce different BN statistics, which may arise in different domains.} 

{\textbf{Fourier perspective analysis}} To verify our multi-domain hypothesis, we study the differences between perturbation types from the Fourier perspective \citep{Yin2019Fourier}. We visualize the Fourier heatmap following \cite{Yin2019Fourier} on 4 different ResNet-20 models (\ie, a vanilla model and 3 adversarially-trained models using only $\ell_1$, $\ell_2$, and $\ell_{\infty}$ adversarial examples, respectively) and shift the low frequency components to the center of the spectrum. Error rates are averaged over 1000 randomly sampled images from the test set on CIFAR-10. The redder the zone, the higher the error rate of the model to perturbations of specific frequency.

\begin{figure}[]
\vspace{-0.1in}
\centering
\includegraphics[width=1.05\linewidth]{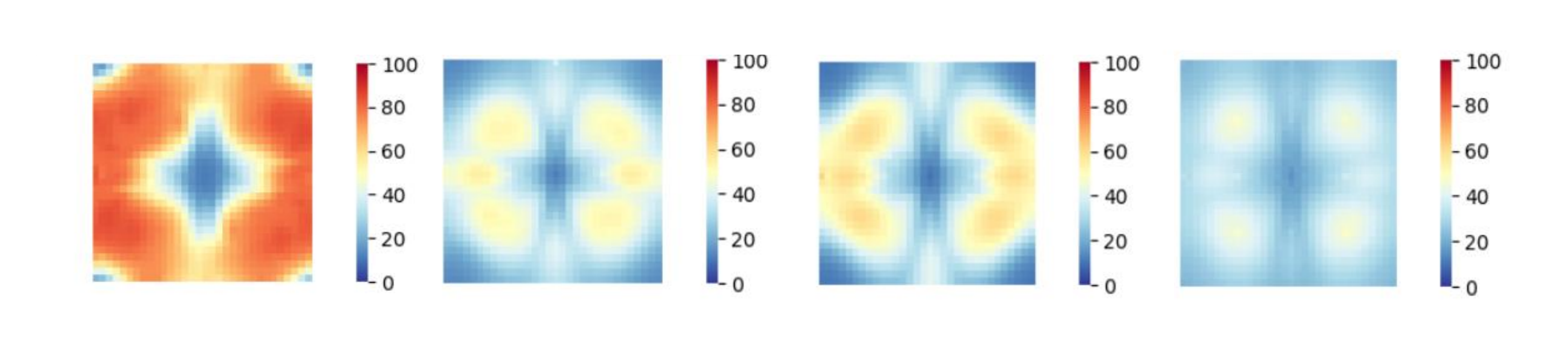}

\vspace{-0.15in}
\caption{Fourier analysis on model sensitivity to additive noise aligned with different Fourier basis vectors on CIFAR-10. From left to right: vanilla model, PGD-$\ell_1$ trained model, PGD-$\ell_2$ trained model, and PGD-$\ell_{\infty}$ trained model. The numbers indicate the model error rates.}
\label{fig:fourier_sen}
\end{figure}

As shown in Figure \ref{fig:fourier_sen}, models trained using different $\ell_p$ perturbations demonstrate different model weaknesses in the Fourier heatmap (\ie, different hot zones). For example, model trained on PGD-$\ell_2$ is more susceptible to high-frequency perturbations while less susceptible to some low-frequency perturbations (the smaller blue zone in the center), while model trained on PGD-$\ell_{\infty}$ is comparatively more robust to middle-frequency perturbations (light yellow zones). Therefore, different perturbation types have different frequency properties, and models trained on different perturbations are sensitive to perturbations from different frequencies. This observation further demonstrates that different perturbation types may arise from different domains.

{\textbf{Training with different ratios of perturbation types on a single BN}.} {Our Theorem~\ref{th:dist} demonstrates that training with a mixture of perturbation types on a single BN would cause an inevitable domain gap between training and testing data, which causes the performance degeneration. To verify this, we hereby train several individual models with different ratios of perturbation types for the estimation of the single BN layer while using the same perturbation type (\ie, $\ell_1$) to update the convolution layers of the model.} Specifically, for each individual model with a single BN, we use different ratios of perturbation types (\eg, clean, $\ell_1$, $\ell_2$, and $\ell_{\infty}$ adversarial examples) to estimate the running statistics of BN, but only use $\ell_1$ adversarial examples to optimize the model parameters (\eg, convolutions). For model $M_1$, we use $\ell_1$ adversarial examples to estimate the running statistics of the BN layer; for model $M_2$, we use $\ell_1$ and $\ell_2$ adversarial examples to estimate the running statistics of the BN layer; for model $M_3$, we use clean example, $\ell_1$, and $\ell_{\infty}$ adversarial examples to estimate the running statistics of the BN layer; for model $M_4$, we use clean example, $\ell_1$, and $\ell_{\infty}$ adversarial examples to estimate the running statistics of the BN layer; for model $M_5$, we use clean example, $\ell_1$, $\ell_2$, and $\ell_{\infty}$ adversarial examples to estimate the running statistics of the BN layer. For all these models, we use $\ell_1$ adversarial examples to optimize the model weights, and we keep the same amount of training data. For all these models, they use different ratios of perturbations to estimate the BN layer, and use the same data to optimize the model weight. {Thus, we could control the variables and the differences between these models ($M_1$ to $M_5$) are the estimation of BN layer.}

On CIFAR-10, the predication accuracy for PGD-$\ell_1$ attacks (the same settings for our main experiment) from $M_1$ to $M_5$ is shown as: \textbf{92.5\%}, 84.0\%, 89.1\%, 85.5\%, and 83.8\%, respectively. $M_1$ achieves the strongest robustness on $\ell_1$ attacks, which empirically verify our Theorem~\ref{th:dist} that the model achieves the best performance when the proportion of the different perturbations in the training set is most close to that in the test set. Thus, we should not learn a mixture distributions with a single BN.

\subsection{Feature visualization before/after normalization}
\label{sec:t-sne}

\begin{figure}[]
\vspace{-0.15in}
\begin{center}

\subfigure[before GBN block]{
\includegraphics[width=0.47\linewidth]{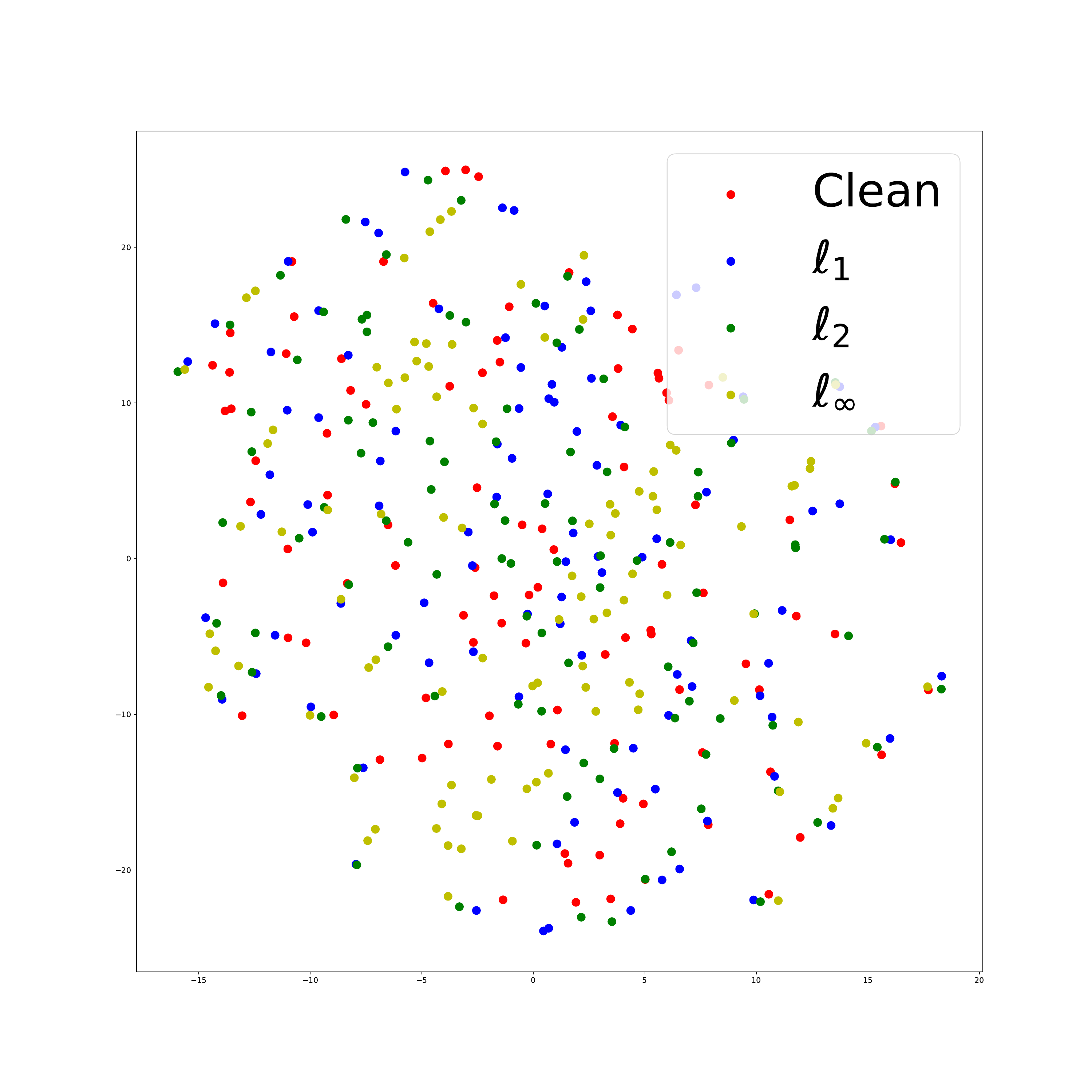}
\label{fig:gbn_before}
}
\subfigure[after GBN block]{
\includegraphics[width=0.47\linewidth]{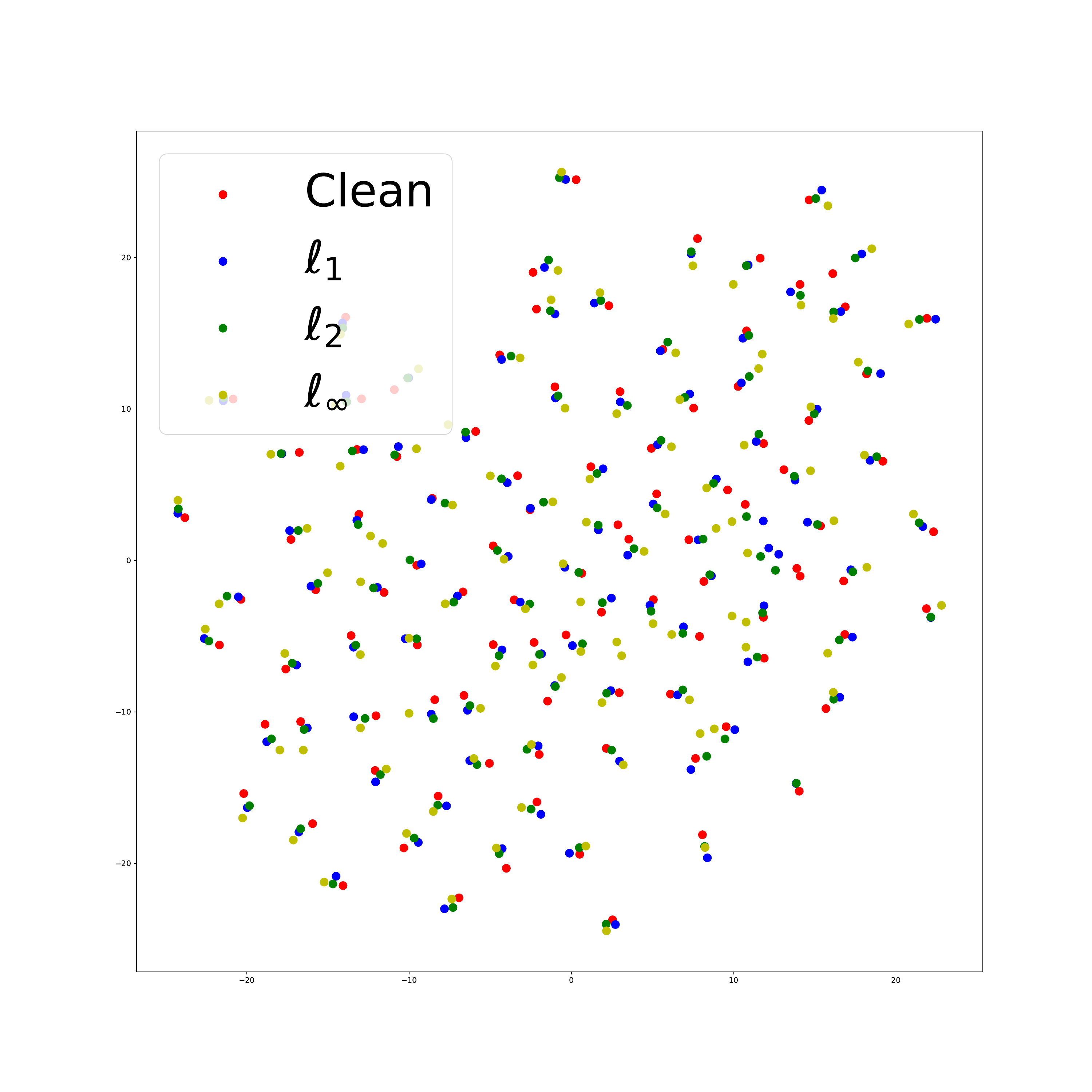}
\label{fig:gbn_after}
}
\subfigure[before BN block]{
\includegraphics[width=0.47\linewidth]{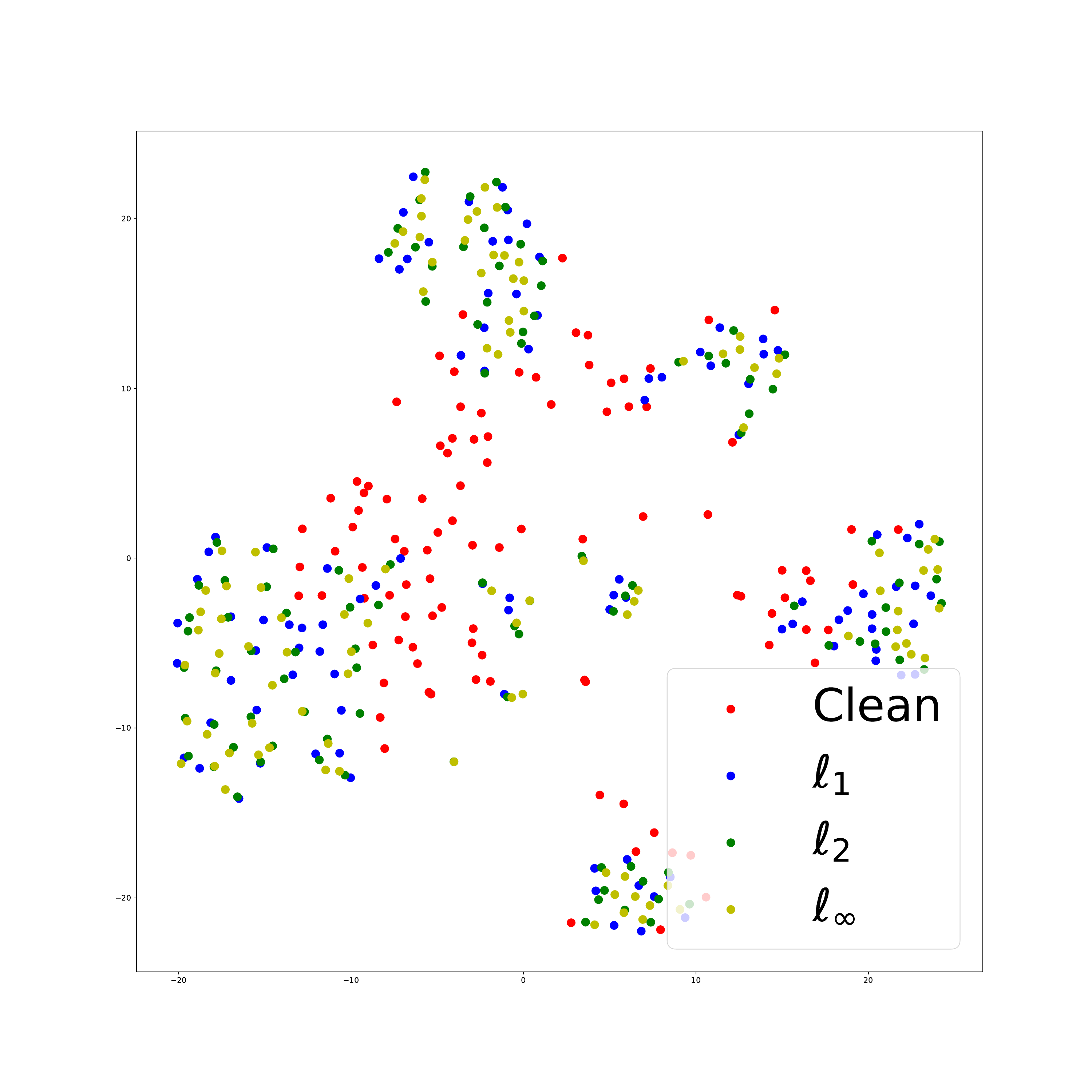}
\label{fig:bn_before}
}
\subfigure[after BN block]{
\includegraphics[width=0.47\linewidth]{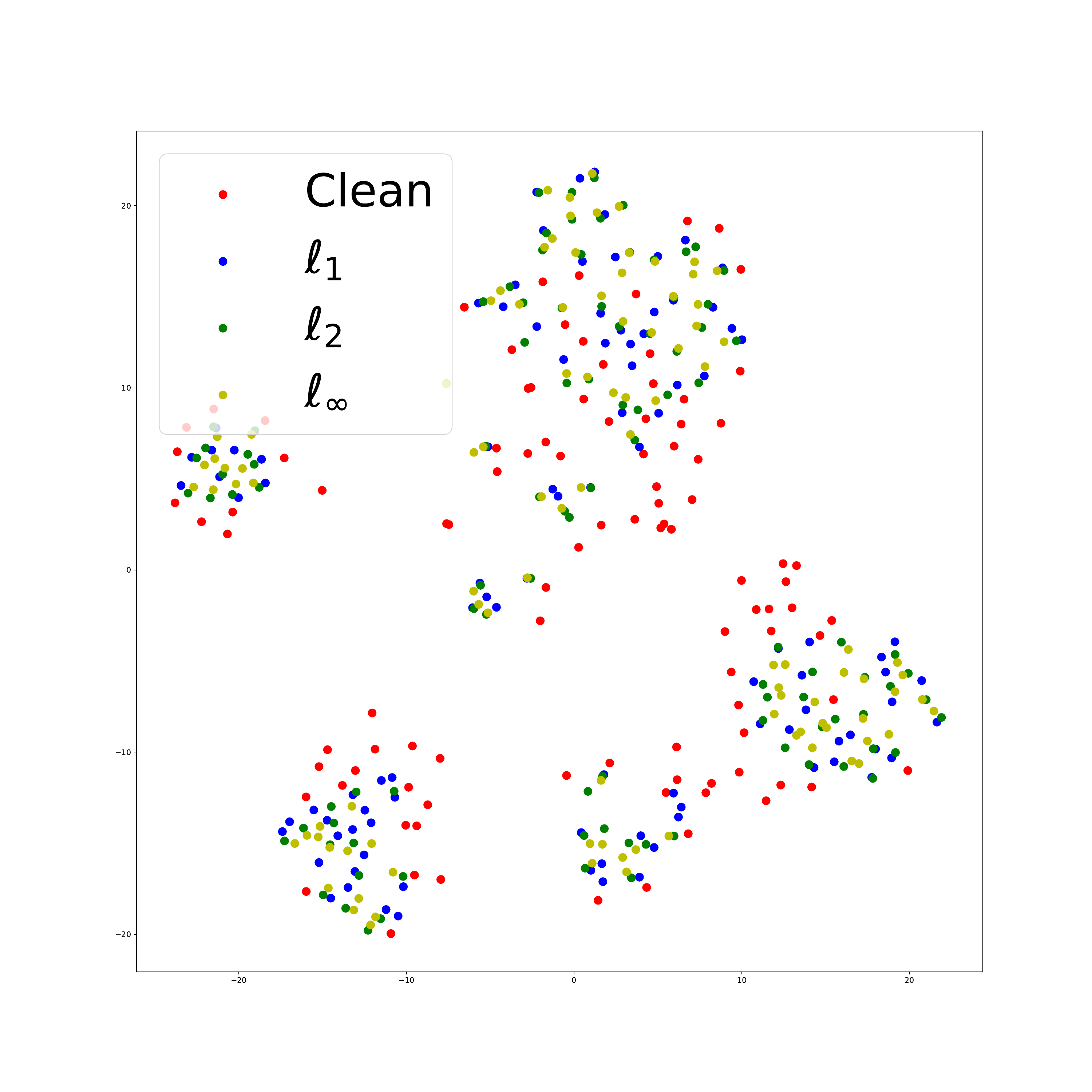}
\label{fig:bn_after}
}

\end{center}
\caption{Feature visualization. We use the BN block at layer $conv3\_1$ of a VGG-16 model trained using AVG (c-d), and the GBN block at layer $conv3\_1$ of a VGG-16 (a-b). We visualize the features before and after BN/GBN blocks. Features of different domains are aligned to domain-invariant representations after GBN, while distances between the clean example and its corresponding adversarial examples are still far after BN.}
\label{fig:t-sne-bn}

\end{figure}

In this part, we aim to better understand the effect of our GBN via feature visualization. Specifically, we use t-SNE~\citep{Laurens2008Visualizing} to visualize the features before and after the normalization. As shown in Figure \ref{fig:t-sne-bn}, the features of different domains are aligned to domain-invariant representations after GBN, \ie, one clean example (the red dot) and its corresponding adversarial examples (the blue, green, and yellow dots) tend to be close in the space of normalized output. However, after the BN block of AVG, distances between the clean example and its corresponding adversarial examples are still far. All the above experiments are conducted on CIFAR-10 with VGG-16.

Therefore, we can draw an important conclusion that our GBN provides model robustness against multiple perturbation types by normalizing them into aligned features for better learning. In particular, GBN first aligns the distribution among different perturbations types by using one BN branch to handle one perturbation type (learn domain-specific features), and then the aligned distribution contributes to the subsequent modules or layers (\eg, convolutional layer) for learning domain-invariant representations.

\begin{table}[]
\caption{Classification accuracy (\%) of ResNet-20 on CIFAR-10, where models need to generalize to attack types not included for training (the higher the better). \revise{``$\ell_1$+$\ell_2$+$\ell_{\infty}$'',``$\ell_1$+$\ell_{\infty}$'', and ``$\ell_2$+$\ell_{\infty}$'' denote models trained on different combinations of perturbation types.}}
\small
\label{tab:out_model_attack}
\begin{center}
\small
\resizebox{\linewidth}{!}{
\begin{tabular}{cc|cccc}
\toprule
                            &         & Clean                    & PGD-$\ell_1$                   & PGD-$\ell_2$                   & PGD-$\ell_{\infty}$ \\
                            \hline
                            
                            & Vanilla & 0.1                      & 0.0                      & 0.0                      & 89.4                     \\
                            \hline
\multirow{6}{*}{$\ell_1$+$\ell_2$+$\ell_{\infty}$} & AVG     & 80.0                     & 47.8                     & 57.2                     & 35.2                     \\
                            & MAX     & 76.7                     & 43.9                     & 54.3                     & 38.6                     \\
                            & MSD     & 78.4                     & 49.3                     & 62.3                     & 42.2                     \\
                            & \revise{MN}      & \revise{\textbf{82.0}}                     & \revise{43.7}                     & \revise{48.9}                     & \revise{32.1}                     \\
                            & \revise{MBN}     & \revise{79.1}                     & \revise{47.3}                     & \revise{60.1}                     & \revise{54.8}                     \\
                            & GBN     & 80.2                     & \textbf{58.1}                     & \textbf{68.9}                     & \textbf{58.0}                    \\
                            \hline
                            
\multirow{5}{*}{$\ell_1$+$\ell_{\infty}$}                    & AVG     & 74.2                     & 50.2                     & 58.7                     & 35.2                     \\
                            & MSD     & 71.2                     & 49.4                     & 60.0                     & 44.1                     \\
                            & \revise{MN}      & \revise{\textbf{83.1}}                     & \revise{41.0}                     & \revise{47.2}                     & \revise{28.9}                     \\
                            & \revise{MBN}     & \revise{80.3}                     & \revise{47.8}                     & \revise{58.4}                     & \revise{54.1}                     \\
                            & GBN     & 81.1                     & \textbf{57.1}                     & \textbf{64.1}                     & \textbf{57.2}                     \\
                            \hline
                            
\multirow{6}{*}{$\ell_2$+$\ell_{\infty}$}                     & AVG     & 77.0                     & 30.0                     & 57.7                     & 34.0                     \\
                            & MSD     & 73.0                     & 23.2                     & 56.1                     & 45.9                     \\
                            & \revise{MN}      & \revise{\textbf{82.0}}                     & \revise{25.1}                     & \revise{47.4}                     & \revise{30.3}                     \\
       & \revise{MBN}     & \revise{79.4} & \revise{30.1} & \revise{58.9} & \revise{53.2} \\
      & GBN     & 80.8 & \textbf{40.3} & \textbf{64.7} & \textbf{56.3} \\
      \bottomrule
\end{tabular}}
\end{center}

\end{table}

\subsection{Defending unseen $\ell_p$ perturbations}
Finally, we examine the generalization of GBN to unseen $\ell_p$ bounded adversarial perturbations. In other words, during testing, the model may encounter the types of adversarial examples that has never been trained on. Specifically, we train a variant of GBN that only includes 3 BN branches. These GBN models are trained on two $\ell_p$ perturbations only, but are evaluated on all the three $\ell_p$ perturbations, including the unseen perturbation. 

As shown in Table~\ref{tab:out_model_attack}, unsurprisingly, compared to the full GBN model with 4 BN branches and trained on (clean examples, $\ell_1$, $\ell_2$, and $\ell_{\infty}$ adversarial examples), the robustness of the held-out perturbation type decreases. However, the robustness is still significantly better than other baselines with the same setting, and sometimes even outperforms others that are trained on all perturbation types.

\begin{table}[!t]

\caption{GBN in batch-unrelated normalization scenario. Results are shown using accuracy (\%) of ResNet-20 on CIFAR-10 on different types of perturbations.}

\begin{center}
\subtable[Comparison with models trained on single GN, LN, and IN]{
\resizebox{\linewidth}{!}{
\label{tab:GBN-GN}
\begin{tabular}{ccccc}
\toprule
& {Clean} & {PGD-$\ell_1$} & {PGD-$\ell_2$} & {PGD-$\ell_{\infty}$}\\
\hline
\textbf{{GBN(ours)}} & \textbf{{80.2}} & \textbf{{58.1}} & \textbf{{68.9}} & \textbf{{58.0}}\\

{BN} & {80.0} & {47.8} & {57.2} & {35.2} \\

{GN} & {76.4} & {48.7} & {57.5} & {35.2} \\

{LN} & {75.9} & {47.6} & {58.1} & {36.0} \\

{IN} & {67.9} & {46.4} & {53.9} & {32.8} \\

\bottomrule
\end{tabular}}}

\subtable[\revise{Replacing LN with GBN}]{
\resizebox{\linewidth}{!}{
\label{tab:GBN-VIT}
\begin{tabular}{ccccc}
\toprule
& \revise{Clean} & \revise{PGD-$\ell_1$} & \revise{PGD-$\ell_2$} & \revise{PGD-$\ell_{\infty}$}\\
\hline
\textbf{\revise{ViT$_{GBN}$}} & \textbf{\revise{81.0}} & \textbf{\revise{44.5}} & \textbf{\revise{51.9}} & \textbf{\revise{49.9}}\\

\revise{ViT$_{LN}$} & \revise{78.8} & \revise{38.9} & \revise{31.2} & \revise{46.1}\\

\bottomrule
\end{tabular}}}

\end{center}
\vspace{-0.1in}
\end{table}

\subsection{{Batch-unrelated normalization scenario}}
{In contrast to BN that computes statistics on the batch of samples, there also exists several normalization techniques that use batch-unrelated information for estimating
normalization statistics such as Group Normalization (GN) \cite{wu2018group}, Instance Normalization (IN) \cite{dmitry2016instance}, and Layer Normalization (LN) \cite{ba2016layer}. \revise{Therefore, we further investigate the potential and effectiveness of our GBN in the batch-unrelated scenario.}

\revise{\textbf{Adversarial training for GN/IN/LN}}. We first adversarially train models with the batch-unrelated normalization layer GN, LN, and IN, so that we could avoid exploiting the batch dimension to calculate statistics. Specifically, we replace all the GBN blocks in ResNet-20 with GN or IN and adversarially train the models on CIFAR-10 using AVG. For GN, we split the channel into 4 groups; for LN and IN, we use the default setting. From the results in Table \ref{tab:GBN-GN} we can observe that the robust accuracy for ResNet-20 with GN/LN/IN under PGD-$\ell_1$, PGD-$\ell_2$, and PGD-$\ell_{\infty}$ attacks are much lower than ResNet-20 with GBN blocks, which demonstrates the effectiveness of our GBN. Meanwhile, we also found that (1) single GN achieves slightly better robustness than single BN, which somewhat reveals the fact that erasing the adversarial domain-related statistics may help improve the robustness for adversarial training; and (2) single IN shows the lowest performance on both clean and adversarial examples, which might be caused by its strict sample-wise constraints that influence the model learning ability. 

\revise{\textbf{Replacing LN with GBN.} In addition, we try to study whether the concept of GBN can be transferred to other normalization layers. Specifically, we select the Vision Transformers (ViTs) architecture \cite{dosovitskiy2020image} that uses LN as default and replace the LN layers with our GBN layers; we then adversarially train the ViT with GBN (denoted as ``ViT$_{GBN}$''). We also report the results of adversarial training on ViT with LN (denoted as ``ViT$_{LN}$'') using AVG. As shown in Table \ref{tab:GBN-VIT}, on CIFAR-10, the robust accuracies for ViT with GBN blocks are higher than ViT with LNs, which further demonstrates the potential of our GBN in other normalization layers. We will explore the possibilities of extending our motivation to GLN (gated layer normalization) in the future.}

\subsection{{Computational overhead}}
\revise{Since our method imports additional blocks into DNNs, we further analyze its computational cost in this part. As for the parameters, in each layer, we have $N$+1 pairs of $\{\hat{\mu}, \hat{\sigma}\}$ to estimate compared to one pair of standard BN (other baselines also adopt models that contain standard BN layers); our GBN also introduces extra learnable parameters $\theta$ in the gated sub-network.}

We then empirically evaluate the time consumption during training and inference for our GBN and other defenses. All the experiments are conducted on an NVIDIA Tesla V100 GPU cluster. We compute the overall training time using ResNet-20 on CIFAR-10 for 40 epochs and the inference time for 10000 images. According to the results in Table \ref{tab:time}, our GBN spends the comparative time during training and inference with other adversarial defenses. \revise{During training, our GBN has extra parameters $\theta$ in the gated sub-network to learn. However, the most time-consuming procedure of adversarial training involves generating adversarial examples. Therefore, our training time cost is comparable to other adversarial defense baselines. As for the inference, our GBN requires to classify the input samples into the specific domain using the gated sub-network. Therefore, our method is slightly slower than methods that are solely training the models (\eg, AVG). However, we achieve similar and even faster speeds compared to other methods that require run-time predictions on input domains (\eg, MN). Therefore, though extra parameters are introduced, our GBN achieves comparable time cost to other defenses.}

\begin{table}[!t]
\caption{\revise{Training and inference time evaluation of different methods using ResNet-20 on CIFAR-10.}}
\label{tab:time}

\begin{center}
\footnotesize
\setlength{\tabcolsep}{0.6mm}

\begin{tabular}{ccccccc}
\toprule
 & {AVG} & {MAX} & {MSD} & {MN} & {MBN}& \textbf{{GBN(ours)}}\\
\hline \\
 Training speed\\(seconds/ epoch) & 1837.9 & 1113.2 & 689.8& 19329.1 & 2737.9 & 2802.6 \\
 \hline\\
 Training duration\\(hours) & {20.4} & {12.4}& {7.7}& {214.8}& {30.4}& {31.1}\\
 \hline \\
 Inference speed\\(seconds/ 10000 images) & {0.6} & {0.6}& {0.6} & {4.8} & {1.9} & {2.0} \\
\bottomrule
\end{tabular}
\end{center}
\end{table}

%% file: 6_conclusion.tex
\section{Conclusions}

Most adversarial defenses are tailored to a single perturbation type (\eg, small $\ell_{\infty}$-noise), but offer no guarantees against multiple $\ell_p$ bounded adversarial attacks. To better understand this phenomenon, we explored the \emph{multi-domain} hypothesis that different $\ell_p$ bounded adversarial perturbations are drawn from different domains. Thus, we propose a novel building block for DNNs, \emph{Gated Batch Normalization} (GBN), which consists of a gated network and a multi-branches BN layer. The gated sub-network separates different perturbation types, and each BN branch is in charge of a single perturbation type and learns the domain-specific statistics for input transformation. Then, features from different branches are aligned as domain-invariant representations for the subsequent layers. Extensive experiments on MNIST, CIFAR-10, and Tiny-ImageNet demonstrate that our GBN outperforms previous proposals against $\ell_1$, $\ell_2$, and $\ell_{\infty}$ adversarial attacks by large margins.

Though promising, our GBN block introduces extra parameters and requires additional computation, and we will investigate ways to accelerate training in future work. Also, we are interested in applying GBN in more complex scenarios to defend other unseen noises (\eg, common corruptions).